# On the Evolution of A.I. and Machine Learning: Towards a Meta-level Measuring and Understanding Impact, Influence, and Leadership at Premier A.I. Conferences

Rafael B. Audibert
*Institute of Informatics, Federal University of Rio Grande do Sul, Brazil*
rbaudibert@inf.ufrgs.br

Henrique Lemos
*Institute of Informatics, Federal University of Rio Grande do Sul, Brazil*
hlsantos@inf.ufrgs.br

Pedro Avelar
*Department of Informatics, Kings College London*
*Machine Intellection Department, Institute for Infocomm Research, A*STAR, Singapore*[*]
pedro_henrique.da_costa_avelar@kcl.ac.uk

Anderson R. Tavares
*Institute of Informatics, Federal University of Rio Grande do Sul, Brazil*
artavares@inf.ufrgs.br

Luís C. Lamb
*Institute of Informatics, Federal University of Rio Grande do Sul, Brazil*
*MIT Sloan School of Management, Cambridge, MA*
lamb@inf.ufrgs.br, luislamb@mit.edu

[*]Work done while at Institute of Informatics, Federal University of Rio Grande do Sul, Brazil

**Abstract**

Artificial Intelligence is now recognized as a general-purpose technology with ample impact on human life. This work aims at understanding the evolution of AI and, in particular Machine learning, from the perspective of researchers' contributions to the field. In order to do so, we present several measures allowing the analyses of AI and machine learning researchers' impact, influence, and leadership over the last decades. This work also contributes, to a certain extent, to shed new light on the history and evolution of AI by exploring the dynamics involved in the field's evolution by looking at papers published at the flagships AI and machine learning *conferences* since the first International Joint Conference on Artificial Intelligence (IJCAI) held in 1969. AI development and evolution have led to increasing research output, reflected in the number of articles published over the last sixty years. We construct comprehensive citation-collaboration and paper-author datasets and compute corresponding centrality measures to carry out our analyses. These analyses allow a better understanding of how AI has reached its current state of affairs in research. Throughout the process, we correlate these datasets with the work of the ACM Turing Award winners and the so-called two AI winters the field has gone through. We also look at self-citation trends and new authors' behaviors. Finally, we present a novel way to infer the country of affiliation of a paper from its organization. Therefore, this work provides a deep analysis of Artificial Intelligence history from information gathered and analysed from large technical venues datasets and suggests novel insights that can contribute to understanding and measuring AI's evolution.

# 1 Introduction

Artificial Intelligence is now seen as a general-purpose technology that impacts the world's economy in significant ways Crafts [2021]. AI research started in academia, where seminal works in the field defined trends first in machine intelligence McCulloch and Pitts [1943], Turing [1950] and later on the early development and organization of the area ranging from symbolic to connectionist approaches Feigenbaum and Feldman [1963], Minsky [1961]. However, AI has become more than a research field explored in-depth in academia and research organizations. Applied AI research has led to uncountable products, technologies, and joint research between universities and industry, see e.g., Gomez-Uribe and Hunt [2016], Ramesh et al. [2021], Amini et al. [2020]. Recent business research Gartner [2019] has shown that AI is now being implemented widely in organizations, at least to some extent. AI research has led to groundbreaking results that caught the media's attention. For instance, in the 1990s, Deep Blue Campbell et al. [2002] became the first computing system to win a chess match against the then reigning chess world champion, Garry Kasparov, under tournament conditions.

Later, AI research would eventually lead to even higher grounds in many applications. AlphaGo Silver et al. [2016] has won a series of matches against Go world champions, Brown et al. [2020] can generate texts that suggest a future of possibly human-like competence in text generation, Cobbe et al. [2021] has shown how to solve math word problems, Jumper et al. [2021] significantly improved 3D protein structure prediction, and Park et al. [2019] can render seemingly authentic life-like images from segmentation sketches, to name a few.

Even though the area has seen a noticeable technological impact and progress, we claim that there is a need to analyse the history and evolution of AI and the dynamics involved in transforming it into a well-established field within Computer Science. Some influential researchers, such as Gary Marcus, have discussed the developments that happened in the area in recent years Marcus [2018]. Moreover, Marcus reflected upon what is to come in the next decade Marcus [2020]. The current debate has also motivated the research on new approaches to integrating the symbolic and connectionist paradigms, leading to Neurosymbolic AI d'Avila Garcez and Lamb [2003, 2006], d'Avila Garcez et al. [2009]. This approach aims to achieve more robust AI systems endowed with improved semantic understanding, cognitive reasoning, and learning abilities d'Avila Garcez and Lamb [2023], Riegel et al. [2020], Besold et al. [2022]. Further, it is even more evident now than in the dawn of AI that the field not only draws inspiration from – but also inspires advances in other areas – including cognitive psychology, neuroscience, economics, and philosophy see, e.g., Booch et al. [2020], Marcus and Davis [2021], Smolensky et al. [2022]. Kautz has recently pointed out that given AI's recent contributions, "we may be at the end of the cyclical pattern; although progress and exuberance will likely slow, there are both scientific and practical reasons to think a third winter is unlikely to occur.Kautz [2022]" making the case that we might not see another AI Winter shortly. The section on AI history briefly details the field's evolution.

## A Note on Methodology and Contributions

In this paper, we look further back in Artificial Science history, explore its evolution, and contribute to understanding what led to the AI impacts we have in society today. To do so, we will investigate how the collaboration and citation networks of researchers evolved since 1969 [1] within three major AI conferences. We start our analyses with IJCAI, AAAI, and NeurIPS, *together with flagship conferences*

[1]The date was chosen because it marks the first International Joint Conference on Artificial Intelligence - IJCAI-69.

*of related research areas which are impacted and influenced by AI*, namely ACL, EMNLP, NAACL, CVPR, ECCV, ICCV, ICML, KDD, SIGIR, and WWW. Even though not all of these conferences had many AI-related papers published in their early years, more recently, it is clear that AI techniques are playing a more prominent role in these research communities. Therefore, we add them to our analyses to compose a "big picture" of how AI has not only grown itself but also gradually started to influence other fields. These include, e.g., computer vision, image processing, natural language processing, and information retrieval.

We proceed by exploring and enhancing an extensive dataset of papers published in Computer Science and AI venues since 1969, the v11 Arnet datasetTang et al. [2008]. We use version v11 from this data dataset, containing data originating from DBLP with further disambiguation regarding paper authorship, spanning from 1969 to 2019. There are Arnet versions v12 and v13 available with data until 2021. However, the data for the recent years are somewhat degraded in these recent datasets, thus rendering their statistical analysis inadequate and error-prone (See Section 3.1 to understand our trade-offs on using v11 instead of v13). We then use this dataset to create a new dataset of our own, modeled in several different graph representations of the same data, allowing us to explore them in a true network fashion. With their centralities already computed, these graphs are available for future research. The process to generate them involves using considerable compute power, with amounts of memory and processing not easily found outside laboratories at large-scale research universities or corporations.

Using citation and collaboration networks, our analyses then use centralities to rank both papers and authors over time. We then correlate these rankings to external factors, such as conferences' localization or the ACM's Turing Award – the most prestigious research recognition in Computer Science. These data will allow us to explore what/who, were/are the influential papers/authors in every area/venue under study here. Additionally, we will also examine the dynamics of where all this research is being produced, trying to understand the recent growth of scientific output in China concerning the other countries that led the first decades of AI research.

In these analyses (Section 4), we try to understand and show how authors do not maintain their influence in the field for an extended period. We also analyse this influence regarding ranking papers by importance, as papers can be considered relevant for a more extended period. We also show that the *average number of authors per paper is increasing* in the analysed venues and the number of self-citations too. Furthermore, we also investigate the authors who introduce most co-authors to these conferences. We also show the dynamics behind citations between conferences, showing how some meetings work better together than others.

Because of the nature of our work – converting large amounts of non-organized data into a structured data format – we also generate some side contributions besides our main work. These contributions are: *(i) a new and efficient Python library to convert XML to JSON that uses file streams instead of loading the whole data in memory; (ii) a parallel Python implementation to compute some centrality measures for graphs, using all physical threads available in a machine, and (iii) a novel structure to avoid reprocessing data already processed when its underlying structure is a graph.*

### Paper Structure

We organize our paper as follows: Section 2 provides a brief history of Artificial Intelligence[2], some background information on the analysed computer science conferences, the ACM's Turing Award, and a review of graph theoretical concepts. Section 3 describes the methodology, including information about the underlying dataset and the process behind the generation of the graphs/charts used throughout this work. Section 4 presents and discusses the analyses of the aforementioned data from various perspectives. Section 5 concludes our work and presents suggestions for future work using the new datasets. The Appendix brings some tables and figures that we avoid including in the main body of the paper to facilitate the reading flow.

# 2 Background

## 2.1 A Short on Artificial Intelligence History

Artificial Intelligence history is generally defined in terms of main time periods, where the field grew stronger, interluded by two periods (the so-called AI Winters) where the area was somewhat discredited and underfunded and thought to be of limited real-world impact. The coming material is not exhaustive but provides historical background to understand how the data analysed here relates to these periods. Several works describe aspects of AI history under different perspectives on how the field evolved in time, see e.g. Kautz [2022], Russell and Norvig [2020].

[2]We present a brief, non-comprehensive history of AI, focusing only on topics related to the analysed AI publications. Our historical account is by no means comprehensive. We choose to focus only on the topics related to the AI venues analysed in *this* paper. Of course, a comprehensive history of AI would not fit into the space of a research paper. Therefore, the reader should not see or interpret our brief historical account as definitive.

### 2.1.1 The Dawn of AI *(1940-1974)*

Although debatable, some of the first XX-century "modern artificial intelligence" papers were published in the 1940s. One of the first artificial neural network-related papers arguably dates back to 1943, when Warren McCulloch and Walter Pitts formally defined how simple logical operations from propositional logic could be computed in a connectionist setting McCulloch and Pitts [1943]. Later, in 1950, Alan Turing published the widely cited "Computing Machinery and Intelligence" paper Turing [1950], one of the first philosophical papers reflecting upon the interplay of intelligence and machines and on the possibility of building intelligent machines. In this paper, Turing reflects if machines are able to think and also proposes the "imitation game" (now widely known as the Turing Test) in order to verify the reasoning and thinking capabilities of computing machines. Nevertheless, it was in 1956 that the term Artificial Intelligence (AI) was coined by John McCarthy during the Dartmouth Summer Research Project on Artificial Intelligence workshop. From the workshop onward, A.I. rapidly evolved into a potentially world-changing research field – at that time, especially focusing on the symbolic paradigm, influenced by logical reasoning and its computational implementations. One of the first collections of artificial intelligence articles would be published in Feigenbaum and Feldman [1963].

A well-known example of rule-based systems of the 1960s is Eliza Weizenbaum [1966], the first-ever chatbot, created in 1964, by Joseph Wiezenbaum at the Artificial Intelligence Laboratory at MIT. Today's chatbot market is considerably large, powering multi-million dollar companies revenues like Intercom[3] or Drift[4]. Eliza was created to be an automated psychiatrist, as if the human was talking to someone who understood their problems, although the system worked in this rule-based format, replying to the user with pre-fed answers. Besides the main artificial intelligence approach, we can already see how related areas are easily influenced with a chatbot clearly involving natural language processing as well.

It would also be in 1964 that Evans [1964] would show that a computer could solve what they described as "Geometry Analogy Problems", which correlates with the problems usually displayed in IQ tests where one needs to solve a question in the format "figure A is to figure B as figure C is to which of the given answer figures?" such as the one represented in Figure 1

Important research would also vouch in favor of the area, causing DARPA (the American Defense Advanced Research Projects Agency) to fund several different AI-related projects from the mid-1960s onward, especially at MIT. This era was marked

[3]`https://www.intercom.com/`
[4]`https://www.drift.com/`

Evans [1964]'s Figure 1

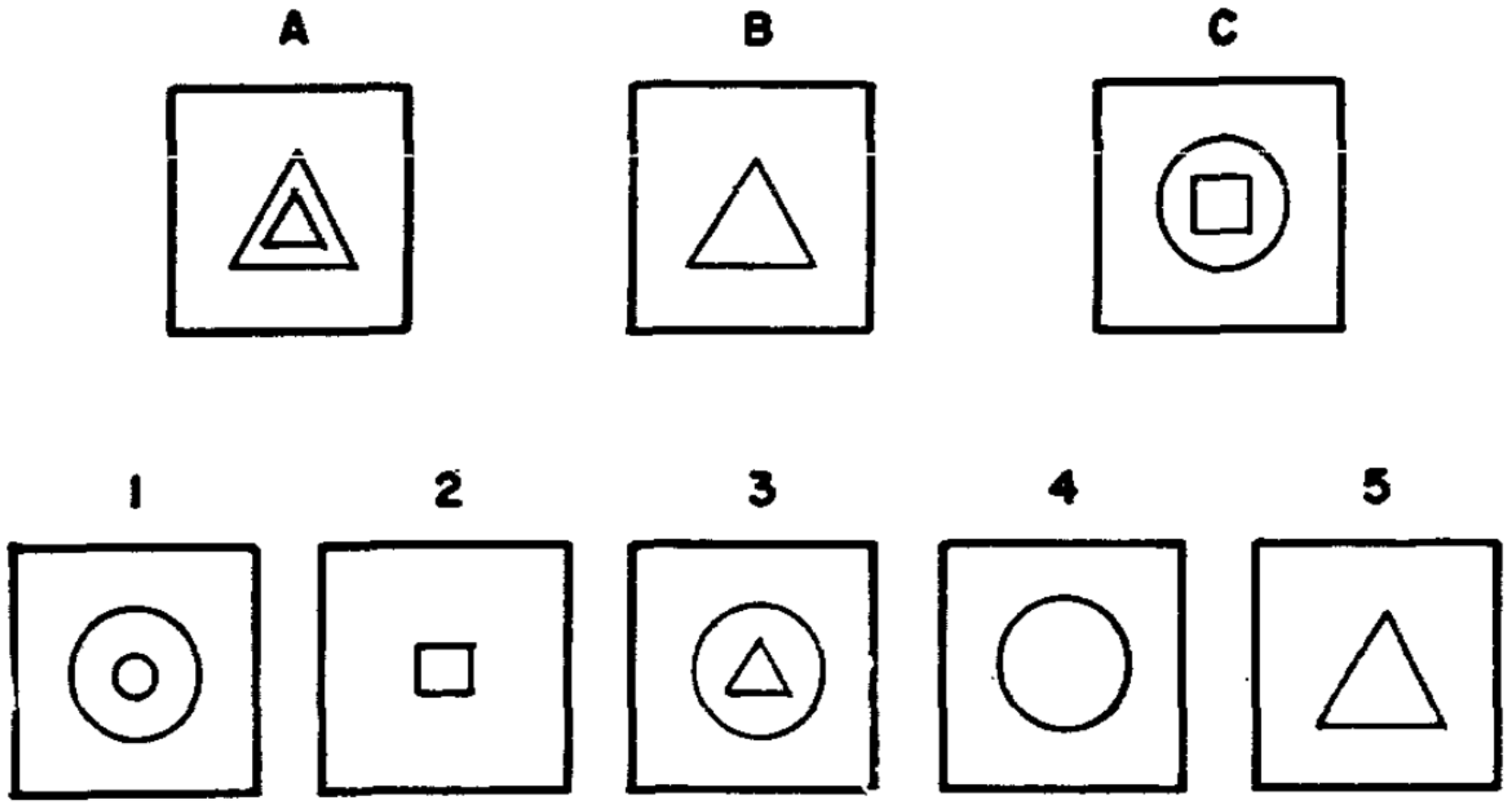

Figure 1: Geometry Analogy Problem example, which correlates with problems usually deployed in IQ tests where one needs to solve a question in the format "figure A is to figure B as figure C is to which of the given answer figures?"

by the extreme optimism in the speeches of the area practitioners. Marvin Minsky said in a 1970s Life magazine interview – one year after receiving the Turing Award (See Section 2.2) – that "from 3-8 years we will have a machine with the general intelligence of a human being". He would also, in the same interview, claim that "If were lucky, they might decide to keep us as pets.". Science Fiction fully adopted the Artificial Intelligence utopia, with the release of famous movies like the French "Alphaville" in 1965 by Jean-Luc Godard, and "2001: A Space Odyssey" by Stanley Kubrick (and screenplay from Arthur C. Clarke) in 1968.

Prior to its first Winter, however, AI had grown into a sizeable academic community. The First International Joint Conference on Artificial Intelligence (IJCAI), was then held at Stanford, in 1969. In it, out of the 63 published papers, some have been influential, such as Stanford's AI work in a "system capable of interesting perceptual-motor behavior" Feldman et al. [1969], Nilsson [1969]'s Mobius automation tool, and Green [1969]'s QA3 computer program that can write other computer programs and solve practical problems for a simple robot.

It was also before the first winter that Alain Colmerauer and Robert Kowalski would develop Prolog, now a prominent programming language, which was widely deployed in the 1970s and 1980s Kowalski [2014], Körner et al. [2022] that also influenced the field of inductive logic programming and statistical relational learning

Raedt et al. [2016]. The feature that makes Prolog stand out among other languages is the fact that it is mostly a declarative language: the program logic is expressed in terms of logic predicates (relations), represented as facts and rules. A computation is initiated by running a query over these relations Lloyd [1984]. More recently, Prolog would become a programming language also used in Watson[5], IBM's question-answering computer system.

### 2.1.2 The First AI Winter *(1974-1980)*

The first AI winter was defined by the hindrances found by researchers and practitioners while trying to develop deployable artificial intelligence technologies. The biggest challenge is today recognized as the lack of computing power needed by artificial intelligence algorithms, which simply did not exist at the time. Computers did not have enough memory to store the overwhelming amount of data required to build these complex rule-based systems or just did not have enough computational power to solve problems fast enough.

Minsky and Papert [1969] may have played a part in this process. Criticism of the "perceptron" (seen as a learning algorithm used in binary classifiers) possibly has influenced the Artificial Intelligence research agenda on going deeper into neural networks – highly influential now – and instead focusing on symbolic methods.[6] New results in NP-Completeness established by Cook [1971] and Karp [1972] in the early 1970s could have also played a role in raising skepticism, as results in computational complexity showed that many problems can only be solved in exponential time. This posed a risk to AI because it meant that some of the basic problems being solved by the era models would probably never be used in real-life data, where data is not represented by just a few data points. The well-known Lighthill report Lighthill [1973] also played a role, as it was interpreted as being critical of the state of AI research and its deployed results in the last 25 years leading to its publication in 1972.

### 2.1.3 The First AI Summer: Knowledge, Reasoning and Expert Systems *(1980-1987)*

The external perception of AI research results slowly recovered again in the early 1980s, mainly due to increasing commercial interest in *expert systems*. At that time, the symbolic school achieved higher prominence than other fields, and developments in non-monotics and temporal logic have had a lasting influence on the field. Symbolic and logical analyses of time, in particular, led to two Turing Award recognitions: Amir Pnueli (for introducing the methods of temporal logic in Computer Science) and Edmund Clarke, Allen Emerson, and Joseph Sifakis for their work on program verification and model checking, which are based upon the theoretical foundations of temporal logics. During the 1980s, we also witnessed the creation of the US-centered National Conference on Artificial Intelligence (AAAI), now a flagship international conference. Besides that, the funds that were somewhat reduced in the first AI winter would also be back on the table, with the Japanese government funding AI research as part of their Fifth Generation Computer Project (FGCP). Prolog had a central role in the fifth generation project and this led to several developments in computational logic at the time, as reported in Shapiro [1983].[7] Some other countries would also restart their funding projects, like UK's Alvey project and DARPA's Strategic Computing Initiative.

[5] `https://www.ibm.com/watson`

[6] Pollack [1989] reviews Minsky and Papert's *Perceptrons* and clarifies several issues around this influential book.

After the *Perceptrons* book criticism, connectionism would come back to prominence in the early 1980s. Hopfield [1982] proved that what we today call a "Hopfield network" could learn in a different way than what it was being done before with perceptrons and simple artificial neural networks. Also, at the same time, Rumelhart et al. [1986] and the "PDP research group" would show the potential of "Backpropagation", a new method to easily train and "backpropagate" the gradient in (neural) machine learning models.

### 2.1.4 The Second AI Winter *(1987-2000)*

Criticisms over the deployment of expert systems in real-world applications, however, may have caused the second AI winter (from the late 1980s to the early 2000s), which ended up reducing AI research funding.

Hans Moravec wrote in Moravec [1988] that "it is comparatively easy to make computers exhibit adult level performance on intelligence tests or playing checkers, and difficult or impossible to give them the skills of a one-year-old when it comes to perception and mobility". This, with some contributions from Rodney Brooks and Marvin Minsky, would emphasize what is now known as Moravec's Paradox: the idea that reasoning *per se* does not require much computation power, and can easily be thought/learned to/by an AI machine, but building an intelligent machine able to do what is "below conscience level for humans", i.e. motor or "soft" skills, is what

[7] Please note that a comprehensive history of symbolic AI and of the impact of logic in Artificial Intelligence is well beyond the scope of this paper. For a complete analyses of the many contributions of logic to AI and an understanding of the developments of logic-based AI methods, see Gabbay et al. [1998, 2014].

actually required enough computation power that did not yet exist at the time.

However, much happened in the 1990s as regards AI technology and its impacts. IBM success, represented by Campbell et al. [2002] Deep Blue's greatest achievement – finally beating in a match series under tournament rules the then world chess champion Garry Kasparov in 1997. Previously, in 1994, Tesauro [1994] TD-GAMMON program would illustrate the potential of reinforcement learning, by creating a self-teaching backgammon program able to play it at a master-like level. Also, although self-driving cars are usually considered recent technology, the ground for it was laid in this era, with Ernst Dickmmans's "dynamic vision" concept in Dickmanns [1988] and Thomanek and Dickmanns [1995] work where they had a manned car riding in a Paris' 3-lane highway with normal traffic at speeds of up to 130 km/h.

The late 1990s would also see an increase of research in information retrieval with the World Wide Web's boom, with research in web and AI-based information retrieval/extraction tools Freitag [2000].

### 2.1.5 Recent Advances in the XXI Century *(2000-present)*

The 2000s present us with wider public recognition of AI, especially if we look at the commercial impact of Machine Learning (ML), specifically Deep Learning (DL). In this context, NeurIPS (at the time, NIPS) arose, again, as perhaps the most prominent AI conference, where several influential DL papers have been published, featuring convolutional neural networks, graph neural networks, adversarial networks, and other (deep) connectionist architectures.

In the early 2000's we would see AI reaching a broader customer base in most developed countries. iRobot[8] introduced its Roomba Robot Vacuum in 2002. Apple, Google, Amazon, Microsoft, and Samsung released Siri, Google Assistant, Alexa, Cortana, and Bixby, respectively, AI-based personal assistants capable of better understanding natural language and executing a wider variety of tasks. Admittedly, they did not work that well at the beginning, circa 2010, but these technologies have improved over the last decade.

Most of the visibility in the area since 2000 is related to Deep Learning, basing itself in the Artificial Neural Network (ANN) concept, a system that tries to mimic how our brain cells work.[9] It is interesting to observe that already in 1943, in McCulloch and Pitts [1943], efforts were made to define artificial neural networks.[10]

[8] `https://www.irobot.com/`

[9] Of course, explaining the technical details of how artificial neural networks are deployed in machine learning, the many ANN models successfully developed over the last 40/50 years, and their technical complexities are beyond the scope of this work.

[10] Perhaps it is curious and relevant to observe that McCulloch and Pitts described how **propo-**

However, the immense computing power we have now allowed us to stack several layers of "neurons" one after the other – thus "deep" neural networks – and compute the results extremely fast. Also, given the natural parallelism of the process, the advent of Graphics Processing Units (GPUs) created the necessary hardware that led to the increasing number of deep models and their applications.

Some of the most noticeable recent achievements base themselves on Generative Adversarial Networks (GANs). They are a "framework for estimating generative models via an adversarial process, in which we simultaneously train two models: a generative model G that captures the data distribution, and a discriminative model D that estimates the probability that a sample came from the training data rather than G" Goodfellow et al. [2014]. This framework is responsible for a wave of photo-realistic procedurally-generated content at the likes of some viral websites, such as `https://this-person-does-not-exist.com/en`, `https://thiscatdoesnotexist.com/`, `http://thiscitydoesnotexist.com/`, or the recursive `https://thisaidoesnotexist.com/`.

GANs are associated with what one colloquially calls "deepfakes" – a mash of "deep learning" with "fake". They work by superimposing one's face with another face through a machine-learning model. Some more recent deepfakes can also alter the subject's voice, improving the experience. These are especially bad from an ethics standpoint when one imagines that these can be used to fake audio and images of influential people Hwang [2020]. A thorough review of the area can be found in Nguyen et al. [2019].

Several researchers have recently received the ACM Turing Award for work related to AI, (probabilistic) reasoning, causality, machine learning, and deep learning. In 2010 Leslie Valiant was awarded for his foundational articles dating back to the '80s and '90s, with some of his most prominent works associated with AI and Machine learning having defined foundational concepts in Computational Learning Theory (specifically, PAC Learning) Valiant [1984]. Judea Pearl won it in the following year, 2011, for his "contributions [...] through the development of a calculus for probabilistic and causal reasoning" with a representation of these contributions illustrated in Pearl [1988] and Pearl [2009]. Deep learning pioneers were recognized in 2018: Geoffrey Hinton, Yoshua Bengio, and Yann LeCun are widely recognized for their work on deep (belief) networks Hinton et al. [2006], image recognition Krizhevsky et al. [2012], text recognition and convolutional neural networks Lecun et al. [1998], LeCun et al. [1989], GANs Goodfellow et al. [2014] and neural translation Bahdanau et al. [2014] among many relevant contributions. However, it is essential to recognize

**sitional logic inference** could be described via neural networks. At that time, AI was not an established field, and thus no division among what came to be known as the connectionist and symbolic schools of AI.

that back in the 1980s, the PDP research group played a crucial role in showing the effectiveness of neural learning through backpropagation Rumelhart et al. [1985].

As regards the impact of Deep Learning reported in the media, in particular in the growing industry of entertainment and games, Google's AlphaGo won a series of matches against the Chinese Go grandmaster Ke Jie in 2017[11], after having already won 4 out of 5 matches against Go player Lee Sedol in 2016[12]. Also in 2017, OpenAI's Dota 2 bot[13] won a 1v1 demonstration match against the Ukrainian pro player Dendi, a noticeable demonstration of power in a game with imperfect information, with almost infinite possible future states. Later, in 2019, a new version of the same bot, called OpenAI Five, won back-to-back 5v5 games against the then-world-champion Dota team, OG[14]. Also in 2019 DeepMind's AlphaStar bot reaches the most significant possible tier in Starcraft II[15].

In recent years another sign of the impressive growth in AI research is the increasing number of submitted (and published) papers in the three biggest AI-related venues. Figure 6 shows that we have over 1500+ papers at these conferences in recent years. For exact numbers, please check Table 8. By checking the figure above, it is also important to notice how Computer Vision arguably became the most visible of the related areas, with CVPR having the biggest number of papers in their proceedings, thanks to the boom in applications for image recognition and self-driving cars. We give more details of AI-related publications in Section 4.

#### 2.1.6 Going Beyond Deep Learning: The Recent Debate on Robust and Neurosymbolic AI

The impacts of AI go beyond the results achieved by deep learning. Recently, the AI community witnessed a debate on how to build technologies that are explainable, interpretable, ethical, and trustworthy Rossi and Mattei [2019]. These led to increased attention to other fields that contribute to the challenge of constructing robust AI technologies. In particular, research that combines learning and reasoning in a principled way, for instance, neurosymbolic AI and hybrid models, have been the subject of growing research interest in academia and industry d'Avila Garcez et al. [2009], Marcus [2020], Riegel et al. [2020], Besold et al. [2022]. Further, in a recent *Communications of the ACM* article Hochreiter [2022], Sepp Hochreiter - deep learning pioneer and proponent of the Long Short-Term Memories (LSTM), one of the most deployed deep learning models - reflects upon a broader AI that "is a sophisticated and adaptive system, which successfully performs any cognitive task by virtue of its sensory perception, previous experience, and learned skill." Hochreiter states that graph neural networks (GNNs) can play a key role in building neurosymbolic AI technologies: "GNNs are a very promising research direction as they operate on graph structures, where nodes and edges are associated with labels and characteristics. GNNs are the predominant models of neural-symbolic computing Lamb et al. [2020]." Further, Hochreiter defends that "the most promising approach to a broad AI is a neuro-symbolic AI, that is, a bilateral AI that combines methods from symbolic and sub-symbolic AI" Hochreiter [2022]. He also states that contributions to neurosymbolic AI can come from Europe, which "has strong research groups in both symbolic and sub-symbolic AI, therefore has the unprecedented opportunity to make a fundamental contribution to the next level of AIa broad AI" Hochreiter [2022]. Although much is yet to be done and shown by AI researchers and professionals, it is clear that the field has grown in stature over the last decades. Also, testimony to the impact of AI is the prominence and growth in the areas of AI ethics, policies, and regulations, as well as annual AI global impact analyses made by several leading research organizations Mishra et al. [2020], Zhang et al. [2021].

[11] https://wired.com/2017/05/googles-alphago-continues-dominance-second-win-china/
[12] https://www.bbc.co.uk/news/technology-35797102
[13] https://openai.com/blog/dota-2/
[14] https://openai.com/five/
[15] https://www.ft.com/content/d659b056-fb28-11e9-a354-36acbbb0d9b6

## 2.2 The Association for Computing Machinery Alan M. Turing Award

The annual ACM A.M. Turing Award is regarded as the foremost recognition in computer science. It is bestowed by the Association for Computing Machinery (ACM) to people with outstanding and lasting contributions to computer science.

The award was introduced in 1966, named after the British Mathematician and Computer Science pioneer Alan M. Turing. Turing influenced several different branches of science, formalizing the concept of computation that led to the concept of a universal computing machine (today's "Turing Machines") through influential publications Turing [1936]. Turing is also considered by most as a modern AI pioneer after having designed the Turing test to decide if a machine is "intelligent" or not Turing [1950]. He is also known for his work in the Second World War, helping the British to decode the Nazi German Enigma machine with his *Bombe* machines, named after the Polish *bomba kryptologiczna* decoding machine. Since 2014, however, the winners receive US$1 million, financed by Google CACM [2014] for their exceptional achievement.[16]

The prize has been awarded to 62 researchers in diverse areas of computer science research - Table 11 lists every Turing Award winner and their nationalities. 37% of the winners were not born in the United States (some places of birth are not listed in the table) - and only 27%[17] of them credit a country other than the United States as the country where they did their main contribution. The first woman to receive the prize, Frances "Fran" Allen, received the prize for her work on the theory and practice of optimizing compiler techniques only in 2006.

[16] `https://amturing.acm.org`

For our AI evolution analyses, the relevant Turing Award Winners are those who had important contributions to this field. The Turing Award has recognized since 1966 seven researchers for their contributions to AI. The following information is provided by ACM at `https://amturing.acm.org/byyear.cfm`:

- **Marvin Minsky** (1969): *For his central role in creating, shaping, promoting, and advancing the field of Artificial Intelligence;*[18]
- **John McCarthy** (1971): *Dr. McCarthy's lecture "The Present State of Research on Artificial Intelligence" is a topic that covers the area in which he has achieved considerable recognition for his work;*[19]
- **Herbert Simon** and **Allen Newell** (1975): *In joint scientific efforts extending over twenty years, initially in collaboration with J. C. Shaw at the RAND Corporation, and subsequentially with numerous faculty and student collegues at Carnegie-Mellon University, they made basic contributions to artificial intelligence, the psychology of human cognition, and list processing;*[20][21].
- **Edward Feigenbaum** and **Raj Reddy** (1994): *For pioneering the design and construction of large-scale artificial intelligence systems, demonstrating the practical importance and potential commercial impact of artificial intelligence technology;*[22][23]
- **Leslie Valiant** (2010): *For transformative contributions to the theory of computation, including the theory of probably approximately correct (PAC) learning, the complexity of enumeration and of algebraic computation, and the theory of parallel and distributed computing;*[24]

[17]See every author page in their ACM Turing Award website: `https://amturing.acm.org/byyear.cfm`

[18]`https://amturing.acm.org/award_winners/minsky_7440781.cfm`

[19]`https://amturing.acm.org/award_winners/mccarthy_1118322.cfm`

[20]`https://amturing.acm.org/award_winners/simon_1031467.cfm`

[21]`https://amturing.acm.org/award_winners/newell_3167755.cfm`

[22]`https://amturing.acm.org/award_winners/feigenbaum_4167235.cfm`

[23]`https://amturing.acm.org/award_winners/reddy_9634208.cfm`

[24]`https://amturing.acm.org/award_winners/valiant_2612174.cfm`

- **Judea Pearl** (2011): *For fundamental contributions to artificial intelligence through the development of a calculus for probabilistic and causal reasoning;*[25]
- **Geoffrey Hinton**, **Yann LeCun**, and **Yoshua Bengio** (2018): *For conceptual and engineering breakthroughs that have made deep neural networks a critical component of computing.*[26][27][28]

## 2.3 Computer Science and AI Conferences

Today, there are thousands of conferences in Computer Science. In our analyses, we obviously had to narrow them down to the ones considered the flagship venues in order to properly analyse the field at its core. CSRankings is a metrics-based ranking of top computer science research institutions[29], which identifies the works from each institution for each venue. They comprise a small set of conferences considered as the top ones in the subfields of computing: In this work, we will only focus on institutions available in the CSRankings "AI" category, which are briefly described below. Some conferences are acronyms borrowed from namesake scientific associations - such as AAAI.

### 2.3.1 IJCAI

The International Joint Conferences on Artificial Intelligence (IJCAI) was first held in California in 1969, being the first comprehensive AI-related conference. The conference was held in odd-numbered years, but since 2016 it has happened annually. It has already been held in 15 different countries, while the 2 most COVID-19 pandemic years were virtually held in Japan and Canada. The next editions will be held in Austria (2022), South Africa (2023), and China (2024), increasing the number of countries that hosted the conference to 17 – China has already hosted it before. Similarly to AAAI, IJCAI is a comprehensive AI conference, with some publications ranging from the philosophy of AI to symbolic AI, and machine learning and applications. IJCAI has over the years published important papers from Turing Award winners, e.g. Avin et al. [2005], Feigenbaum [1977], McCarthy [1977], Valiant [1985], Verma et al. [2019].

[25] `https://amturing.acm.org/award_winners/pearl_2658896.cfm`
[26] `https://amturing.acm.org/award_winners/hinton_4791679.cfm`
[27] `https://amturing.acm.org/award_winners/lecun_6017366.cfm`
[28] `https://amturing.acm.org/award_winners/bengio_3406375.cfm`
[29] `http://csrankings.org`

### 2.3.2 AAAI

The Association for the Advancement of Artificial Intelligence (AAAI – pronounced *"Triple A I"*) was founded in 1979 as the American Association for Artificial Intelligence. This association is responsible for promoting some prominent AI conferences since 1980, including The AAAI Conference on Artificial Intelligence (formerly the National Conference on AI). The conference used to be held once every one or two years (depending on whether IJCAI was organized in North America or not). AAAI has been held yearly since 2010. It is worthy of note that although the conference was renamed, it has actually only been held in North America (and remotely in 2021 due to the COVID-19 pandemic). The conference covers AI comprehensively as its (older) sister conference IJCAI. Similarly to IJCAI, AAAI has published several papers from influential researchers and Turing laureates, see e.g. Hinton [2000], de Mori et al. [1988], Pearl [1982], Valiant [2006].

### 2.3.3 NeurIPS (formerly NIPS)

The Conference and Workshop on Neural Information Processing Systems (NeurIPS) is a machine learning and computational neuroscience conference held every December since 1987. It has been held in the USA, Canada, and Spain. NeurIPS published hundreds of influential papers over the years on several learning models and architectures ranging from Long Short-Term Memories Hochreiter and Schmidhuber [1996], to Transformer architectures Vaswani et al. [2017]. The Conference Board decided to change the meeting name to NeurIPS [30] in 2018.

CSRankings defines it as a "Machine Learning & Data Mining" conference, publishing important papers, recently featuring technologies such as GPT-3 Brown et al. [2020] and PyTorch's technical papers Paszke et al. [2019], which have 31 and 21 authors, respectively. The sheer size of the venue is noticeable, with 2,334 papers accepted in 2021, outnumbering every other conference studied in this work. The "most influential papers in the recent years" snippet, is due to `https://www.paperdigest.org/`.

### 2.3.4 CVPR

The Conference on Computer Vision and Pattern Recognition (CVPR) is an annual conference on Computer Vision and Pattern Recognition, regarded as one of the most important conferences in its field, with 1,294 accepted papers in 2019. It will in 2023, for the first time, be organized outside the United States in Vancouver,

[30] `https://www.nature.com/articles/d41586-018-07476-w`

Canada. It was first held in 1983 and has since 1985 been sponsored by IEEE, and since 2012 by the Computer Vision Foundation, responsible for providing open access to every paper published in the conference. CVPR is a flagship Computer Vision venue and witnessed groundbreaking work in the past including Siamese Representation Learning Chen and He [2020], GANs Karras et al. [2018], and Dual Attention Networks Fu et al. [2018]. Turing award laureate Yann LeCun published in this conference on several occasions, e.g. Boureau et al. [2010], LeCun et al. [2004].

### 2.3.5 ECCV

ECCV stands for European Conference on Computer Vision, being CVPR's European arm – even though ECCV 2022 is going to be held in Tel Aviv - Israel. It is held biennially on even-numbered years since 1990, when it was held in Antibes, France. Even though it is considered CVPR's small sister, it had 1,360 accepted papers in 2019, also heavily focusing on Computer Vision with some publications of note such as RAFT Teed and Deng [2020], a model able to segment and predict image depth with high accuracy.

### 2.3.6 ICCV

Similar to ECCV, the International Conference on Computer Vision is CVPR's International arm, being held every odd-numbered year since 1987, when it was held in London, United Kingdom, been held in 14 other countries ever since. 1,077 papers made the cut in 2019, such as Shaham et al. [2019] who won the 2019's best paper award.

### 2.3.7 ACL

ACL is the Association for Computational Linguistics's conference held yearly since 2002, having surprisingly been held in 15 different countries in the last 20 years. The website announces ACL as "the premier international scientific and professional society for people working on computational problems involving human language, a field often referred to as either computational linguistics or natural language processing (NLP). The association was founded in 1962, originally named the Association for Machine Translation and Computational Linguistics (AMTCL), and became the ACL in 1968." Commonly referred to as an NLP-related conference, it has some highly-cited work in recent years such as Strubell et al. [2019]'s work in investigating the environmental effects of creating large language models.

#### 2.3.8 NAACL

NAACL is the conference held by the North American Chapter of the Association for Computational Linguistics, therefore also referred to as an NLP conference. The conference is actually named NAACL-HLT (or HLT-NAACL, sometimes) – North American Chapter of the Association for Computational Linguistics: Human Language Technologies. It has been held since 2003, and it was co-located with ACL on the occasions ACL happened in North America. One of the most cited papers on the use of transformers in recent years was published there, the BERT model Devlin et al. [2018].

#### 2.3.9 EMNLP

EMNLP stands for Empirical Methods in Natural Language Processing. The conference started in 1996 in the US based on an earlier conference series called Workshop on Very Large Corpora (WVLC) and has been held yearly since then. The recent conferences are marked by works trying to improve the BERT model Devlin et al. [2018] already explained above, such as Jiao et al. [2019], Feng et al. [2020] and Beltagy et al. [2019].

#### 2.3.10 ICML

ICML is the International Conference on Machine Learning, a leading international academic conference focused on machine learning. The conference is held yearly since 1987, with the first one held in 1980 in Pittsburg, USA. The first conferences were all held in the United States, but the 9th conference, in 1992, was held in Aberdeen, Scotland. Since then it has been held in 10 other countries, and twice virtually because of the COVID-19 pandemic. It contains some seminal papers in Machine Learning from Pascanu et al. and Ioffe and Szegedy, and some more recent excellent research like Zhang et al. [2018] and Chen et al. [2020]. Besides Bengio's aforementioned seminal paper, Turing Award laureate Geoffrey Hinton also published important papers on ICML e.g. Nair and Hinton [2010].

#### 2.3.11 KDD

The SIGKDD Conference on Knowledge Discovery and Data Mining is an annual conference hosted by ACM, which had its first conference in 1989's Detroit. Although it is usually held in the United States, it has already been hosted by a few other countries, namely Canada, China, France, and the United Kingdom. It is the most important conference encompassing the Knowledge Discovery and Data Mining field, with 394 accepted papers in 2021: a smaller number if we compare with the other conferences we investigate in this paper. The conference recent years have seen a significant presence of AI-related research, mostly defined by Graph Neural Networks (GNNs), with works from Qiu et al., Wu et al., Jin et al., and Liu et al., all of them accepted in 2020's SIGKDD. It is interesting to note how many of the authors of such influential papers in the 2020 conference are from China; perhaps a trend – see some insights about it in Section 4.6.

#### 2.3.12 SIGIR

SIGIR stands for Special Interest Group on Information Retrieval, an ACM group. It has its own annual conference that started in 1978 and has happened every single year since then. It is considered to be the most important conference in the Information Retrieval (how to acquire useful and organized information from raw, unorganized, and unstructured data) area. After 43 editions, it has been hosted in 21 different countries. It used to alternate between the USA and a different country, but this rule does not hold anymore, with only one conference in the US in the last 8 years. Many papers in recent years have tackled recommender systems such as He et al., Wu et al., and Wang et al..

#### 2.3.13 WWW

The Web Conference (WWW) is the top conference in the field. It "brings together key researchers, innovators, decision-makers, technologists, businesses, and standards bodies working to shape the Web"[31]. It is a yearly event that started at CERN in Geneva, Switzerland, in 1994. The conference heavily focuses on Semantic web and Data mining with some important results in recommender systems as well.

### 2.4 A Brief on Graphs and Centrality Measures

Next, we briefly introduce the concepts of graph theory used in this paper. Sections 2.4.2 to 2.4.5 describe the most widely used graph centralities from the literature. Then in Section 2.4.6 we go over some other centrality measures for completeness' sake.

A graph $G$ is represented by a tuple $G = (V, E)$, where V is a set of nodes (vertexes) and E a set of edges $e_{u,v}$ connecting nodes $u$ to $v$ where $u, v \in V$. These edges can be directed or undirected – thus making us able to differentiate between directed and undirected graphs. In the directed case of $e_{u,v}$ we call $u$ as being the source node and $v$ the destination node. We will always use $n$ to represent the number of nodes in a graph, and $m$ to represent the number of edges in it. Also, a pair of nodes $(u, v)$ might have more than one edge connecting them: in this case, we call the graph a multigraph. Similarly, these edges might have a weight $w$ making the graph a weighted graph. Furthermore, we can also have labeled graphs, where nodes and edges can be of different types. These are useful in knowledge representation systems, such as the graph built in Section 3.3.4. We call $p = u_1, u_2, ..., u_p$ a path between $u_1$ and $u_p$ in $G$ if $\exists\, e_{u_i,u_i+1}\, \forall\, 1 <= i <= p-1$. Basically, we have a path if we can go from node $u_1$ to $u_p$ through a sequence of connected edges. We can also define the shortest path between a pair of nodes $(u, v)$ as the path with the minimum possible quantity of intermediate nodes – note, however, that we can have more than one shortest path between any pair of nodes $(u, v)$.

[31] `https://dl.acm.org/conference/www`

### 2.4.1 Centrality Measures

The interest in Graph centrality measures dates back to the 1940s, but it was more formally incorporated into graph theory in the 1970s Freeman [1978]. A fundamental motivation for the study of centrality is the belief (or relevance) that node position (that can represent a person's position) in a network impacts their access to information, status, power, prestige, and influence Grando et al. [2019], Wan et al. [2021]. Therefore, throughout this work when we want to identify the above concepts we will use graph centralities for the different networks we built. Grando et al. serves as a tutorial and survey for those interested in applying machine learning to this field.

### 2.4.2 Degree Centrality

We represent the degree of a node $u$ as $k_u$ meaning the number of other nodes connected to this node. In a directed graph we can further split this metric into two: $k_u^{in}$ is the in-degree, representing the number of nodes $v \in V$ that have an edge $e_{v,u}$ with $v$ as source and $u$ as destination (i.e. number of nodes with an edge pointing to $u$); the opposite metric $k_u^{out}$ is the out-degree, representing the number of nodes $v \in V$ that have an edge $E_{u,v}$. Therefore, it is possible to extend this metric to a centrality called **Degree Centrality** defined as:

$$\mathcal{C}_{Dg}(u) = \frac{k_u}{n-1}, \tag{1}$$

where $n$ represents the number of nodes $V$ in the graph $G$.

Also, in the same way we have in-degree and out-degree metrics, we can extend Equation 1 and define **In-Degree Centrality** and **Out-Degree Centrality**, respectively:

$$\mathcal{C}_{Dg_{in}}(u) = \frac{k_u^{in}}{n-1} \tag{2}$$

$$\mathcal{C}_{Dg_{out}}(u) = \frac{k_u^{out}}{n-1} \tag{3}$$

These degree metrics are used to identify how well a node is directly connected to other nodes, without considering the influence a node can pass to its neighbors.

### 2.4.3 Betweenness Centrality

The **Betweenness Centrality** was defined in Freeman [1977], and its measure of the importance of a node $u$ is how many shortest paths in the graph go through $u$. It is defined as

$$\mathcal{C}_B(u) = \frac{\sum_{s \neq u \neq t} \frac{\partial_{s,t}(u)}{\partial_{s,t}}}{(n-1)(n-2)/2} \qquad \forall s, u, t \in V, \, \exists \, e_{s,t} \tag{4}$$

where $\partial_{s,t}(u)$ is the number of shortest paths between $s$ and $t$ that go through $u$, and $\partial_{s,t}$ is simply the number of shortest paths between $s$ and $t$. Note that we are only ever counting paths between the pair $(s, t)$ if there is a path between $(s, t)$.

Betweenness is related to the notion of connectivity, where a node with a bigger betweenness actually means that it is a point of connection between several nodes. In a graph with a single connected component, a node can have the highest betweenness if it works as a bridge between two individually disconnected components. It is regarded as a measure of a nodes control over communication flow Freeman [1978].

### 2.4.4 Closeness Centrality

**Closeness Centrality** was created in Sabidussi [1966] representing the closeness of a node with every other node in the graph. It is the inverse of the farness which in turn is the sum of distances with all other nodes Saxena and Iyengar [2020]. It is defined by

$$\mathcal{C}_C(u) = \frac{n-1}{\sum_{v \neq u} d(u,v)} \qquad \forall u, v \in V \tag{5}$$

where $d(u,v)$ is the distance between the nodes $u$ and $v$. This distance is simply the number of edges in the shortest path $p$ between the pair $(u,v)$ if the graph is unweighted, while it is the sum of every edge in the path in case the graph is unweighted. Since distance is not defined between every pair of nodes in disconnected graphs (a graph where not every node can be reached from another node) we cannot compute closeness for disconnected graphs. A node with a higher closeness indicates that the node is in the middle of a hub of other nodes. It also means that a node with big closeness values is "closer", on average, to the other nodes, hence closeness. It represents the nodes level of communication independence Freeman [1978], Cardente [2012].

### 2.4.5 PageRank Centrailty

Pagerank is a global centrality measure that needs the overall network to measure the importance of one node. It measures the importance of one node based on the importance of its neighbors. Saxena and Iyengar [2020]. It was developed by Brin and Page when they created Google, and it is the underlying method behind their search engine. To understand Pagerank, we need to understand that its main idea is to understand how important a web page is in the middle of all the other millions of pages on the World Wide Web. The main idea behind it is that we are considering a web page important if other important web pages link to it.

Think about it as if we had a web crawler randomly exploring the web and increasing a counter every time we enter into a specific page. Then, when you are on a page you either have the option to click on one of the links on the page or go to a random page on the web with probability $0 <= q <= 1$ – this is useful both to model real-life where we simply go to random websites and also to mimic pages without any out link. The usual value for $q$, also called teleportation or damping factor, is 0.15, as defined in the original paper. Therefore, with this in mind, we can define **Pagerank** as

$$\mathcal{C}_{PR}(u) = \frac{q}{n} + (1-q)\sum_{v} \frac{\mathcal{C}_{PR}(v)}{k_v^{out}} \qquad \exists e_{u,v} \in E \tag{6}$$

The equation above illustrates how this process is iterative because we depend on the Pagerank of every neighbor to be able to compute our own Pagerank. The process usually converges or can be stopped after a certain number of iterations.

### 2.4.6 Other centralities

There are other useful centralities present in the literature. They were not used in our work, but they would ideally be used in future work using the dataset created. Recent work has focused not only on the application of machine learning in learning centrality measures of complex graphs Grando et al. [2019], but also on analyzing the own application of Graph Neural Networks capable of multitask learning trained on the relational problem of estimating network centrality measures Avelar et al. [2019]. In summary, the reader interested on centrality measures can refer to Grando et al. [2019], Saxena and Iyengar [2020].

- **Semi-Local centrality** Chen et al. [2012] defines a metric similar to the degree centrality where we expand it to 2 levels of neighbours.

$$\mathcal{C}_{SL}(u) = \sum_{v \in N(u)} \sum_{w \in N(v)} d_2(w), \tag{7}$$

  where $\mathrm{d}_2(w)$ is the number of neighbors plus the number of neighbors for every neighbor of $w$ – basically how many nodes you can reach in two steps.

- **Volume Centrality** Wehmuth and Ziviani [2013] is a kind of generalization from the above centrality parameterizing how far a node influence can reach and is defined by

$$\mathcal{C}_V(u) = \sum_{v \in \tilde{N}_h(u)} k_v, \tag{8}$$

  where $N_h(u)$ is the set of neighbors within a distance $h$ of $u$, and $\tilde{N}_h(u) = N_h(u) \cup \{u\}$. Wehmuth and Ziviani [2013] demonstrated that $h = 2$ results in a good trade-off of identifying nodes with important relations and the cost of computing this relationship.

- **H-index** Hirsch [2005] is a well-known statistic in the research world, being exhibited as a statistic in most research-aggregator portals such as Google Scholar and DBLP. Hirsch [2005] defined that $h$ is the highest integer value for which the author has $h$ papers with at least $h$ citations.

- **Coreness Centrality** Kitsak et al. [2010] represents the idea that the important nodes are at the core of a graph. It can be determined by the process of assigning each node an index (or a positive integer) value derived from the $k$-shell decomposition. The decomposition and assignment are as follows: Nodes with degree $k$= 1 are successively removed from the network until all remaining nodes have a degree strictly greater than 1. All the removed nodes at this stage are assigned to be part of the $k$shell of the network with index $k_S$= 1 or the 1-shell. This is repeated with the increment of $k$to assign each node to distinct $k$-shellsWan et al. [2021]. See Figure 2 to see an example of the definition of $k$-shells. Then, we can mathematically define this centrality as

$$\mathcal{C}_k(u) = \max\{k|u \in H_k \subset G\}, \tag{9}$$

where $H_k$ is the maximal subgraph of $G$ with all nodes having a degree of at least $k$ in $H$Wan et al. [2021].

Source: Tanase et al. [2015]

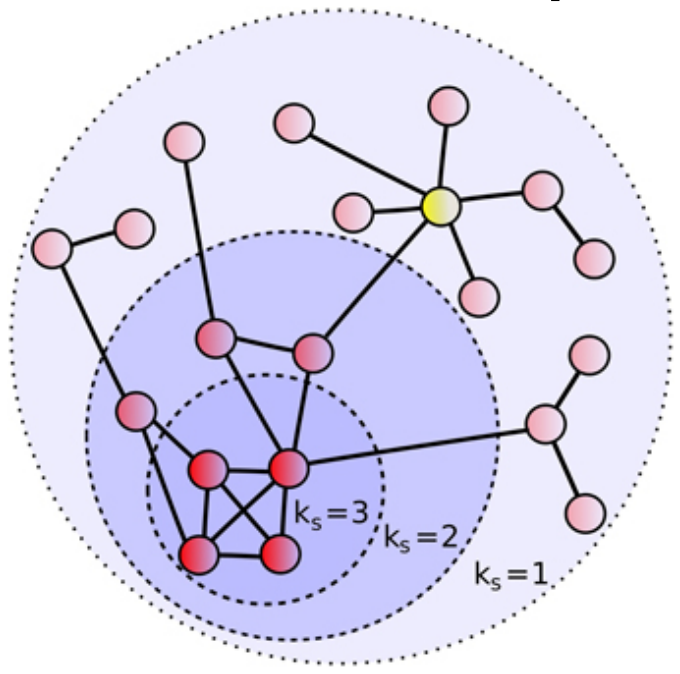

Figure 2: Example of $k$-shell assignment

Some more complex centralities mostly use the fact that we can define a graph by its adjacency matrix $\mathbf{A}$ and its corresponding eigenvalues and eigenvectors. Since we are not carrying out a comprehensive review or using them, we will not describe them.

## 2.5 Related Work

One of the main reasons motivating this work is the fact that the history of Artificial Intelligence and its dynamic evolution has not been researched in depth, at least with respect to our methodology: Xu et al. [2019] focused specifically on "explainable AI" evolution (or de-evolution, in this case); Oke [2008] does deepen their work in several different AI areas, with a review of each area, but does not go back in history further than the mid-1990s. There are also similar approaches to investigating author citation/collaboration networks such as Ding et al. [2010], Guns et al. [2011], Abbasi et al. [2012], and Wu et al. [2019], mostly focusing on the betweenness centrality. von Wartburg et al. [2005] uses closeness to analyse patent networks. Also, Krebs [2002] shows how centrality measures can be used to identify prominent actors from the 2001 Twin Tower terrorist attackers network.

Regarding the authors' country affiliation in papers, Grubbs et al. [2019] investigated coauthor country affiliation in Health research funded by the US National Institute of Health; Michalska-Smith et al. [2014] goes further by trying to correlate country of affiliation with the acceptance rate in journals and conferences; Yu et al. [2021] studied how one can infer the country of affiliation of a paper from its data in WoS[32]; Hottenrott et al. [2019] investigates the rise on multi-country affiliations in articles as well.

# 3 Methodology

## 3.1 Underlying Dataset

The most extensive public bibliography of computer science publications is probably the DBLP Database DBLP [2019], available at `https://dblp.uni-trier.de/`. Recently (in February 2022), it surpassed the 6 million publications mark (See Figure 3), containing works from almost 3 million different authors. Figure 4 shows how large is the increase in publications in the recent years, per DBLP's statistics page[33]. They provide a downloadable 664MB GZipped version of their dataset in XML format[34]. Recently (after this work had already been started and was past the dataset-choosing process), DBLP has also released its dataset in RDF format[35]. However, because their dataset contains duplicated authors and/or incorrectly merged authors, we opted to not use their dataset directly. Instead, in our work, we used Arnet's Tang et al. [2008] V11[36] paper citation network, which dates from May 2019. It contains 4,107,340 papers from DBLP, ACM, MAG (Microsoft Academic Graph), and other sources, including 36,624,464 citation relationships.

[32] `https://www.webofknowledge.com/`
[33] `https://dblp.org/statistics/index.html`
[34] `https://dblp.org/xml/release/`
[35] `https://blog.dblp.org/2022/03/02/dblp-in-rdf/`
[36] `https://lfs.aminer.cn/misc/dblp.v11.zip`

Source: `https://blog.dblp.org/2022/02/22/6-million-publications/`

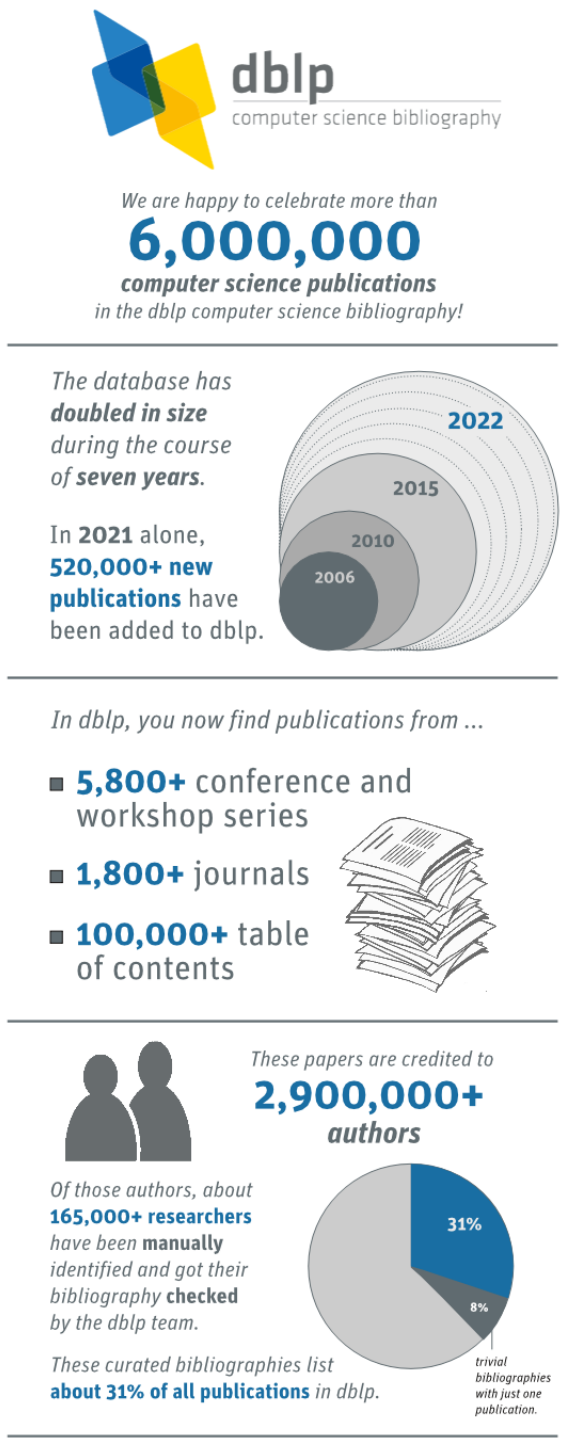

Figure 3: Excerpt of a DBLP poster acknowledging their 6 million papers mark

This dataset contains more information than DBLP's, as they better worked on author disambiguation (merging authors DBLP considered to be different ones, or separating authors DBLP considered to be the same person), providing us the ability to generate truther collaboration/citation networks.

It is important to clarify why we are using Arnet's v11 dataset instead of one of their newer datasets, namely v12 and v13 – the latter, from May 2021, contains 5,354,309 papers and 48,227,950 citation relationships, an increase of 30.3% compared to v11. First, and foremost, this work started in 2019, when versions v12 and v13 were not available yet. Also, when these newer datasets were made available, we did try to use both of them, but we faced some problems that prompted us back to the v11 dataset:

1. v12's and v13's data format is different from v11's. The format of v12 and v13 is a fully-fledged 11GB XML file, which required us to write a new Python library to convert from XML to JSON (our storage method) without loading the whole file into memory by streaming-converting it (see Section H.1). Be-

Source: `https://dblp.org/statistics/publicationsperyear.html`

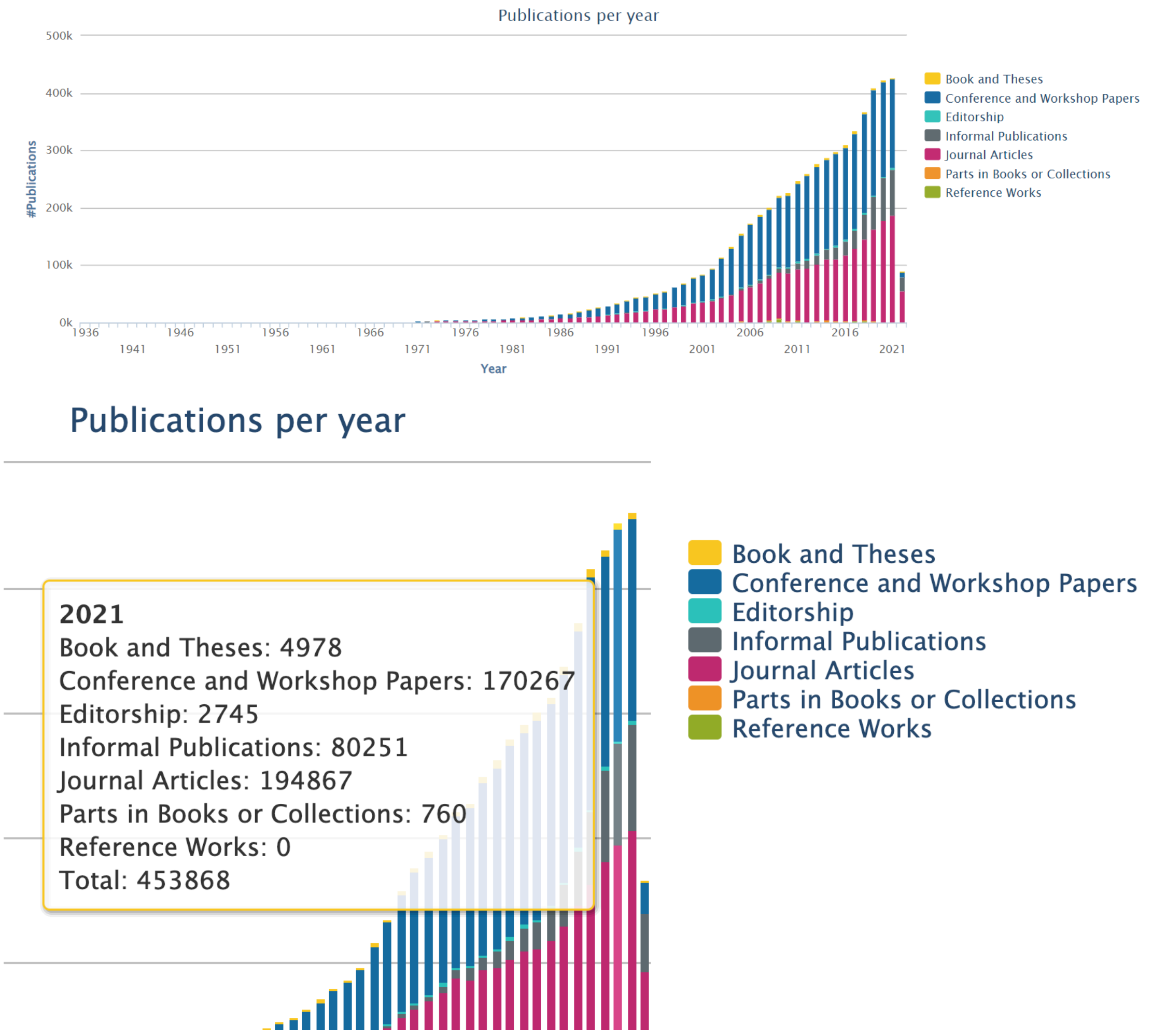

Figure 4: DBLP papers per year, with a detailed view of 2021.

sides the file being harder to read and handle, the new format also changed the IDs from an integer to a UUID-based value, causing us to rewrite the whole logic that was able to detect papers from the main AI conferences based on their past integer values.

2. There are fewer papers from the AI conferences of interest for this work. Even though we have 30% more papers in the most recent version, after carefully finding out which are the new IDs for the conferences, we could only find 58490 papers out of the 89102 ( 65%) present on version v11. As a smoke test, we did reduce our test only for the main AI conferences (AAAI, NeurIPS, and IJCAI): we could manually count 42082 papers in these 3 conferences – and this is a lower bound because we could not find the count of papers in some years for AAAI and IJCAI; v11 and v13 have 41414 and 20371 of them, respectively. We also tried finding the AI Conferences by name instead of IDs (at cost of some false positives) but it did not work, also finding only 20929 papers. This shows how we have twice the data in v11 compared to v13 instead of 30% more in v13 as expected.

3. Missing data in the most recent years. Even though v13 should have data until 2021, there are only a few hundred papers for the main AI conferences in 2019, 2020, and 2021, while in reality there should be 12559 of them.

All of the data compiled to build the points above can be seen in Table 1, and Figure 8. Table 8 has the raw data used to build Figure 8, where "?" data points were considered to be 0 for the sake of simplicity. An interesting statistical information one might get from Figure 5 is the fact that even though IJCAI used to happen only in odd-numbered years, even-numbered years do not have any noticeable NeurIPS and AAAI paper acceptance rates increase.

Section 4.6.1 shows some charts where it can be seen how degraded our data looks if we had used *v13* instead of *v11*.

| | **AI Conferences Total** |
|---|---|
| *Manual Count* | 42082 |
| *v11* | 41414 |
| *v13 detecting conferences by ID* | 20371 |
| *v13 detecting conferences by name* | 20929 |

Table 1: Comparison of paper counts with different methods

Arnet's v11 format is a variation of a JSON file with some improvements to make it easier to load the data in memory without having to load the whole file. Every line is a valid JSON object, requiring us to simply stream the file, iterating over every line, parsing the JSON file, keeping only the required information in memory, and immediately send the JSON file to be garbage collected, using no more than 8kb of memory to read the entire file.

Every JSON object in this file follows the structure defined in Table 2. We, then, for most of the work, keep only the fields tagged with an asterisk ($*$). Also, a question mark symbol (?) indicates the field is optional and is, sometimes, not present in the data provided by Arnet. Figure 43 shows an example of such JSON entry, depicting Glorot and Bengio [2010]'s representation in the dataset.

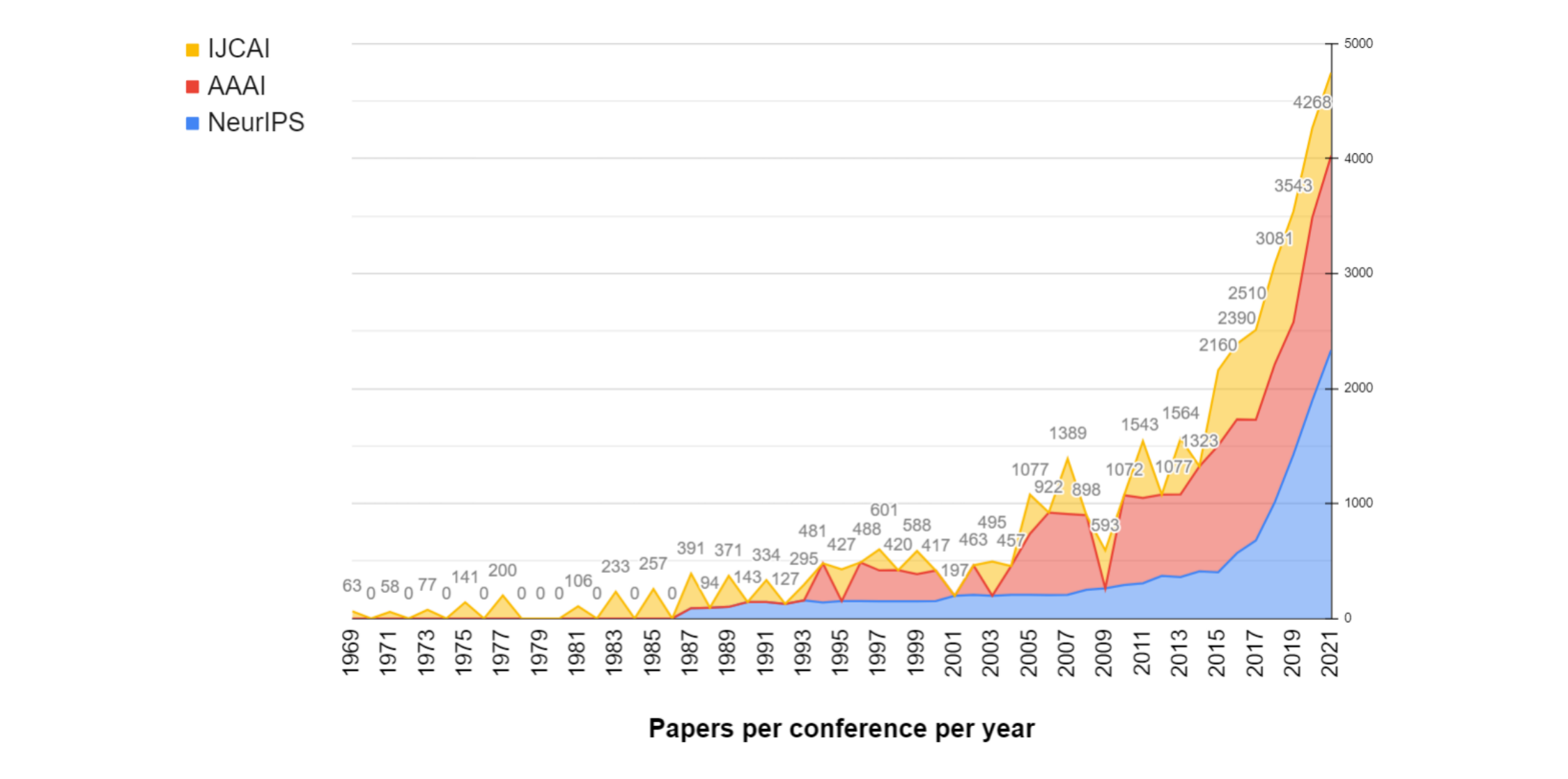

Figure 5: Manual paper count per year in AAAI, NeurIPS and IJCAI

*"*" indicates the field was used in this work*
*"?" indicates the field is optional*

| **Field Name** | **Type** | **Description** |
|---|---|---|
| id* | *string* | Unique identifier for the paper |
| title* | *string* | Paper title |
| authors* | *Author[]* (See Table 5) | List of every single author |
| venue* | *Venue* (See Table 6) | Object with data about the venue |
| year* | *integer* | Year of publication |
| n_citation | *integer* | Citation number |
| page_start? | *string* | Paper start page in the Proceedings/Book/Journal |
| page_end? | *string* | Paper end page in the Proceedings/Book/Journal |
| doc_type | *string* | Place of publication |
| publisher? | *string* | Book/Journal publisher |
| volume? | *string* | Book volume |
| issue? | *string* | Journal issue |
| references* | *string[]* | List of ids this paper references |
| indexed_abstract* | *IndexedAbstract* (See Table 7) | Inverted index holding data about the paper abstract |

Table 2: Data structure for a single entry in the Arnet JSON dataset

Figure 6 shows some raw insights about this dataset, using the conferences de-fined in Section 2.3. It shows that all conferences have seen an increasing trend in the number of papers in the last few years, especially CVPR and AAAI.

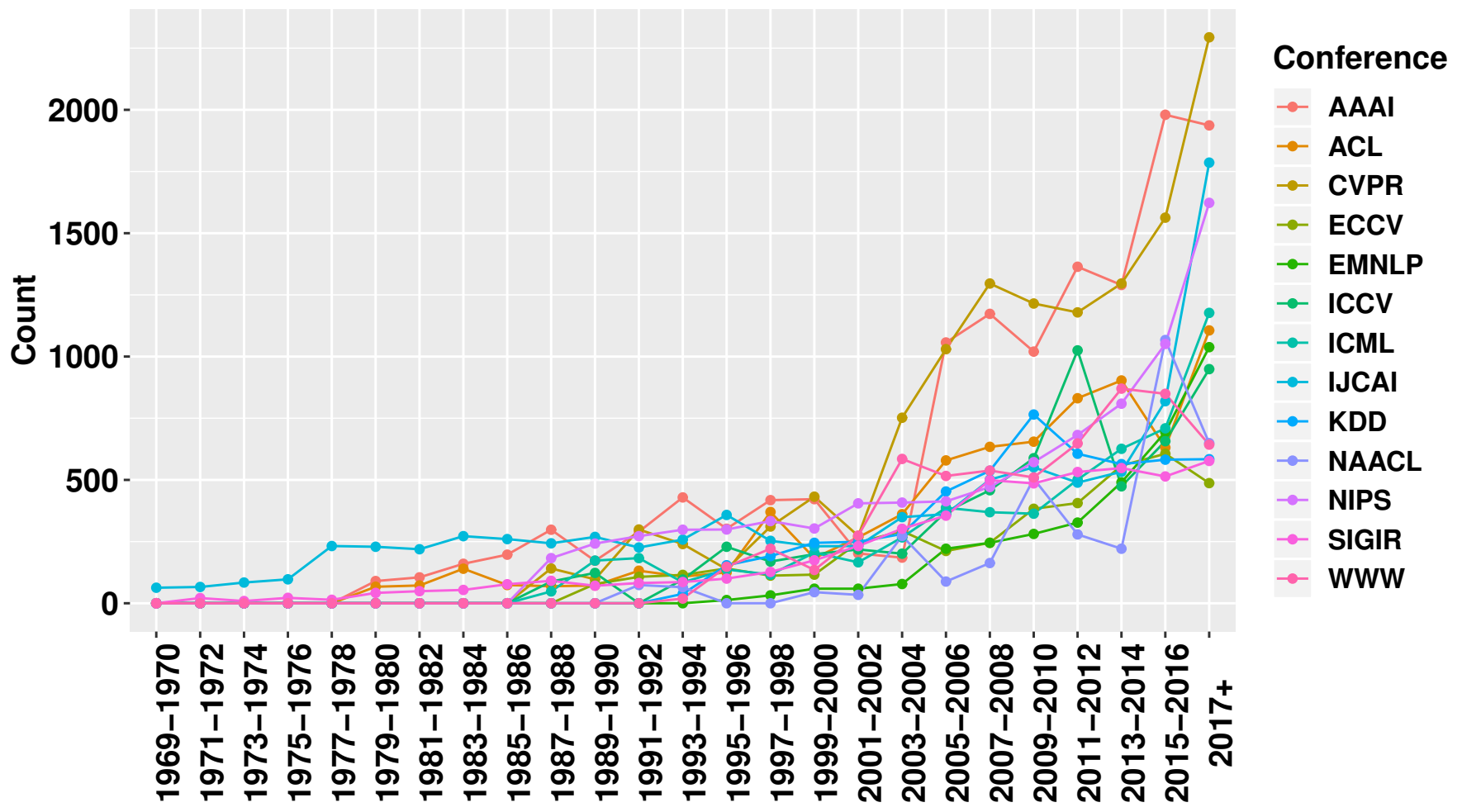

Figure 6: Number of papers per conference per year.

### 3.2 Artifacts

The code used to download the data, parse the dataset, and generate the graphs, analyses, and charts present in this work is available at `https://github.com/rafaeelaudibert/TCC/tree/v11` in Github. The code for this work is in branch *v11*. The *master* branch contains the code used when we were trying to parse Arnet's *v13* dataset, which did not work out as explained in the previous section. Everything data analysis was built using Python, with the help of some open-source third-party libraries (See Table 9) available in PyPi. For the most complex plots, Python was not the right tool for the job, so they were built using R and its built-in counterparts for *matplotlib*, *numpy* and *seaborn*. Unfortunately, the code for these graphs is not available anymore because it was lost during a disk formatting procedure.

#### 3.2.1 Graph Datasets

Throughout this work, we assembled 5 new datasets, modeled in a graph structure, which are briefly described below. A thorough explanation can be found in each respective section below.

**Author Citation Graph (*ACi*)** Directed multigraph, where every author is a node, with edges representing citations.

**Author Collaboration Graph (*ACo*)** Undirected graph, where every author is a node, with edges representing co-authorship

**Paper Citation Graph (*PC*)** Directed graph, where every paper is a node, with edges presenting citations.

**Author-Paper Citation Graph (*APC*)** Directed labeled graph, where nodes can be an author or a paper, and we can have edges between papers (citation) or between authors and papers (authorship).

**Countries Citation Graph (*CC*)** Directed multigraph, where each node represents a country of origin, and edges represent citations.

As our work is focused on the flagship AI and adjacent fields conferences, we filtered their dataset to contain only the papers published in these conferences to build ours. The chosen conferences were based on CSRankings CSRankings [2019] top-ranked AI conferences, which include the following fields: Artificial Intelligence, Computer Vision, Machine Learning & Data Mining, Natural Language Processing, and The Web & Information Retrieval. For each of the graphs explained above, we calculated the following exact centralities: degree (in and out) (Section 2.4.2), betweenness (Section 2.4.3), closeness (Section 2.4.4), and PageRank (Section 2.4.5).

For our work, we created the cumulative graph for each year from 1969 (the first IJCAI conference) until 2019, i.e. the cumulative graph for the year 2000 contains all the papers before and including 2000. A graph for each individual year from 1969 to 2019 was also created, to help with the analysis presented in the sections below. The cumulative graphs containing all the data, including exact centralities, were made available at `https://github.com/rafaeelaudibert/conferences_insights_database`. The cumulative graphs for the entire Authors Citation dataset, not restricting it by conference, were also made available in the same repository, without computing the centralities. We can find the statistics for the size of each graph dataset in Table 3.

## 3.3 Types of Graphs

The graphs were built in Python using *networkx* Hagberg et al. [2008] which provides an easy interface to build various types of graphs, including multigraphs with directed edges, which we routinely use.

All graphs below are based on the data shown in Figure 7.

| | *Graph* | *Nodes* | *Edges* |
|---|---|---|---|
| CS Conferences | ACi | 104179 | 5654596 |
| | ACo | 104179 | 621644 |
| | PC | 89102 | 486373 |
| | APC | 193281 | 759386 |
| | CC | 93 | 4776703 |
| Full DBLP | ACi | 3655049 | 210362459 |

Table 3: Graph Statistics for the cumulative data

#### 3.3.1 Author Citation Graph

This is a directed multi-graph, where every author is a node. An edge $e_{u,v}$ represents a paper from author $u$ having a citing to a paper by author $v$. As author $u$ can have more than one paper citing a paper by author $v$ there might be more than one edge between the nodes, therefore we have a multi-graph. Also, authors might cite another paper from themselves, therefore we might have self-loops.[37]

Because of the way our data is organized, when we are iterating over the papers we have only the id of the papers that were referenced, but not the ID of the authors in the other papers. So, we first create a hash table with keys as the paper IDs and the value as the authors of that paper. We use this as a lookup table to identify which authors should be connected when we are iterating over the papers. See Algorithm 2 to see how this works when building the graph.

The above means that we first need to iterate over all papers and create this huge lookup table. In practice, because you cannot cite papers that have not yet been published, we split the papers into buckets by the year they were published and iterate in ascending years, which will make us keep only the "past" papers in this hash table. Algorithm 1 shows the year bucket-splitting algorithm and Algorithm 2 shows how we build this graph, with this more efficient hash table where at any year $y$ we only have papers from years $i <= y$ in the hash table. Although at the end of the process, the table has the same size as it would have if we had built it from the beginning, this method increases local consistency improving cache results when we are iterating over the first years making this process more efficient.

Note that we might not have data for the cited paper because we are filtering the data out for only a few conferences. In this case, we simply do not include this paper.

[37] The most recent version of the code for this graph generation process can be found in `https://github.com/rafaeelaudibert/conferences_insights/blob/v11/graph_generation/generate_authors_citation_graph.py`.

```
[
  {
   id: '1',
    title: 'Survey about Graphs',
    authors: [
      { id: '1', name: 'John Doe', org: 'MIT' }
    ],
    venue: { raw: 'Some Conference' },
    references: [],
    year: 1967
  },
  {
   id: '2',
    title: 'Survey about Bigger Graphs',
    authors: [
      { id: '2', name: 'Mary Jane', org: 'UFRGS' },
      { id: '3', name: 'Jane Carl', org: 'TU KL' },
    ],
    venue: { raw: 'Some Conference' },
    references: ['1'],
    year: 1970
  },
  {
   id: '3',
    title: 'Survey about Huge Graphs',
    authors: [
     { id: '2', name: 'Mary Jane', org: 'UFRGS' }
    ],
    venue: { raw: 'Some Conference' },
    references: ['1', '2'],
    year: 2003
  }
]
```

Figure 7: Sample data for graphs

Figure 8 shows an example of such a graph, given the input data from Figure 7.

### 3.3.2 Author Collaboration Graph

This is an undirected graph, where every author is a node. In this graph, an edge $e_{u,v}$ represents that $u$ and $v$ worked together in at least one paper.[38]

[38] The most recent version of the code for this graph generation can be found in `https://github.com/rafaeelaudibert/conferences_insights/blob/v11/graph_generation/generate_collaboration_graph.py`.

**Algorithm 1** Bucket-splitting paper per year

**Require:** $L$ ▷ List of papers such as the example in Figure 7

papers_per_year ← empty hashtable

**for** year = 1969...2018 **do**
  papers_per_year[year] ← empty list
**end for**

**for** paper ∈ $L$ **do**
  papers_per_year[paper.year] ≪ paper ▷ ≪ means append
**end for**

**return** papers_per_year

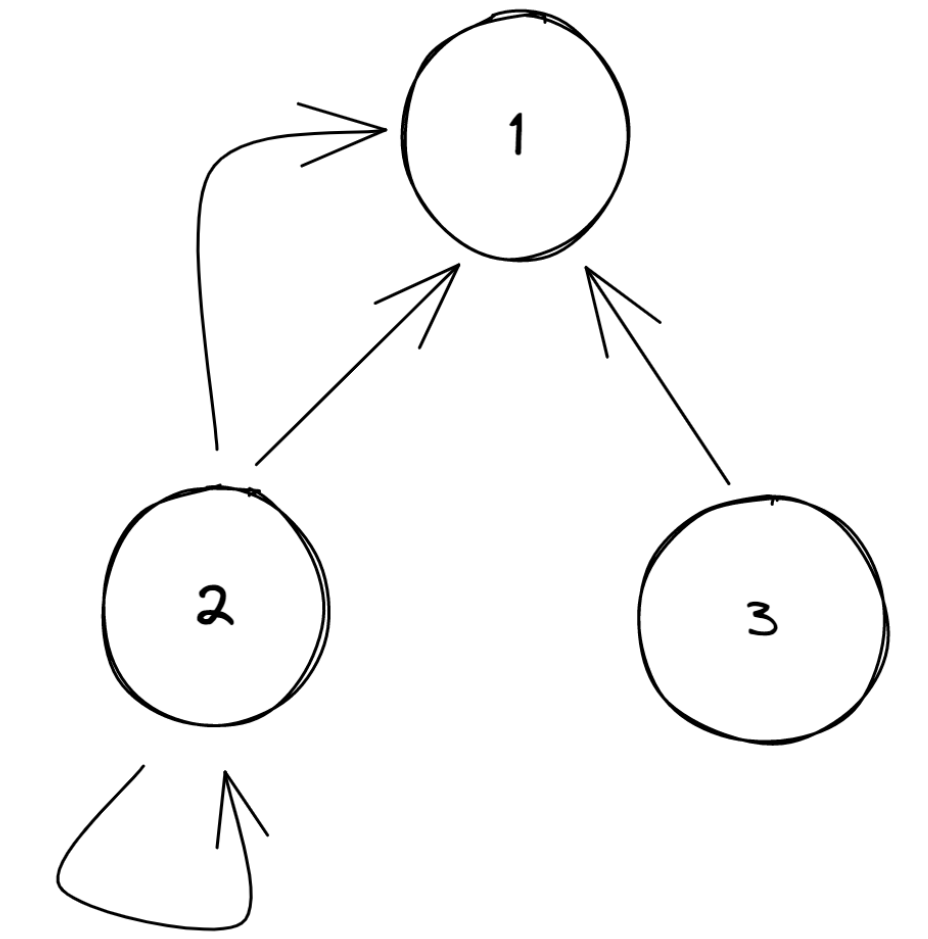

Figure 8: Example of author citation graph

This graph is easier to generate compared to the Author Citation Graph (Section 3.3.1) because data is local and we do not need to iterate twice over the data to generate a lookup table: we can simply iterate over all papers and then connect all co-authors in a clique.

Figure 9 shows an example of such a graph, given the input data from Figure 7.

**Algorithm 2** Author Citation Graph

```
Require: papers_per_year                    ▷ Hash table as returned by Algorithm 1
  G ← new graph with empty V and E
  old_papers ← {}

  for year = 1969...2018 do
      papers ← papers_per_year[year]
      for paper ∈ papers do
          old_papers[paper.id] ← id of every author in paper.authors
      end for

      for paper ∈ papers do
          for author ∈ paper.authors do
              G.V ← G.V ∪ {author.id}
          end for

          for citation_id ∈ paper.references do
              if citation_id ∈ old_papers.keys then
                  for cited_author ∈ old_papers[citation_id] do
                      for author ∈ paper.authors do
                          G.E ← G.E ∪ { (author.id, cited_author.id) }
                      end for
                  end for
              end if
          end for
      end for
  end for

  return G
```

### 3.3.3 Papers Citation Graph

This is a directed graph, where every paper is a node. A directed edge $e_{u,v}$ means that paper $u$ cited paper $v$. Similar to the Authors Citation Graph we need to create a lookup table, using the same incremental procedure of loading in the lookup table data only for years $i <= y$ when iterating over year $y$. [39]

[39] The most recent version of the code for this graph generation can be found in `https://github.com/rafaeelaudibert/conferences_insights/blob/v11/graph_generation/`

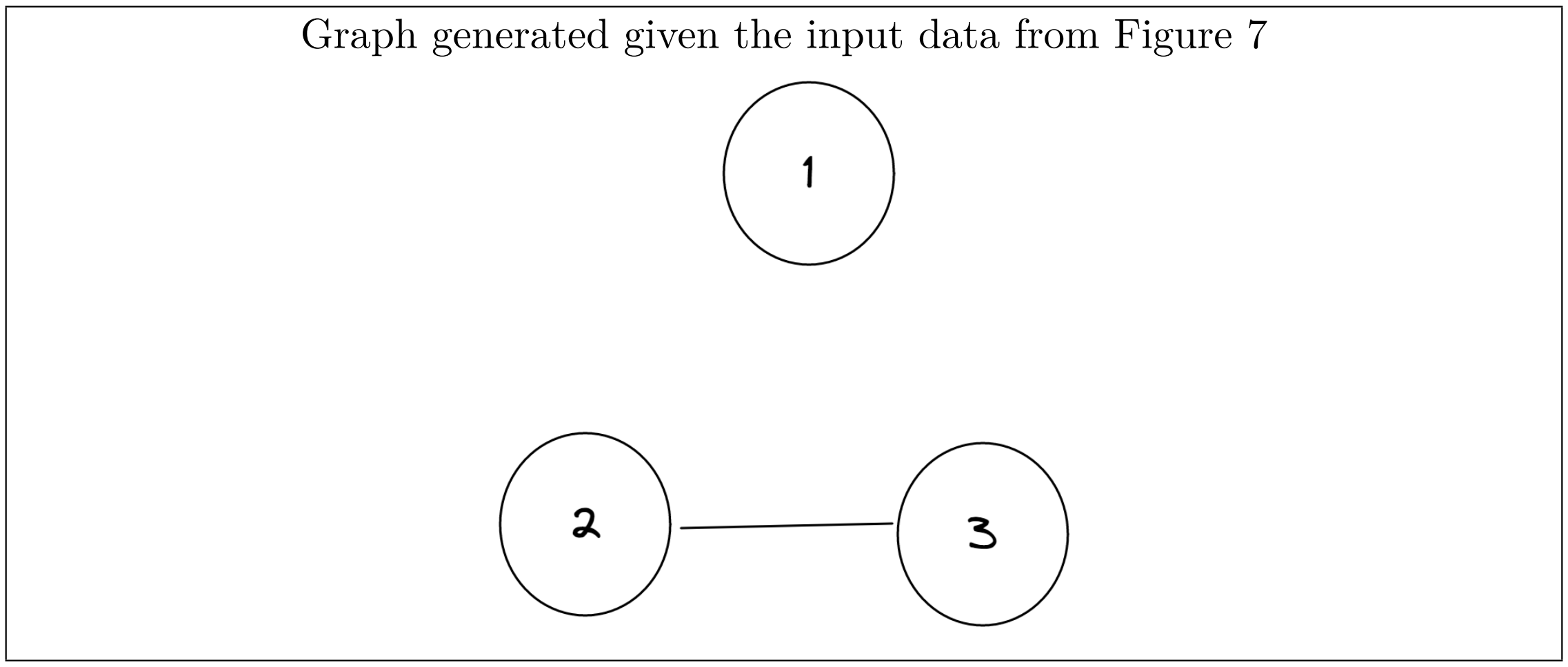

Figure 9: Author collaboration example graph

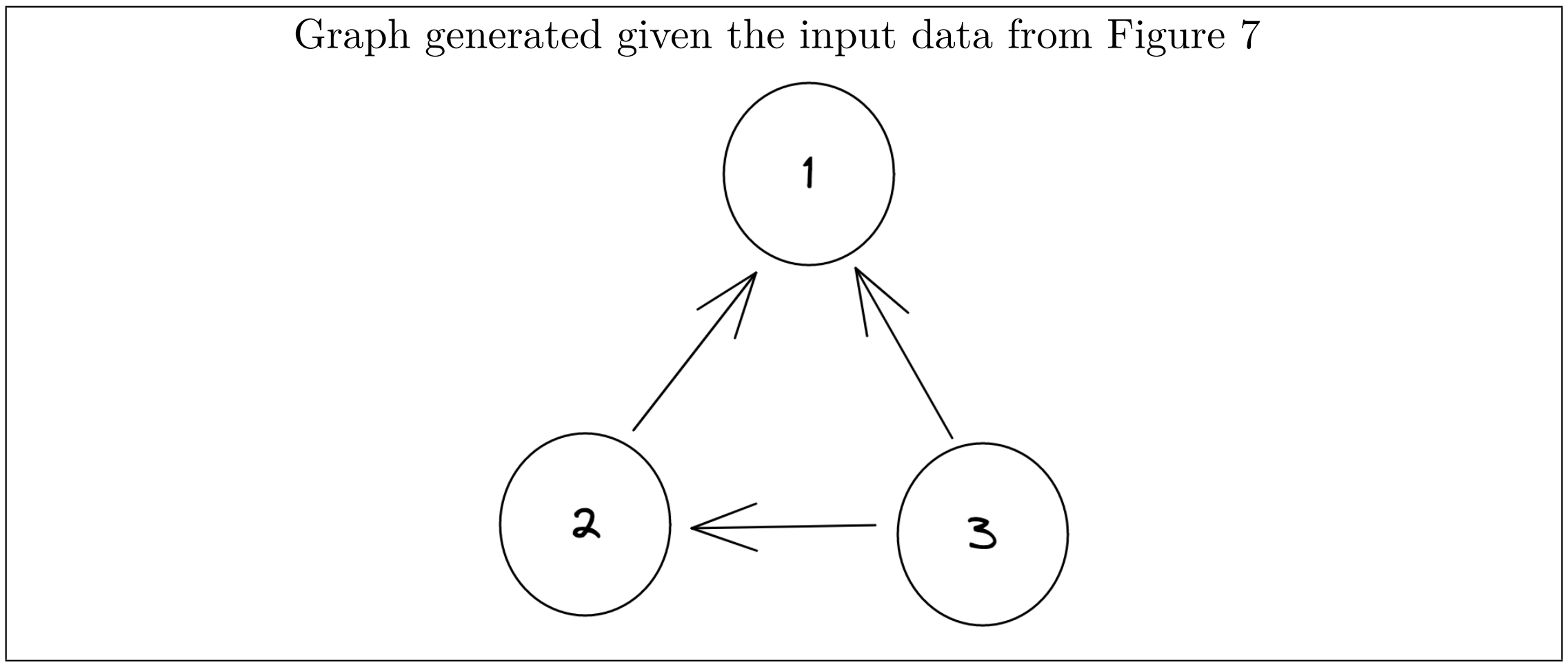

Figure 10: Paper citation example graph

Figure 10 shows an example of such a graph, given the input data from Figure 7.

#### 3.3.4 Author-Paper Citation Graph

This is a directed labeled graph, where nodes can be either an author or a paper, and we can have edges between papers or between authors and papers, therefore this graph is more complex than the previous ones because it can represent both a paper citation network and an author citation network (through intermediate paper

`generate_citation_graph.py`.

nodes).[40]

This graph is built based on the *Papers Citation Graph*, with the already existent nodes being from the type *paper* $V_P$, and the already existent edges being from the type *citation* $E_C$. After, we add a node with type *author* $V_A$ for each author, with a directed edge with type *authorship* $E_A$ for each paper they authored.

This graph is ideal for a full picture of the data, with the possibility of inferring every possible interaction in it. Therefore, it is an ideal representation for knowledge representation tasks or even recommender systems. This is discussed in more detail in Section 5.2.

Figure 11 shows an example of such a graph, given the input data from Figure 7.

### 3.3.5 Country Citation Graph

This is a directed multigraph, where each node represents a country, and an edge $e_{u,v}$ represents that an author from country $u$ cited an author from country $v$ in a paper. Because of this two nodes might have many edges between them.[41]

After we have figured out which country an author is from (Details in Section 3.4) we can create this graph by doing the same procedure for the citation graph. Save the papers already existing by that time in a lookup table; iterate over every paper; iterate over the citations; iterate over the current paper authors and the cited paper authors; connect the country they belong to with an edge. It is possible (and quite common) to create self-loops.

Figure 12 shows an example of such graph, given the input data from Figure 7, in addition to the following mapping from organizations to countries: MIT[42] $\rightarrow$ USA; UFRGS[43] $\rightarrow$ Brazil; TU KL[44] $\rightarrow$ Germany.

## 3.4 Affiliation x Country mapping

It is important to note that the Arnet v11 data we collected does not always provide the country of an author in its “org” field, containing only the organization they belong to – it sometimes does not even provide the organization – which poses a problem.

[40] The most recent version of the code for this graph generation can be found in `https://github.com/rafaeelaudibert/conferences_insights/blob/v11/graph_generation/generate_authors_and_papers_graph.py`.

[41] The most recent version of the code for this graph generation can be found in `https://github.com/rafaeelaudibert/conferences_insights/blob/v11/graph_generation/generate_country_citation_graph.py`.

[42] `https://www.mit.edu/`

[43] `https://www.ufrgs.br`

[44] `https://www.uni-kl.de/`

Graph generated given the input data from Figure 7

*Red nodes indicate they have type $V_P$;*

*Red edges indicate they have type $E_C$;*

*Blue nodes indicate they have type $V_A$;*

*Blue edges indicate they have type $E_A$;*

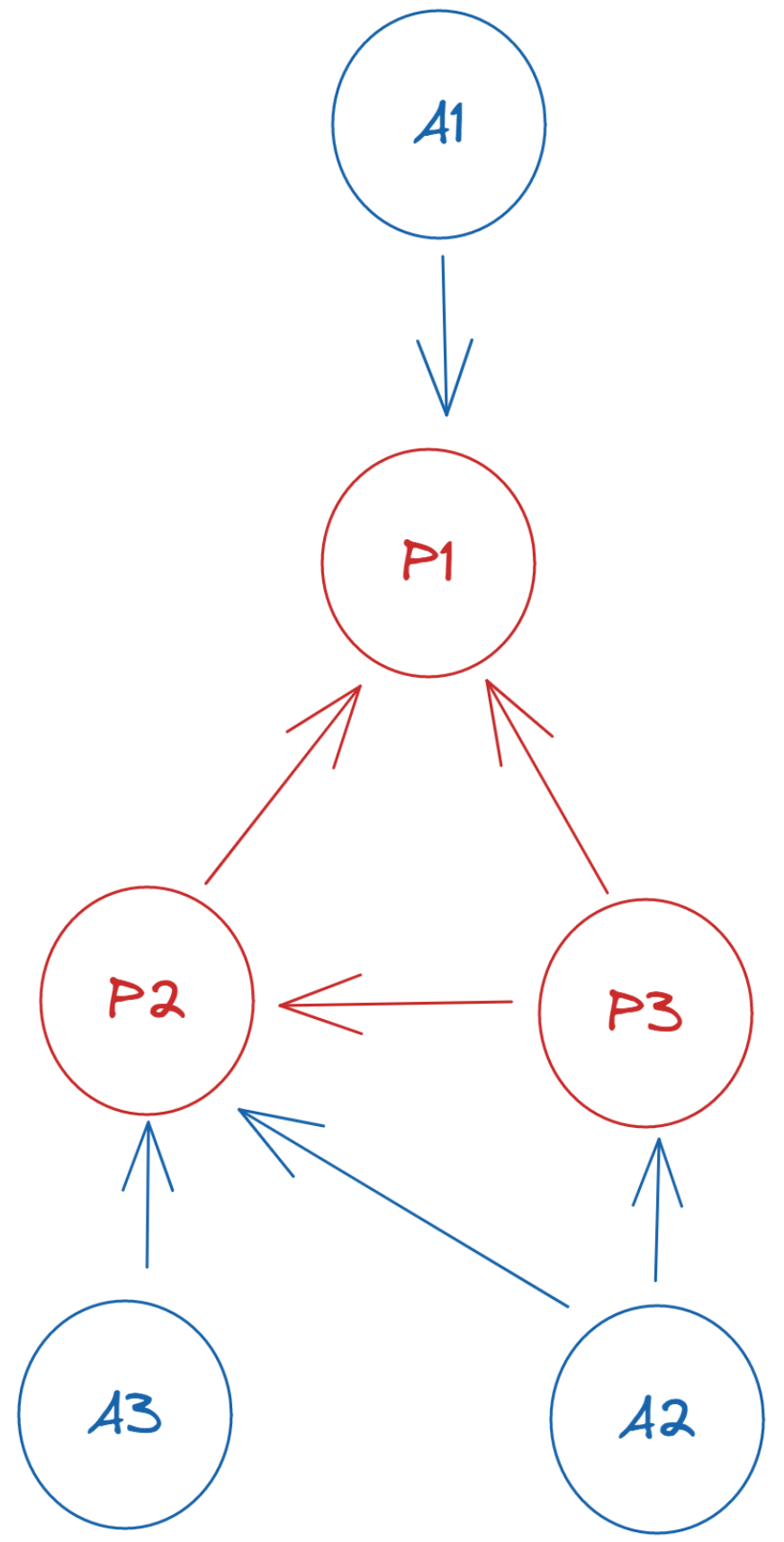

Figure 11: Author Paper Citation example graph

The "organization" field present in the data is in free-form format, which means that it does not have a clear structure from which we can extract the country of an author. Even worse, it might not even be a university name, as both companies and non-affiliated individuals can have papers in flagship venues. There is some structure in it for most of the data, though, so we have developed a pipeline where we iteratively try to detect an organization's country of origin.

In our pipeline, we first preprocess the organization by following Algorithm 4 removing cluttering and using only the text after the last comma – ideally where the country of affiliation should be. After, Algorithm 3 is followed. We try matching the text against a lookup table that maps organizations to countries. If there's a miss, we split the text into spaces and try matching only the first word to the table, and after only the last word. If that still does not work we try matching the text without the preprocessing step.

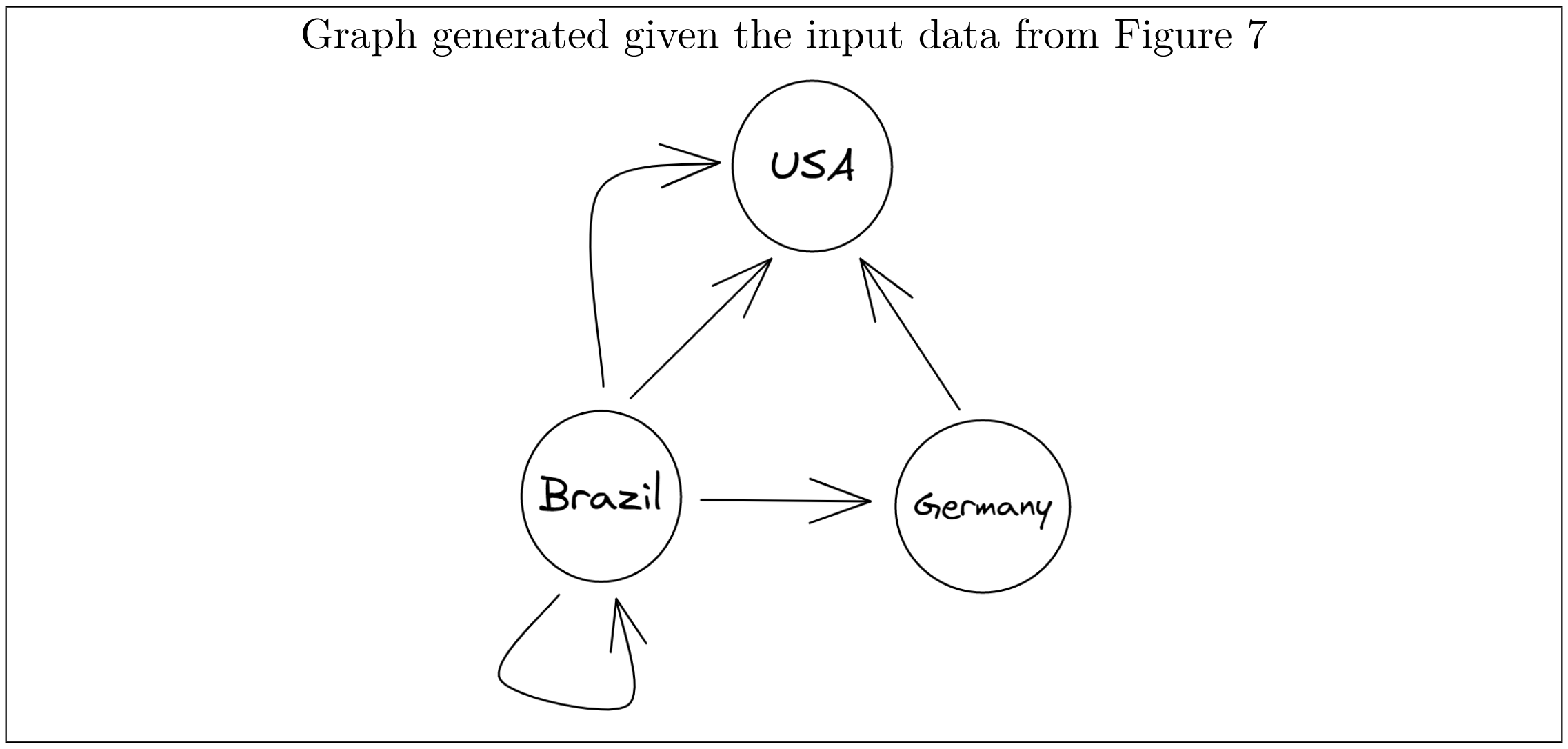

Figure 12: Countries citation example graph

In the end, if everything fails, we check if we matched that author previously. That is our last resort because remember that the author might change organization (and even country) throughout their academic career, so we cannot trust an author will still be in the same organization as they were the last time they published something.

The above process can be seen in the *infer_country_from* function[45].

We do have another important step not fully explained in the steps above: how we created the "lookup table" to map from organizations to countries. We manually created it over the span of 2 months, through a manual iterative labor-intensive process: manually looking at the organizations not matched using Algorithm 3 and mapping them to the countries they belong to using both our own knowledge and web searches to filter the options down.[46]. The mapping for every organization that has ever been published in AAAI, IJCAI, and NeurIPS is complete, and the process

[45] `https://github.com/rafaeelaudibert/conferences_insights/blob/v11/graph_generation/generate_country_citation_graph.py`

[46] This mapping is available at `https://github.com/rafaeelaudibert/TCC/blob/v11/graph_generation/country_replacement.json`

**Algorithm 3** Organization to Country Mapping

```
Require: raw_org                                        ▷ Organization name
Require: org              ▷ Organization name preprocessed by Algorithm 4
Require: author_id
Require: T                   ▷ Lookup table matching organization to country
Require: PT                          ▷ Past author to organization matchings
  if org ∈ T.keys then                 ▷ Check if preprocessed org is in the table
      return T[org]
  end if

  split_org ← split("org", " ") ▷ Split the text into every space, turning it into a list
  if split_org[0] ∈ T.keys then          ▷ Check if first name in org is in the table
      return T[split_org[0]]
  end if
  if split_org[-1] ∈ T.keys then          ▷ Check if last name in org is in the table
      return T[split_org[-1]]
  end if

  if raw_org ∈ T.keys then   ▷ Check if org without preprocessing is in the table
      return T[raw_org]
  end if

  if author_id ∈ PT.keys then    ▷ Check if we have already matched this author
  before
      return PT[author_id]
  end if

  return ∅
```

to map this for the other conferences is still ongoing. We hope this mapping can be used in the future by other works to facilitate the inference of a country from an organization. Figure 13 shows how many organizations we mapped per country – the USA does not fit in the figure for scale purposes and has a value of 2163.

Additionally, there are a few authors whose "org" field is empty. For the first years of the area (1969-1979), we did not have many papers being published, so we manually looked at every single paper with an empty organization field and generated another lookup table available at `https://github.com/rafaeelaudibert/TCC/blob/v11/graph_generation/author_country_replacement.yml`. We then

*The USA does not fit in the figure; and has a value of 2163.*
*This does not reflect the true count of different organizations per country, because some of them could be easily identified by their country, and did not need any*

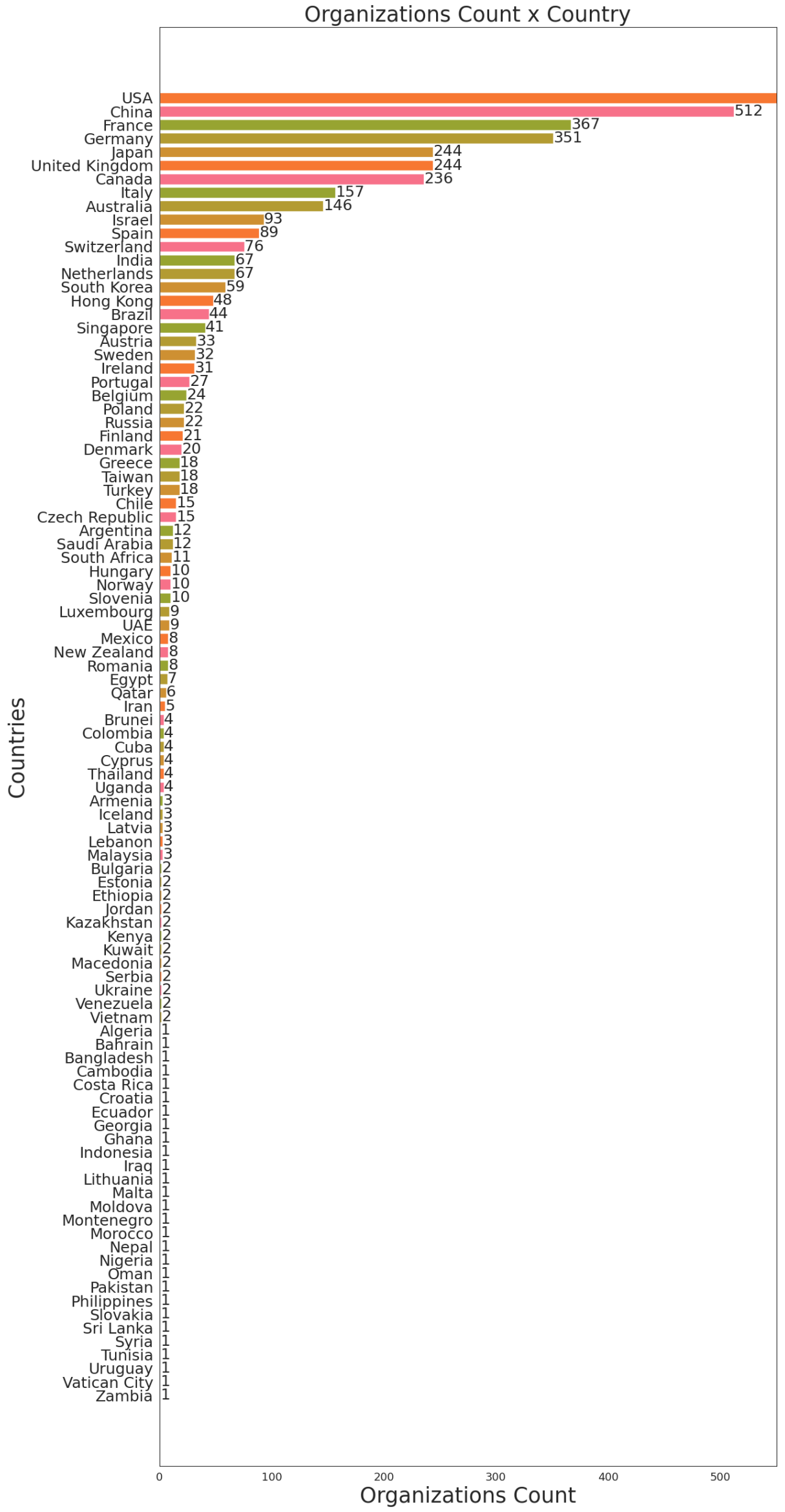

*special treatment.*

Figure 13: Quantity of **mapped** different organizations per country that appeared in our data.

check this table first before attempting the above pipeline, since it is more reliable. This table was specially built in YML instead of JSON for better readability and allows us to add comments in between the entries.

It is noticeable, though, that this problem is worse in more recent years. Arnet's data does not have organizations for most papers published from 2018 onward, so the problem is bigger in recent years. For example, in Figure 36 the "None" stacked part is bigger in recent years.

# 4 Data Analyses

This section presents the main analyses and insights performed on our datasets described in Section 3. We present initial statistics in Section 4.1, then analyse each graph (Sections 4.2 to 4.6). We then investigate the research impact of Turing Award winners in Section 4.7.

As already stated before, the full code for both the data generation and data analysis was made publicly available at `https://github.com/rafaeelaudibert/TCC/tree/v11`. The main code is in the branch *v11* because of the aforementioned problems with Arnet's V13.

## 4.1 Raw Data

Although the bulk of this work is intended to revolve around the graph datasets and their centralities built to support our claims, the raw data itself is also able to provide us with great introductory information to the following sections.

Figure 14 shows a boxplot with a rising trend in the number of authors per paper over the years. In this boxplot graph, the red dot represents the average number of authors per paper, the black line represents the median, the box per se represents the 95% percentile, while the black lines represent the 99% percentile – even showing a failure in the dataset with some papers with 0 authors in the late 1960s. The figure shows how the trend of several authors in a single paper, like Brown et al. [2020], Jumper et al. [2021], and Silver et al. [2016], is recent and rare with not more than 1% of the papers having 7 or more authors since 2004. It is noticeable how the average value jumps to almost 4 in the years past 2014.

We also intersected the authors who published in the same year in different venues. Some interesting trends arose, such as AAAI and IJCAI have the biggest overlap in authors than any combination of them with NIPS and ACL (Figure 15); CVPR has congregated more authors than NIPS and IJCAI since the beginning of the 2000s and its biggest authors overlap is always with NIPS (Figure 16); SIGIR had almost no overlap with these three conferences during the 90s and still has very little overlap nowadays, despite an increase in its intersection with AAAI (Figure 17).

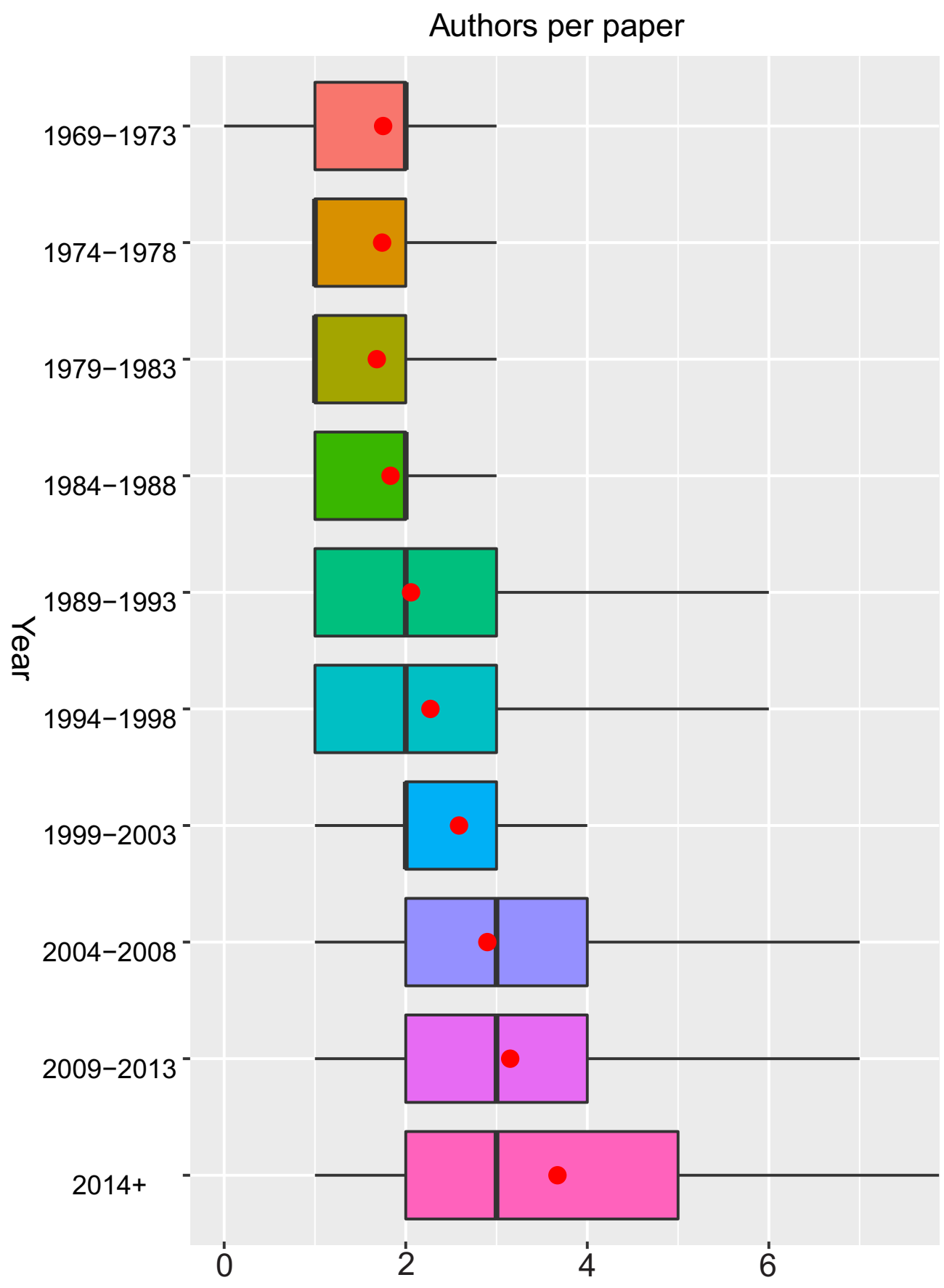

Figure 14: Boxplot of the number of different authors for each single paper per year

## 4.2 Author Citation Graph

### 4.2.1 Ranking over time

We have calculated an authors ranking regarding the aforementioned centralities from 1969 until 2019 using the accumulated citation data AC graph.

Figures 18, 19, and 20 show the evolution of PageRank, Betweenness, and In-Degree centralities, respectively, in our Author Citation Graph. In these figures, a line represents a single author and its ranking evolution over time in some predefined years (chosen to be 1969, 1977, 1985, 1993, 2001, 2009, and 2014). The only authors

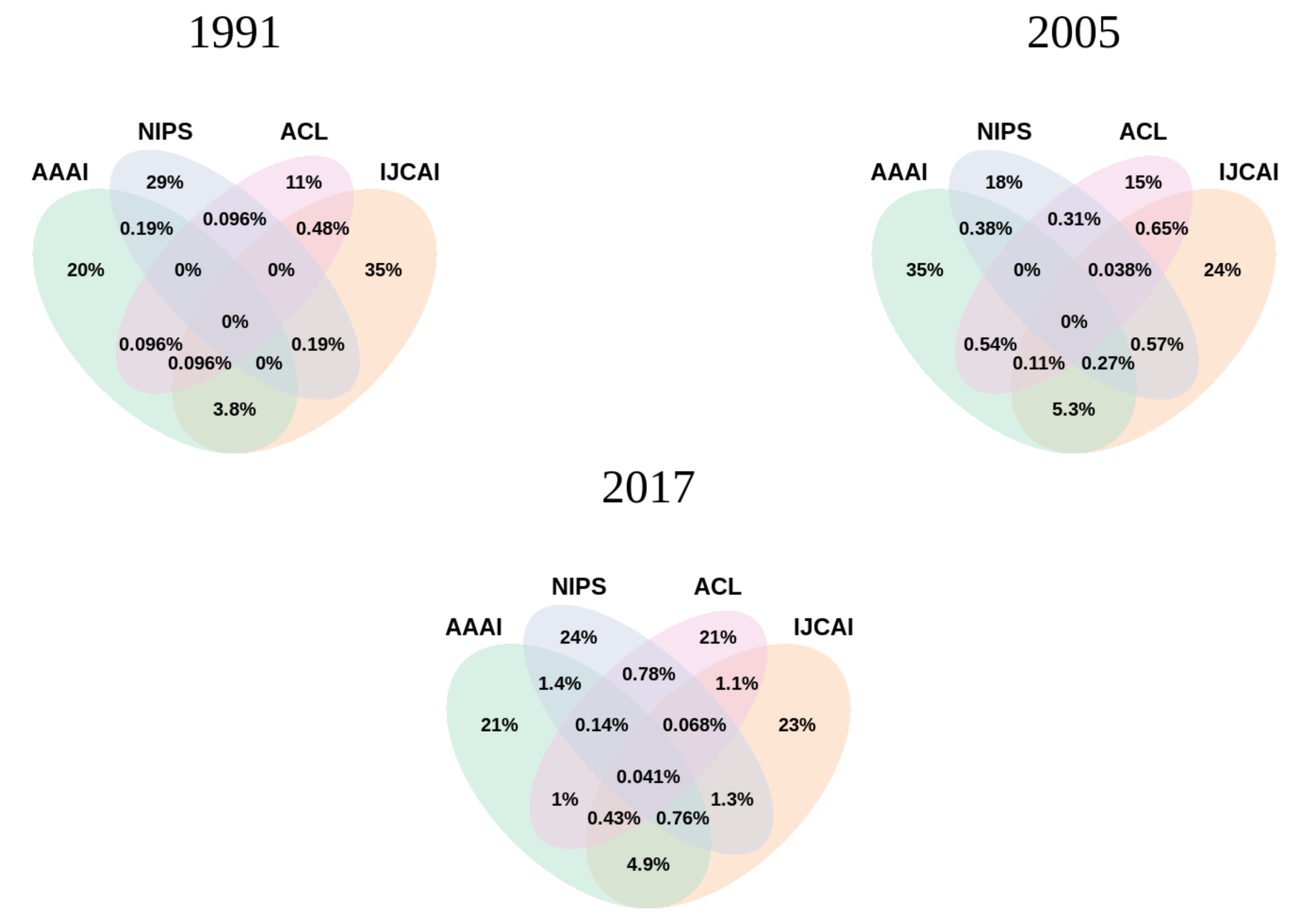

Figure 15: Percentage of overlapping authors in AAAI, NeurIPS (NIPS), ACL, and IJCAI.

shown are those who, at any point in one of these years, reached the top 10 in that specific rank. Authors who had not published yet in one of these years and, therefore, did not have any rank yet, show as *N/A*.

Although they seem chaotic, these graphs do have some interesting insights. Figure 18 is an interesting starting point because it is considerably stable, at least at the top of it. Harry Pople was the top 1 author in this ranking at least from 1977 until 2001, the longest period one will hold this position in any of our analyses. His main work is focused on Artificial Intelligence in Medicine therefore very central in-between different areas Dhar and Pople [1987]. Also in the PageRank graph, one might see that the rises tend to be meteoric with Andrew Ng going from position 974 in his debut year of 2001 to 16th 8 years later, and then 2nd after 5 more years. The same can be said for most of the dynamics present in this graph.

The aforementioned insights also hold for Figure 19 where Betweenness is analysed. This graph is a lot less stable than PageRank's, as betweenness is easier to evolve when new areas in Machine Learning happen, therefore changing the flow of information in the graph, while PageRank will be more stable because important people at one time will continue to be as important as they were forever, only going down in rank if someone even more influential appears. One can see this dynamic, for example, by looking at the last position in both charts: Larry Tesler – the one but last in the PageRank chart because the last position is an outlier – is 4267th in the PageRank, while the last position in the Betweenness chart is 31159th, showing how low one might drop in the Betweenness ranking even though they once were in the top 10 most influential scientists, *in the datasets analysed here*.

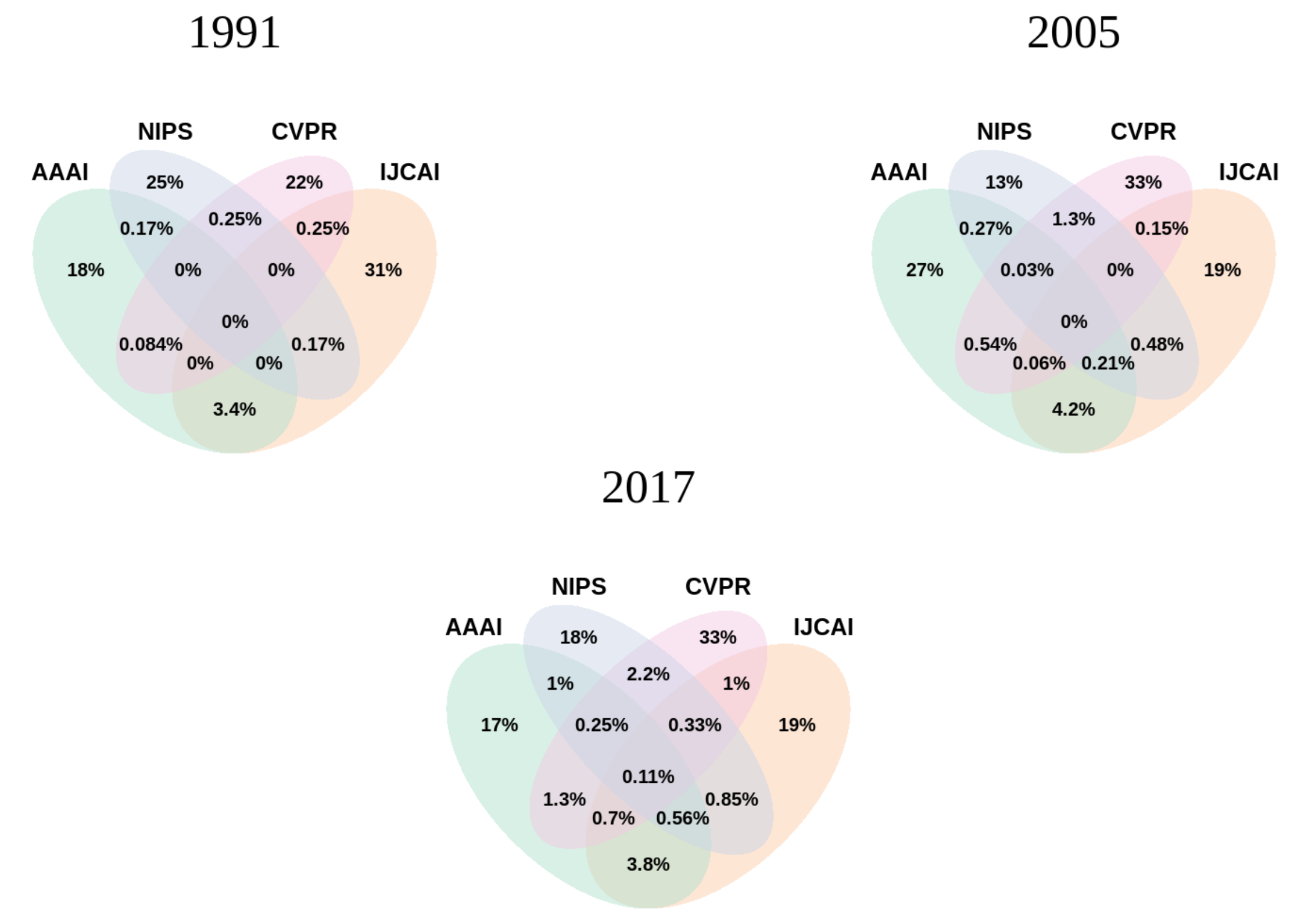

Figure 16: Percentage of overlapping authors in AAAI, NeurIPS (NIPS), CVPR, and IJCAI.

The Indegree chart shows a basic and raw data point: which author is the most cited, which should reward older authors with seminal papers. The first place in this ranking belongs to Andrew Zisserman, author of papers such as Simonyan and Zisserman [2014] and Hartley and Zisserman [2003], having close to 300,000 citations over his whole life – more than 100,000 of those only for the 2 cited papers. The second position is Andrew Ng with just over a third of the number of citations that Zisserman has.

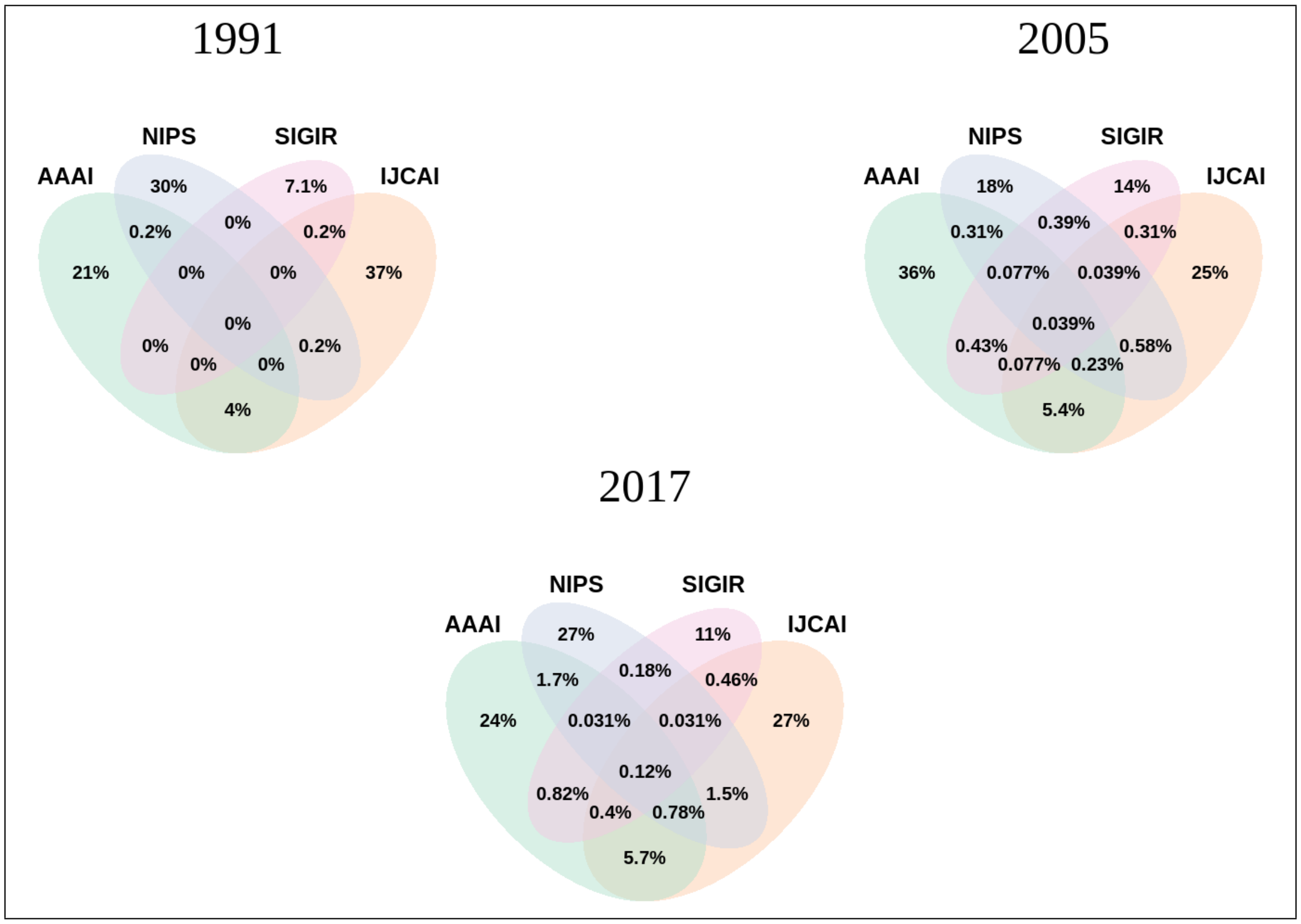

Figure 17: Percentage of overlapping authors in AAAI, NIPS, SIGIR, and IJCAI.

Considering all these charts together, it is interesting to see how Andrew Ng is the most influential overall author in AI when we analyse it from a citation perspective, *in the datasets analysed here*. He is the author of papers such as Blei et al. [2003], and Ng et al. [2001], having an h-index of 134, i.e. 134 papers with at least 134 citations (See H-Index on Section 2.4.6 to understand how this metric is computed), the 1403rd biggest h-index in Google Scholar[47]. He appears with the biggest betweenness value, and second-biggest indegree and PageRank ranking.

Takeo Kanade, the first position in the page rank ranking, is only 14th and 16th when looking at betweenness and indegree, respectively – although it is worthy of note that in 2001 he was first in in-degree and betweenness while third in PageRank. This is the best result, on average, that can be found in our results. Similarly,

[47] `https://www.webometrics.info/en/hlargerthan100`

Andrew Zisserman, the first position in the in-degree ranking, is sixth when looking at betweenness, and 8th on the PageRank ranking.

The ranking for the other computed centralities can be seen in Appendix C.

#### 4.2.2 Self-citations

Authors might build up in their previous work, which would introduce self-edges in our graph representing self-citations. Figure 21 shows a boxplot of the evolution of self-citations count per year. Despite the average beginning stable at around two, increasingly more authors have been increasing their number of self-citations over the years.

This figure, however, does not represent the full truth because there are more papers recently. Figure 22 shows a better view of the same data, clearly showing the average number increasing. The data has its faults because if an author can publish more than one paper per year then it will help to bring the average up by not being divided twice, but this can be said for every single year, so the increasing rate of self-citations would still exist.

### 4.3 Author Collaboration Graph

#### 4.3.1 Ranking over time

We have calculated an authors' ranking regarding the six aforementioned centralities from 1969 until 2019 using the accumulated collaboration data – *ACo* graph.

Figures 23 and 24 demonstrate how the PageRank and Betweenness rankings, respectively, evolved over time. In these figures, we chose to plot the top 10 authors each year, in an 8-year interval. Considering this gap, it is interesting to observe that only in 2009 it is possible to see all authors who appeared in the top 10 ranking during all the selected years. Also, most of the authors entered the ranking during the 80s and the 90s, regardless of the centrality. The remaining rankings (centralities) can be seen in the Appendix D.

#### 4.3.2 Entering the Realm of AI

Every year several researchers publish their first papers in AI-related venues such as the ones we are analyzing throughout this work. Figure 25 shows the yearly share of new authors per conference. The stacked area contains spikes due to the fact that several conferences did not occur yearly. NIPS conference (currently NeurIPS) was the conference that mostly attracted new authors until the mid-90s together with IJCAI. Since then the share has become more and more split into conferences of

*N/A stands for authors who had not published in the selected venues until that year.*

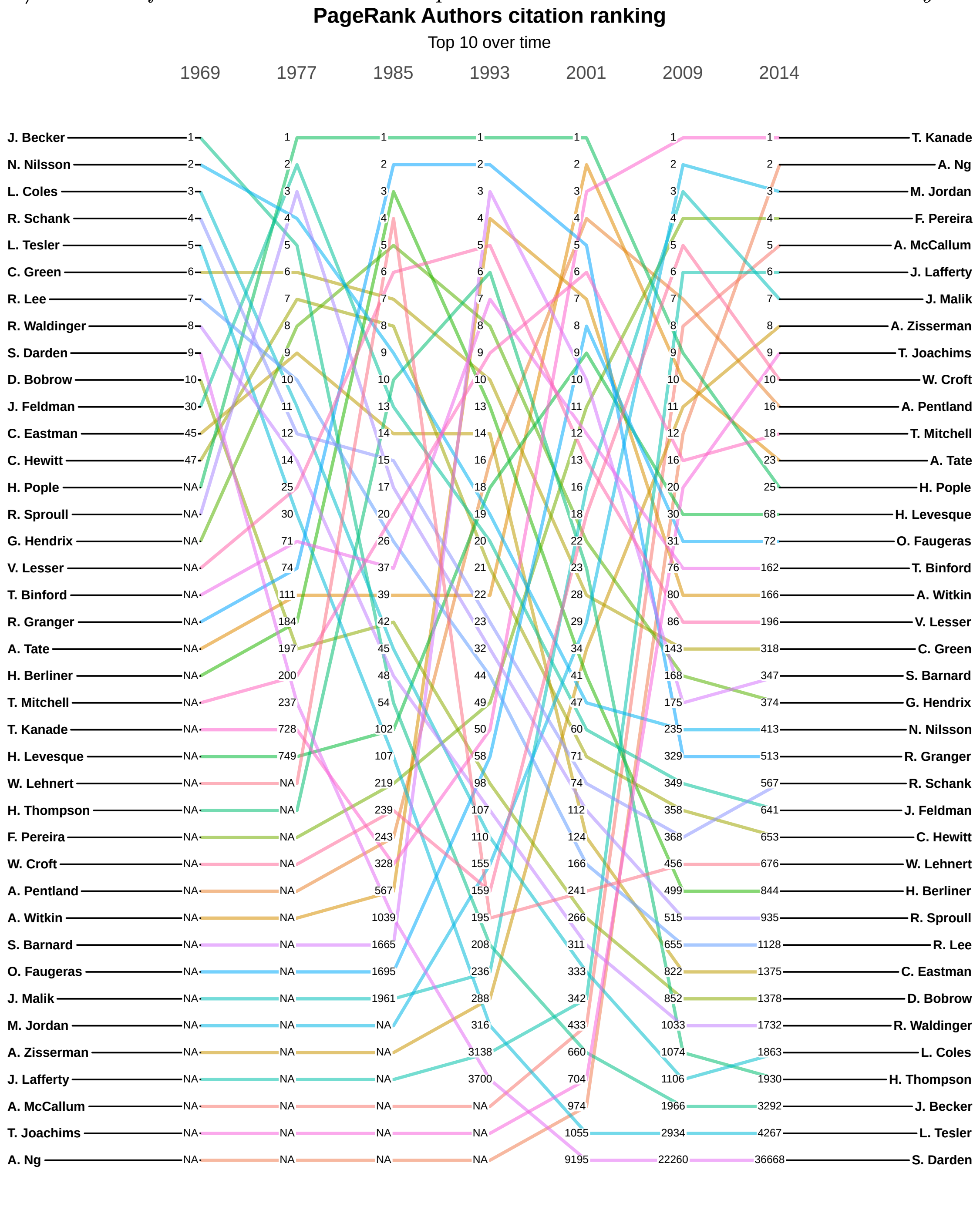

Figure 18: Author citation ranking over time according to PageRank centrality.

*N/A stands for authors who had not published in the selected venues until that year.*

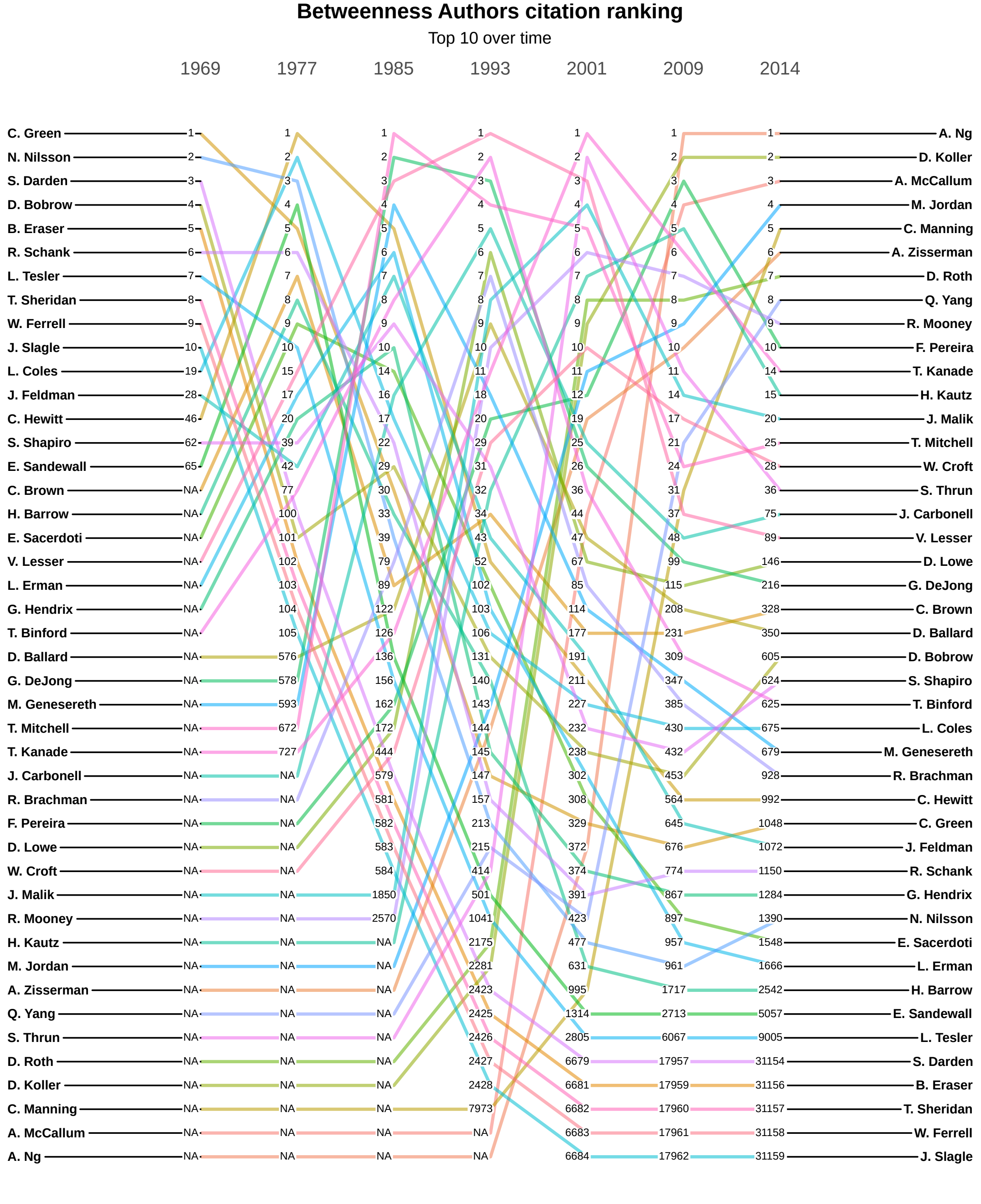

Figure 19: Author citation ranking over time according to Betweenness centrality

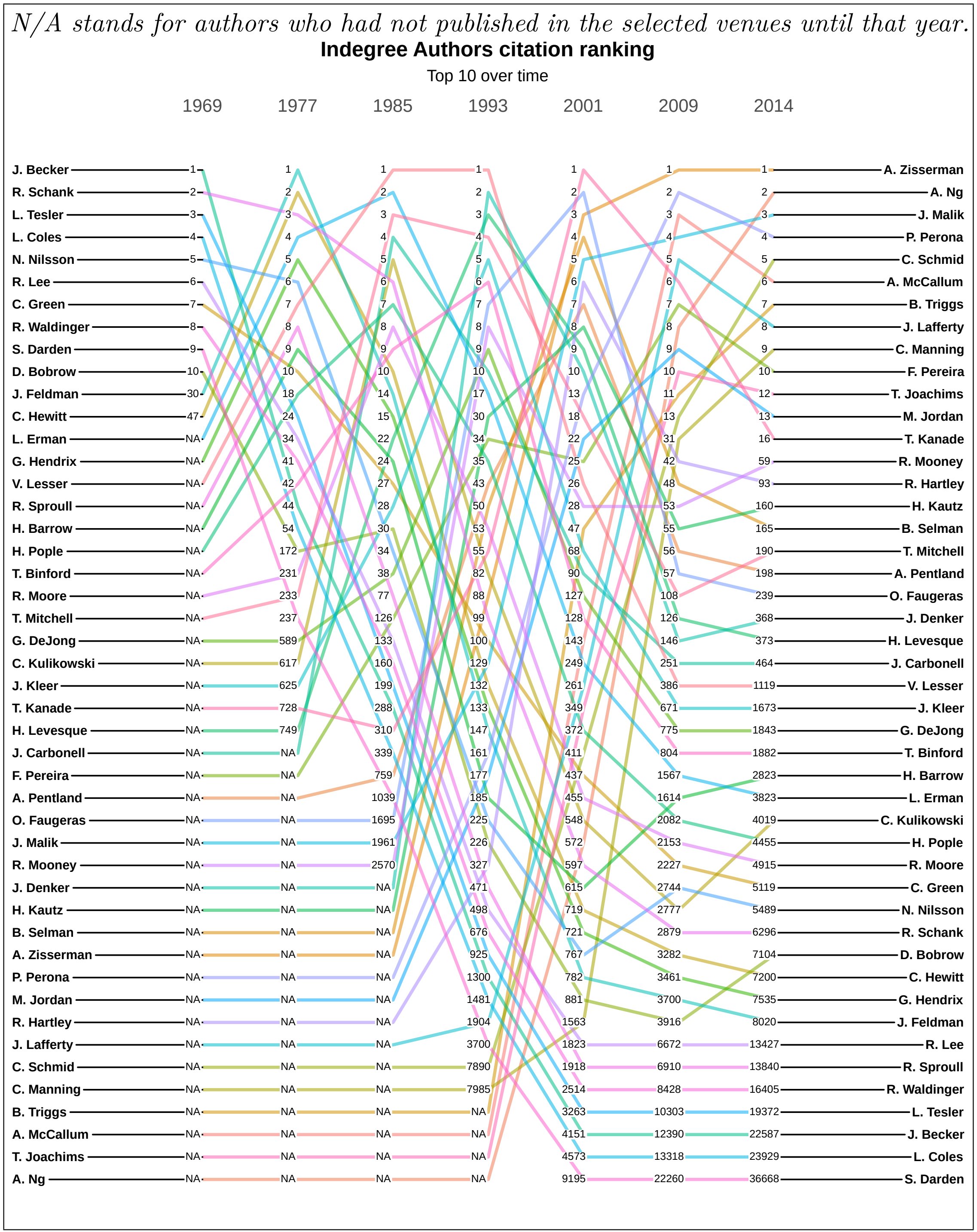

Figure 20: Author citation ranking over time according to In-degree centrality

Despite the average being stable around two, increasingly more authors have been increasing their number of self-citations over the years different areas. CVPR and ICCV have shown some growth in this respect in recent years, as did AAAI and WWW in the early 2000s.

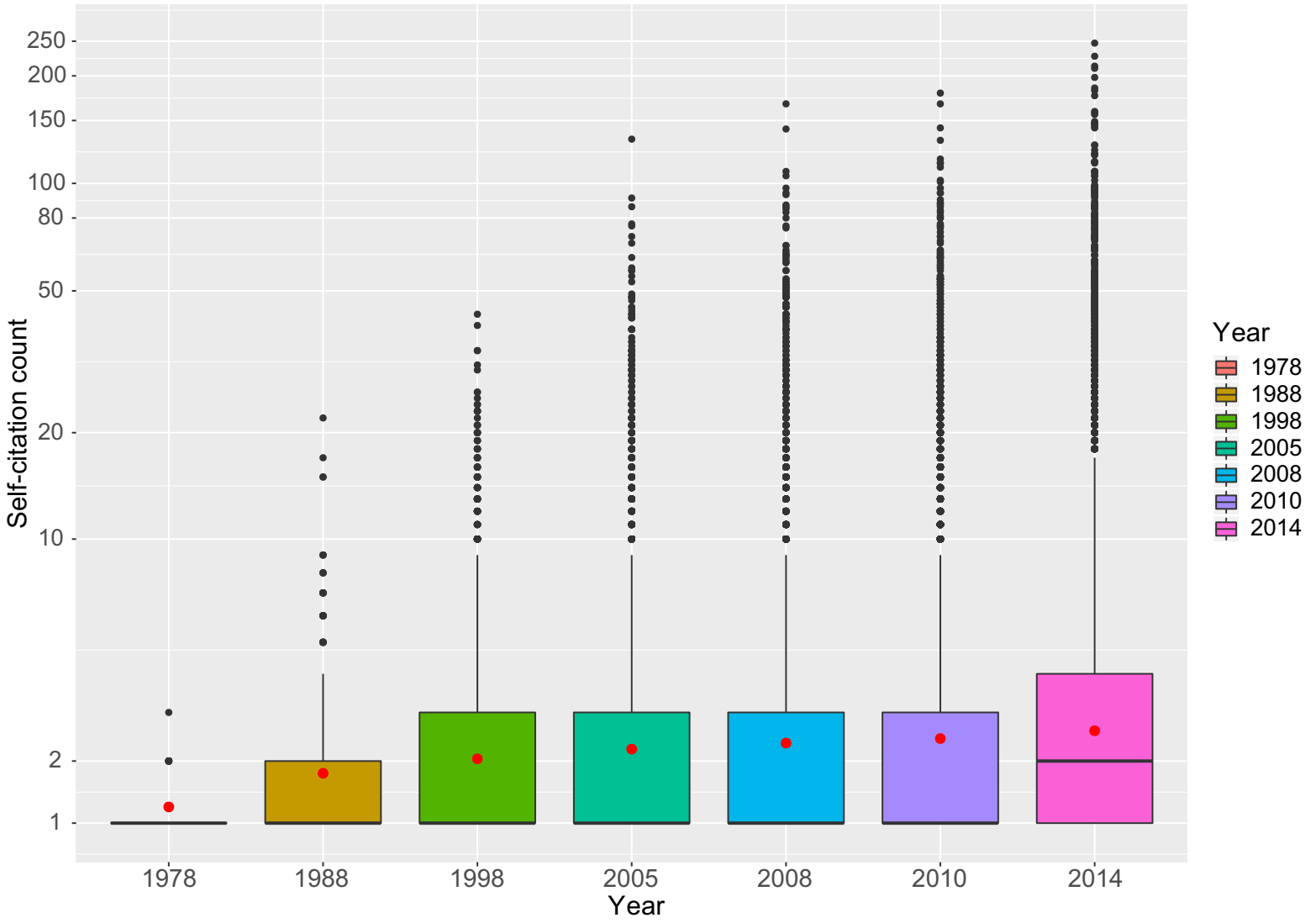

Figure 21: Boxplot of self-citation count per year

Table 4 lists all authors who collaborated with more than 200 new authors since 1969. Several of them appeared in the authors' collaboration ranking (especially regarding Betweenness, Closeness, and PageRank centralities). The regular behavior, however, is better described by the average and standard deviation statistics: the average number of new authors that an author collaborated with is around 4 with a standard deviation of 12. It seems clear that these numbers are highly affected by the career age of a researcher. We estimated this age with the author's first year of publication inside our graph and we used this age to normalize the amount of collaboration with new authors, achieving a normalized average number of new authors per author of 0.3 (career time average: 11) with standard deviation of 0.94 (career time standard deviation: 9.2), which essentially means that a researcher usually brings a new author to these AI venues after 3 years of his entry into the field.

Number of self-citations divided by number of authors who published during that

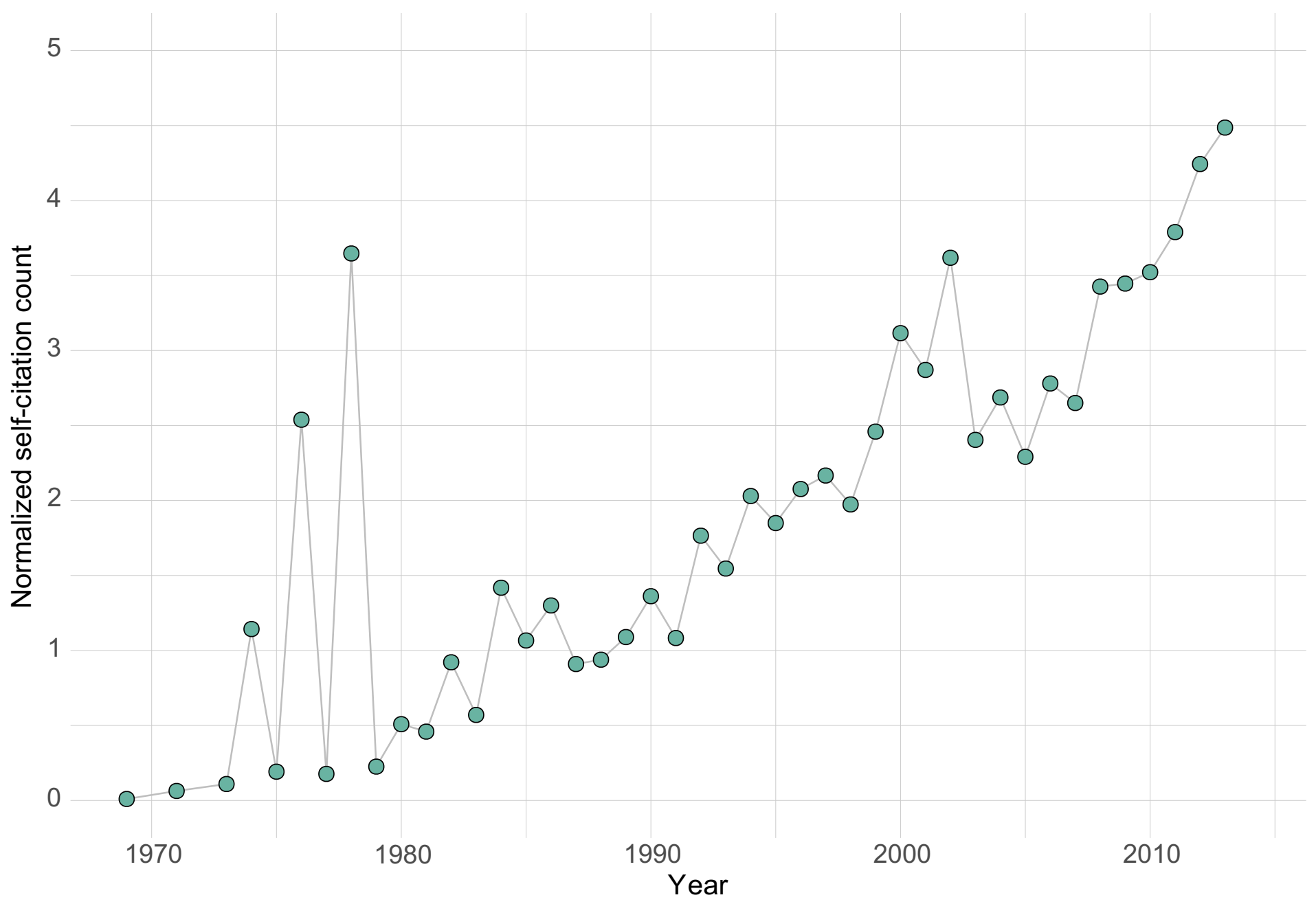

year

Figure 22: Normalized self-citation count per year

## 4.4 Paper Citation Graph

### 4.4.1 Ranking over time

In every centrality measure done for this graph, whenever we plot it, we map the name of the papers to those in Table 10. This format is not ideal for readability, but it was the best method found to show this data in its full form. When it comes to citation networks, the betweenness centrality can be seen as a measure of how a node (paper) is able to connect different research areas, or how it acts to foster interdisciplinarity Leydesdorff [2007].

In this sense, Figure 26 shows how the ranking of most important papers (according to betweenness centrality) evolved. It is possible to see that the ranking itself is very volatile as no paper can remain in the top 5 for more than 2 times (inside our gap of 8 years), nevertheless the paper "Constrained K-means Clustering with Background Knowledge" Wagstaff et al. [2001] (CKCWBK01 in the figure) has been in the top 10 at least since 2009. Also, all papers in the top 5 of 2017 and 2019 were published after the year 2000, which could indicate that, despite not being seminal

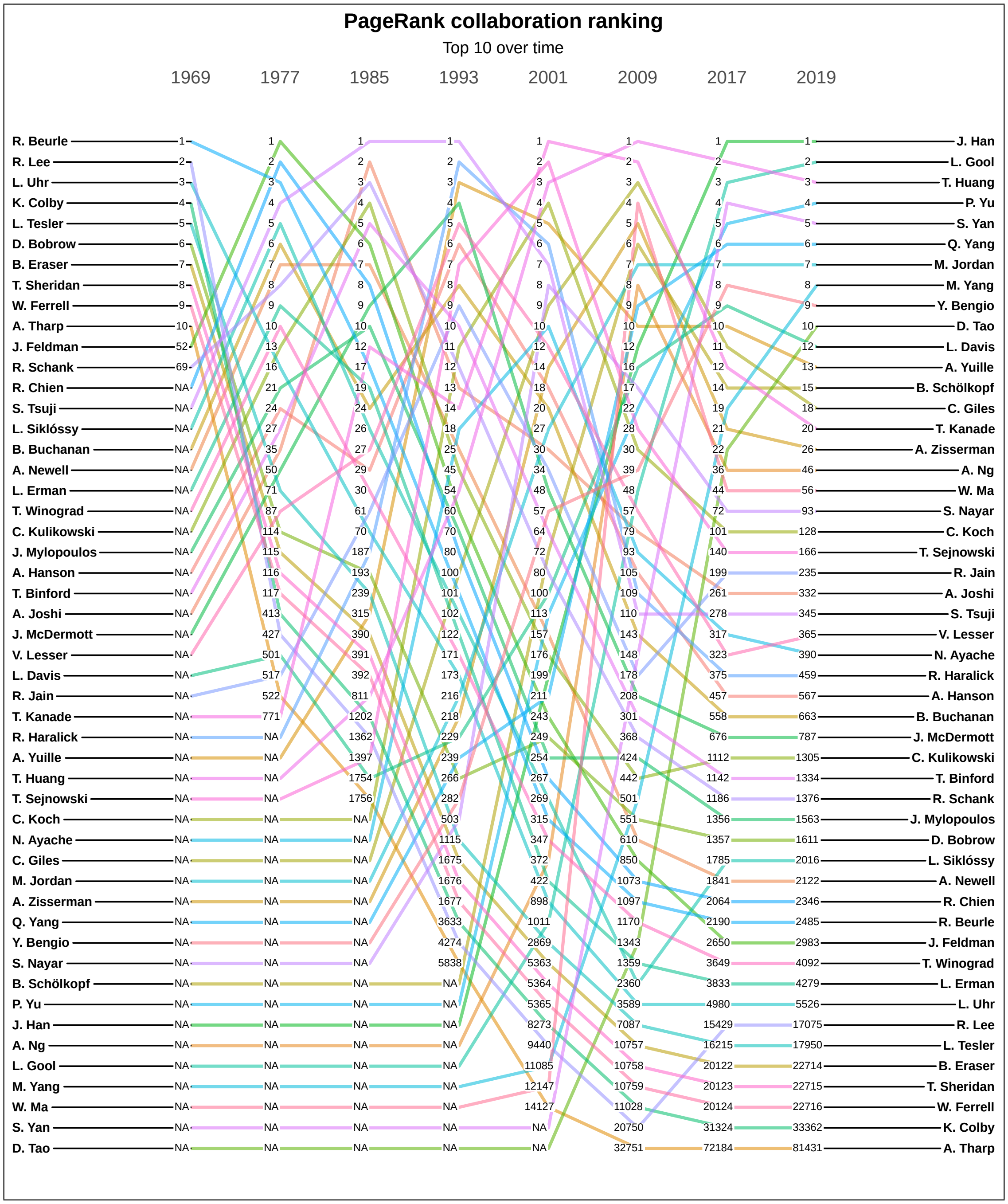

Figure 23: Authors collaboration ranking over time according to PageRank centrality. *N/A stands for authors who had not published in the selected venues until that year.*

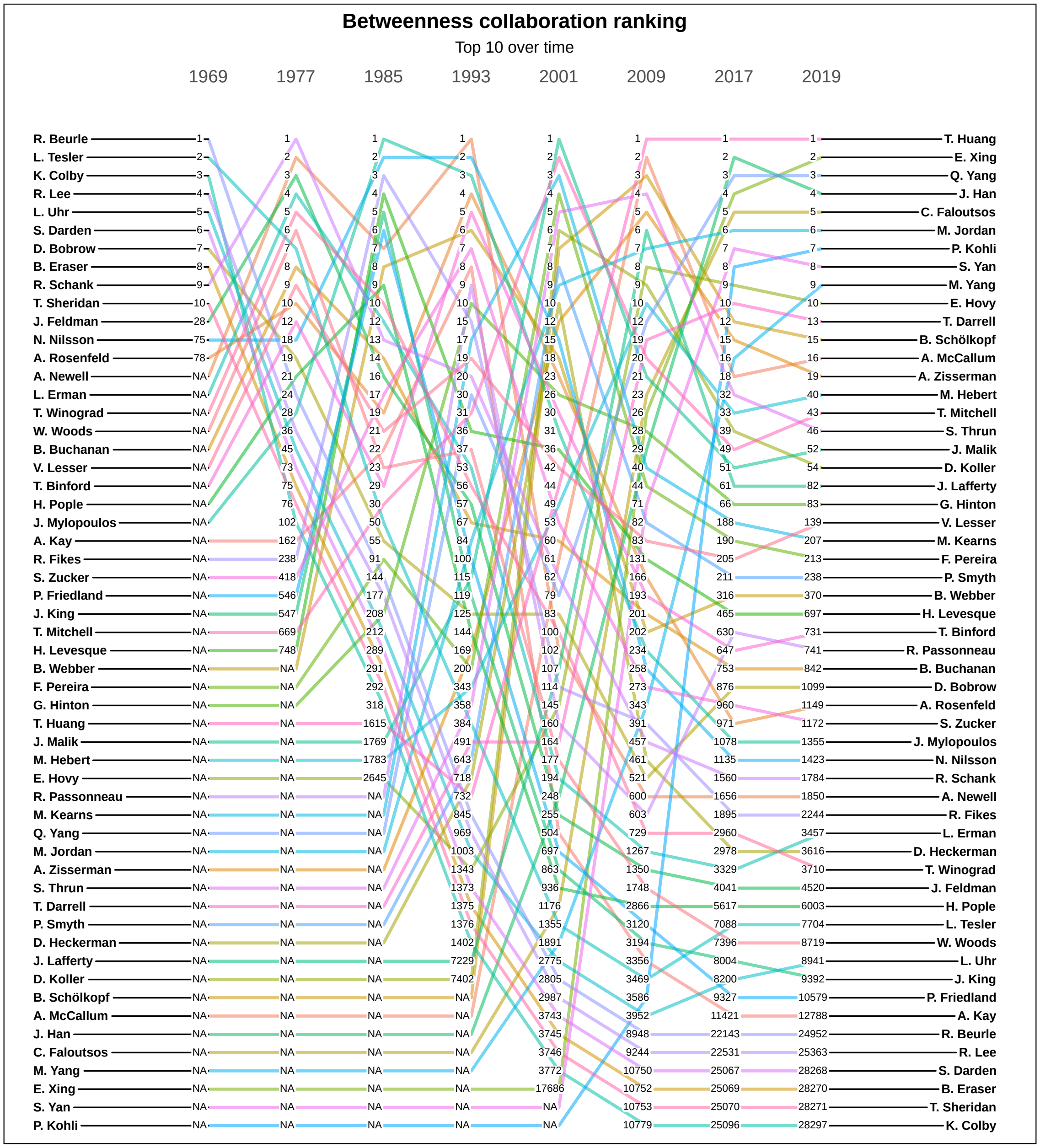

Figure 24: Authors collaboration ranking over time according to betweenness centrality.
*N/A stands for authors who had not published in the selected venues until that year.*

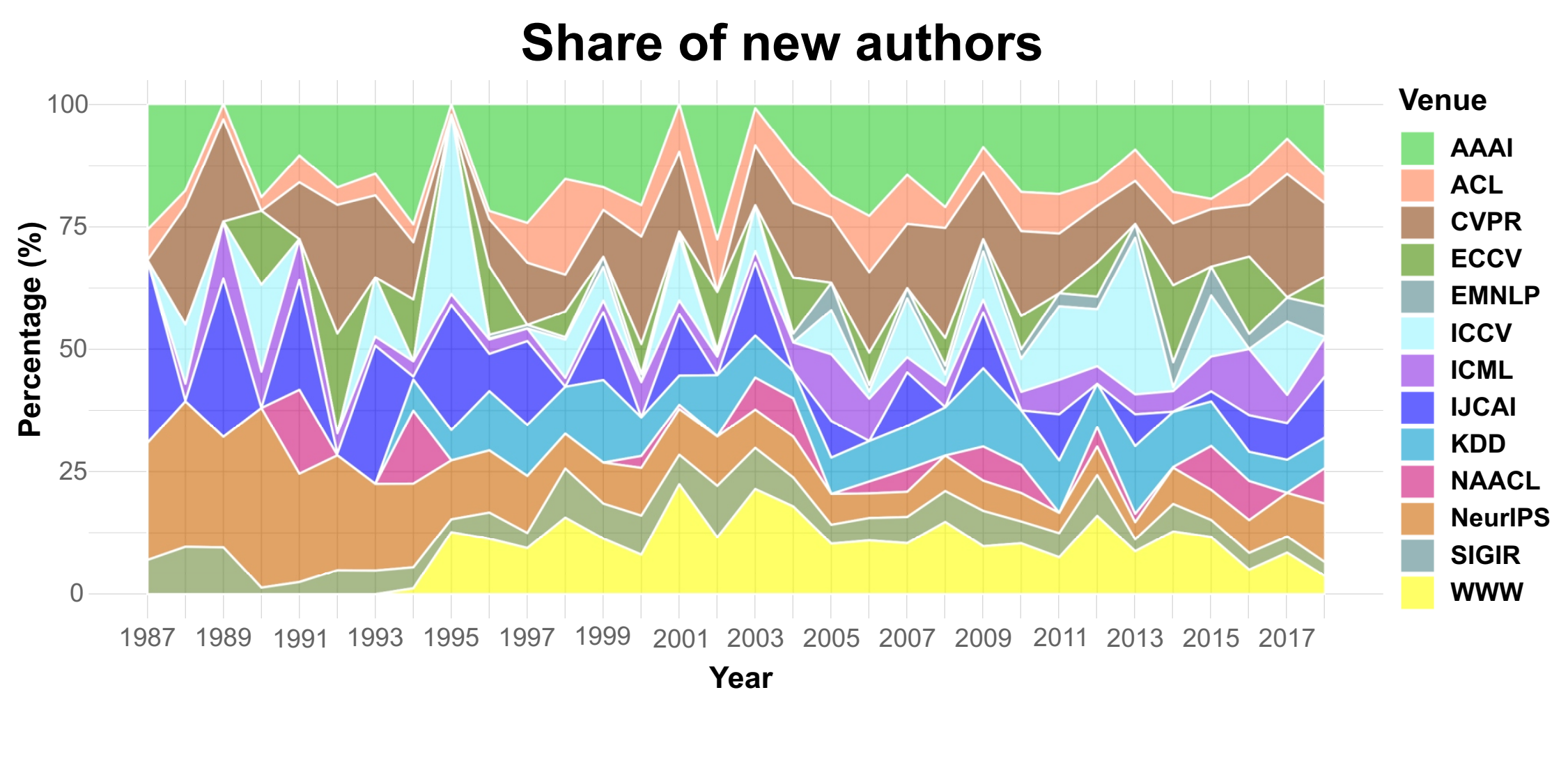

Figure 25: Share of yearly new authors per conference.

papers, these recent researches are more helpful in different areas.

Figure 27 shows the same graph data but ranked by their in-degree centrality, which simply measures how many citations a paper has received until a given year (inside our graph). The latest top 5 is composed of 3 papers related to computer vision and 2 to natural language processing. A similar pattern can be found until 1993, but back in 1985 and before most of the ranking was composed of papers that tackled reasoning, problem-solving, and symbolic learning, such as "Reasoning about knowledge and action" Moore [1977] (RAKAA77 in the figure), "A multi-level organization for problem-solving using many, diverse, cooperating sources of knowledge" Erman and Lesser [1975] (AMOFPSUMDCSOK75 in the figure) and "The art of artificial intelligence: themes and case studies of knowledge engineering" Feigenbaum [1977] (TAOAITACSOKE77 in the figure).

A very stable behavior can be seen in the PageRank ranking (Figure 28): most papers remained in the top 5 for 2 gaps (usually 8 years) and many of them for 3 gaps (16 years in the middle, 10 in the end). The paper "Towards automatic visual obstacle avoidance" Moravec [1977] (TAVOA77 in the figure) has been in the top 5 at least since 1993 and it has been leading the ranking since 2001. The second one, "Feature extraction from faces using deformable templates" Yuille et al. [1989] (FEFFUDT89 in the figure), is also a somewhat old paper related to computer vision.

Similarly, one can see the stableness and invariability to change that PageRank offers by noticing that we only had 20 different papers in the top 10 in the selected years, while we have 28 for Betweenness and Indegree, a more befitting number when compared to the figures shown in the previous sections.

| Author | Count |
|---|---|
| Lei Zhang | 488 |
| Luc Van Gool | 352 |
| Ming-Hsuan Yang | 350 |
| Thomas S. Huang | 322 |
| Andrew Y. Ng | 298 |
| Jiawei Han | 294 |
| Dacheng Tao | 282 |
| Yang Liu | 280 |
| Philip H. S. Torr | 280 |
| Yoshua Bengio | 248 |
| Wei Wang | 238 |
| Milind Tambe | 236 |
| Yang Li | 232 |
| Ale Leonardis | 224 |
| Liang Lin | 222 |
| Qingming Huang | 222 |
| Shuicheng Yan | 222 |
| Christos Faloutsos | 222 |
| Jiri Matas | 214 |
| Michael Felsberg | 212 |
| Horst Bischof | 212 |
| Philip S. Yu | 208 |
| Richard Bowden | 206 |

Table 4: The 23 authors who collaborated with more than 200 new authors since the year 1969

The remaining rankings (Closeness, Degree, and Out-degree) can be seen in Appendix E.

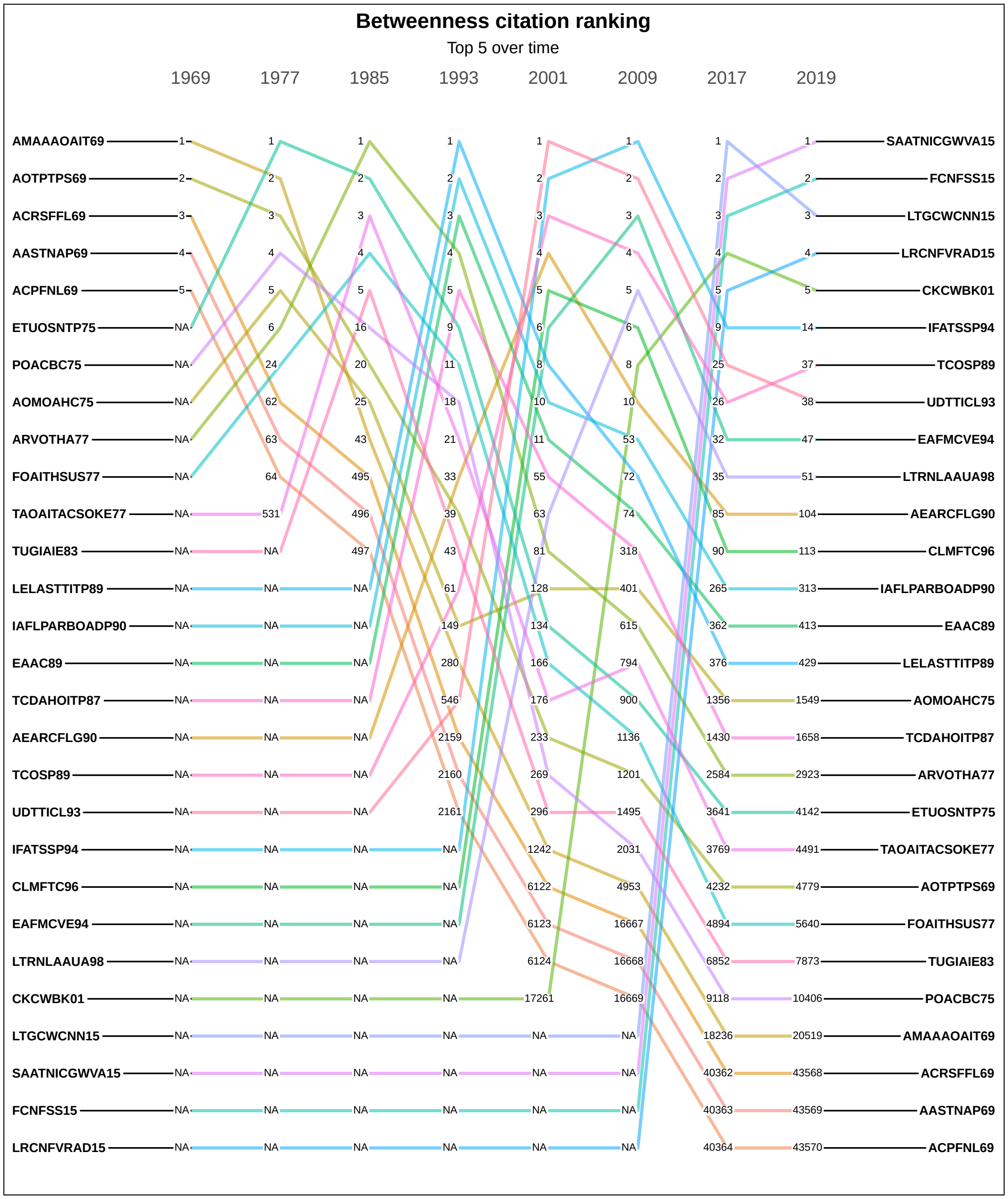

Figure 26: Papers citation ranking over time according to Betweenness centrality. *N/A stands for papers that had not been published in the selected venues until that year. Please refer Table 10 in the Appendix E to see the details of each ranked paper.*

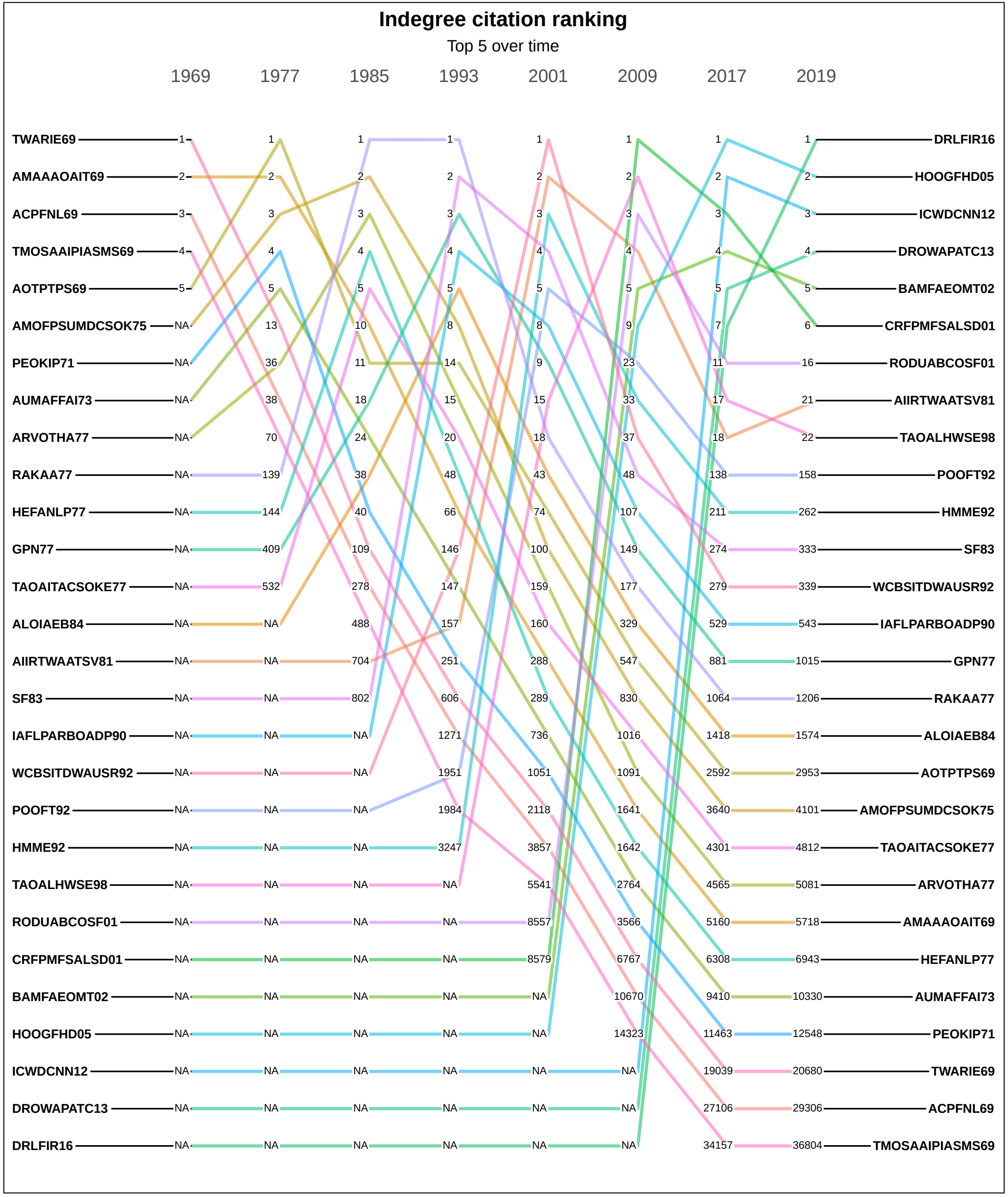

Figure 27: Papers citation ranking over time according to In-degree centrality. *N/A stands for papers that had not been published in the selected venues until that year. Please refer Table 10 in the Appendix E to see the details of each ranked paper.*

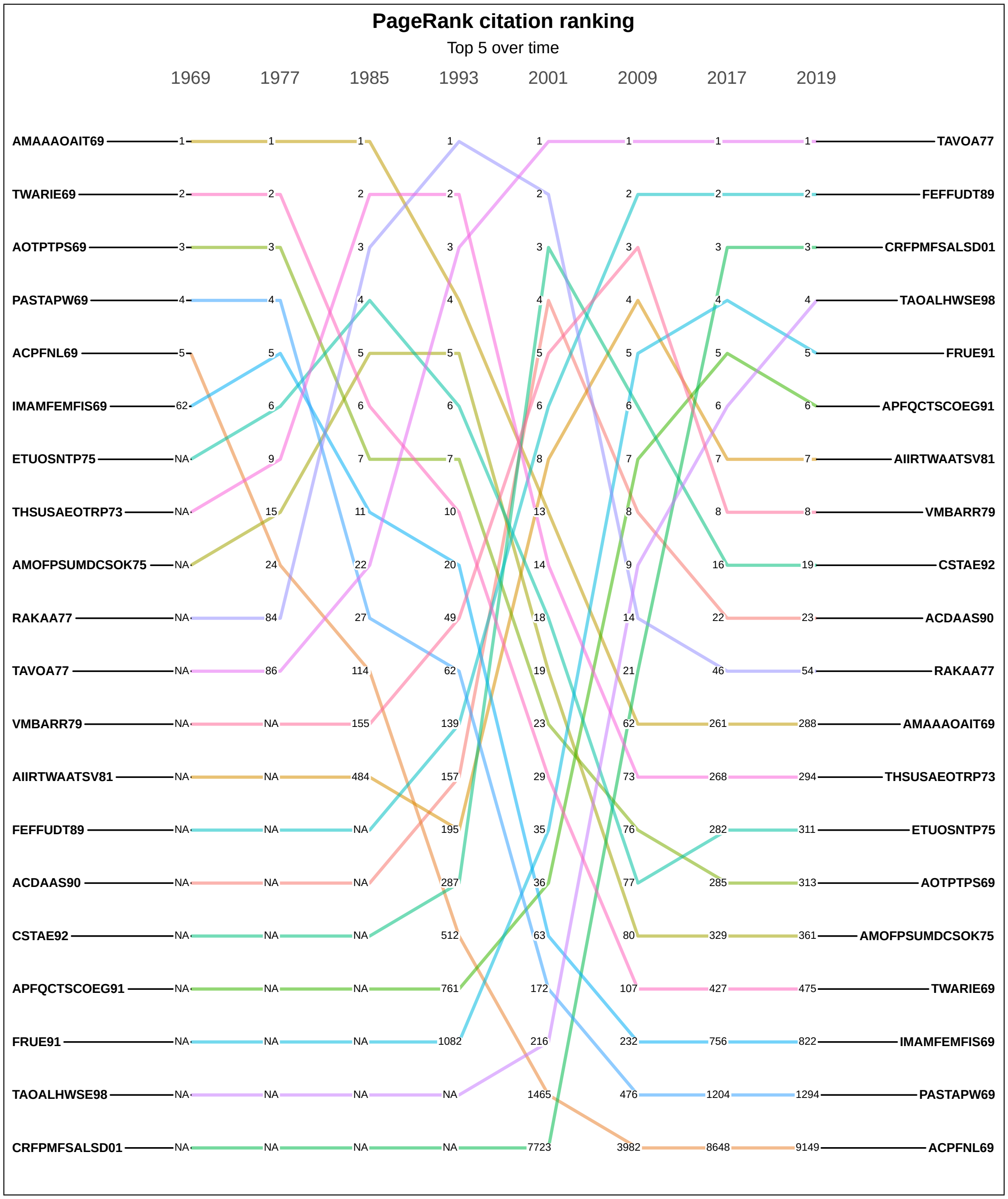

Figure 28: Papers citation ranking over time according to PageRank centrality. *N/A stands for papers that had not been published in the selected venues until that year. Please refer Table 10 in the Appendix E to see the details of each ranked paper.*

### 4.4.2 Share of Top 100 Ranking per Venue

Figures 29 to 31 reinforce the trend seen in the top 5 ranking in the last section: despite the centrality, computer vision-related venues are progressively gaining importance regarding their published papers, especially the CVPR. However, there has been a distinguished contribution by the ACL conference to the most important papers (according to PageRank) since 1984. These three heatmaps also show that the AAAI papers had their peak of importance during the late 1980s and the 1990s, but now they are losing their share of the ranking in the same fashion that IJCAI.

### 4.4.3 Share of Citations per Venue

We were also interested in how the citations of each venue have been evolving in the last few years. In this analysis, we were also able to distinguish citations to papers from arXiv, Journals, and the International Conference on Learning Representations (ICLR). For instance, back in the 1980s and early 1990s, around 50% of citations coming from NIPS papers were directed to papers from journals (see Figure 32), however, this share nowadays has been reduced to less than 25%. More than that, citations to ICLR papers and especially to arXiv papers have been increasing since the early 2010s.

A similar pattern occurs when we consider papers from AAAI and IJCAI, Figures 33 and 34 respectively. However, in their case, there is a much more divided share between all the conferences: it is possible to distinguish some influence from KDD, WWW, ACL, and EMNLP together with the increasing, yet unobtrusive, influence of arXiv and ICLR.

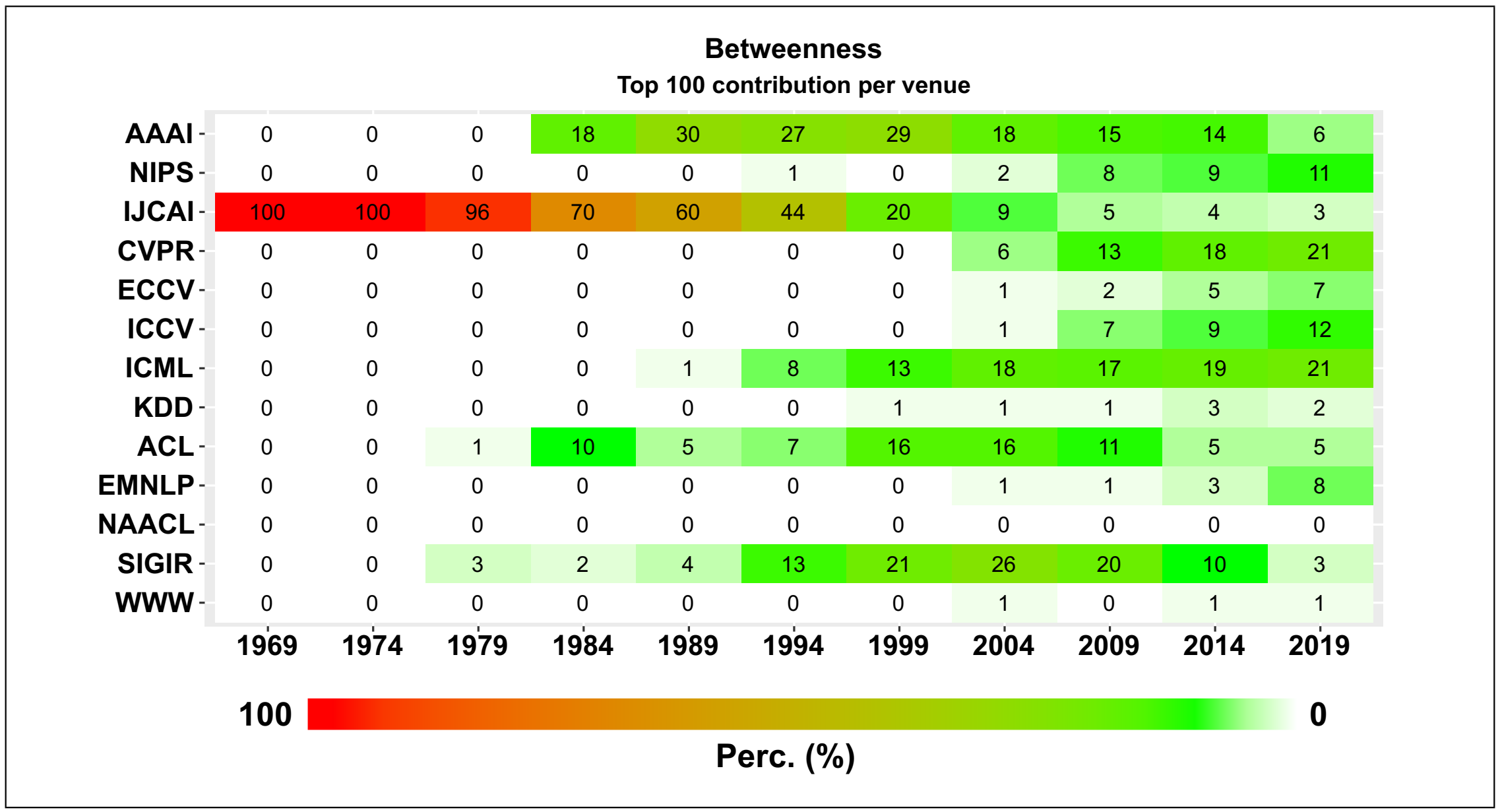

Figure 29: Venue contribution per year (accumulated) in the top 100 most important papers, according to Betweenness.

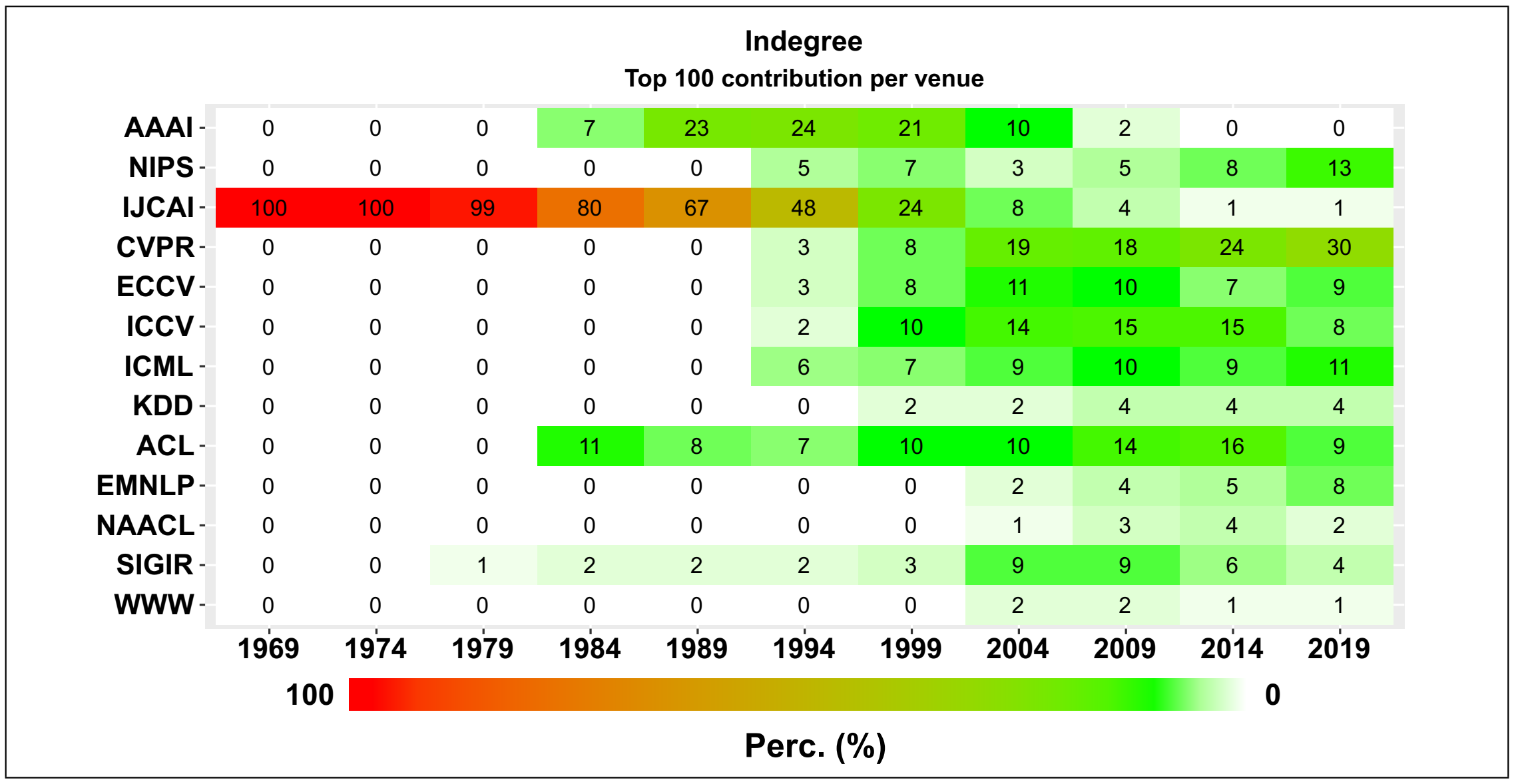

Figure 30: Venue contribution per year (accumulated) in the top 100 most important papers, according to In-Degree.

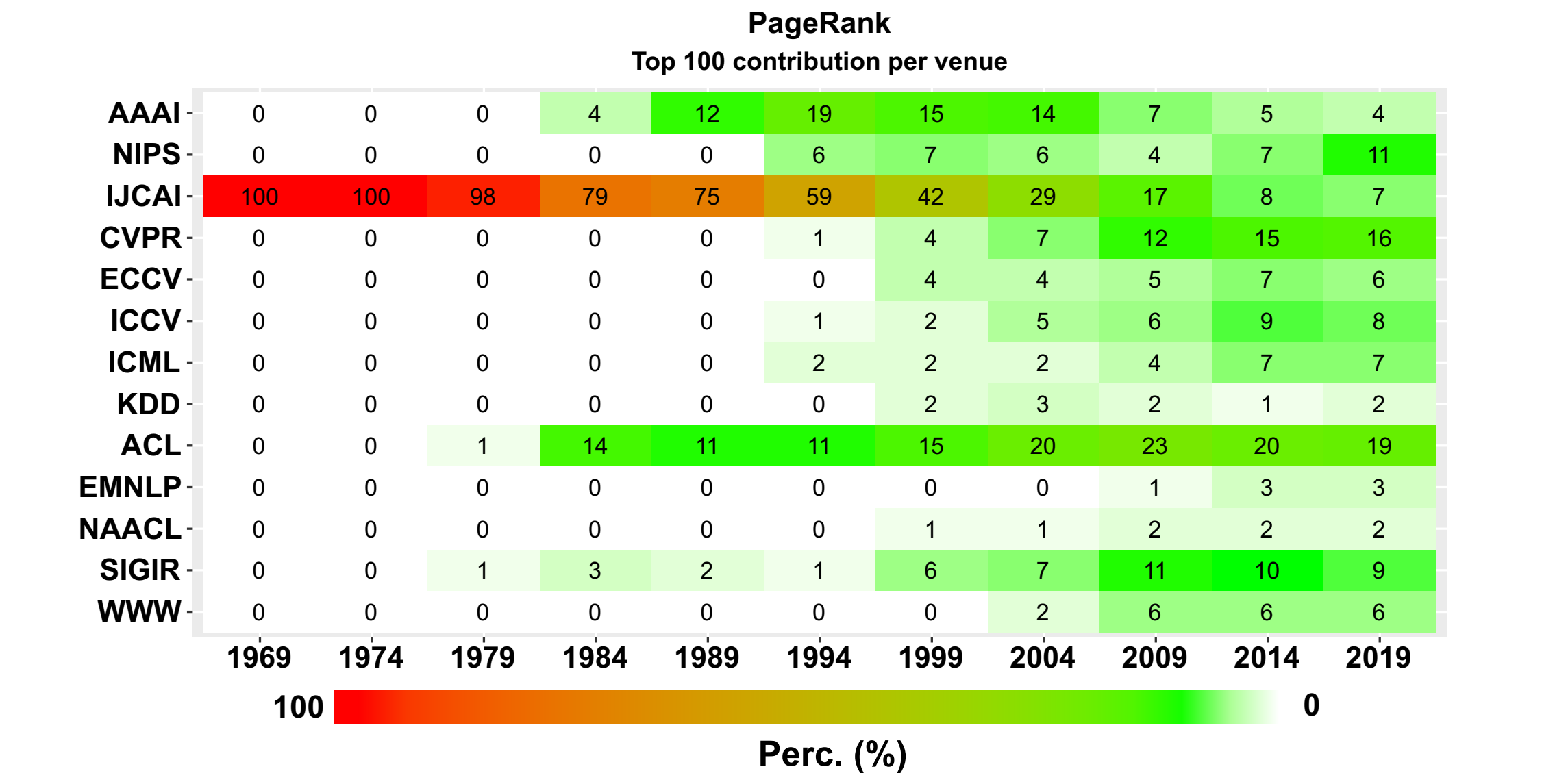

Figure 31: Venue contribution per year (accumulated) in the top 100 most important papers, according to PageRank.

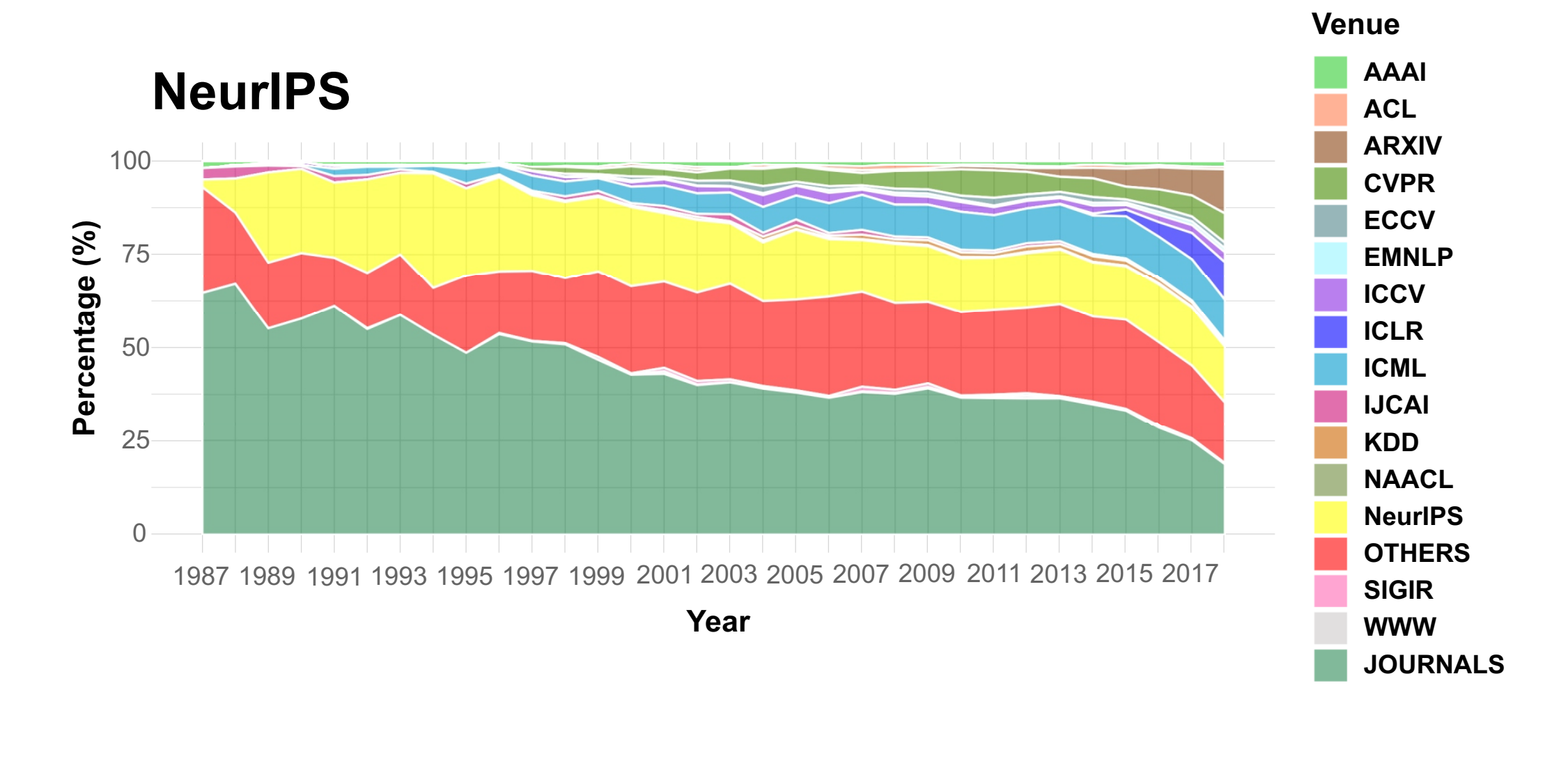

Figure 32: Where the citations coming from NeurIPS papers are pointing to share of each venue.

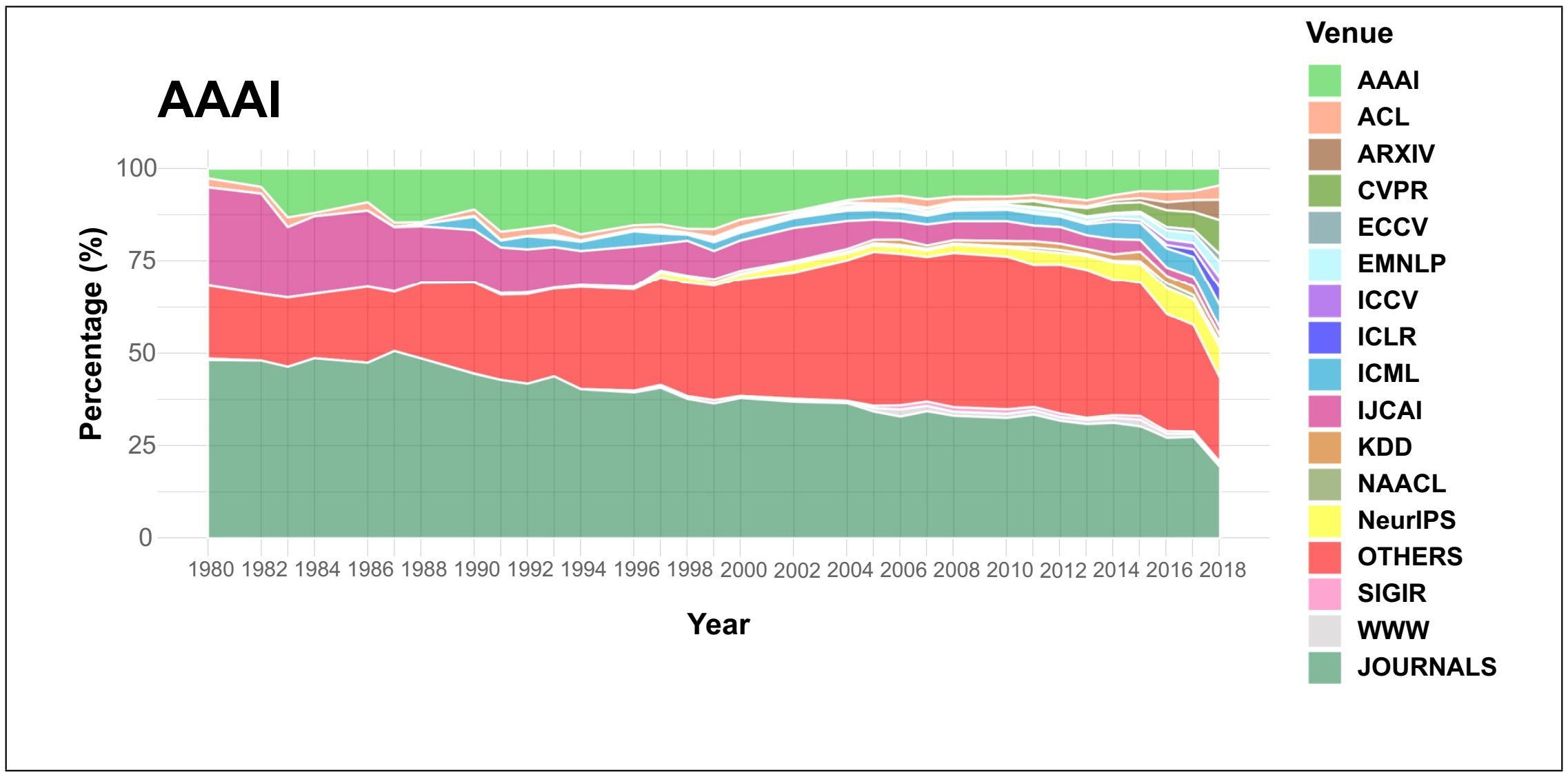

Figure 33: Where the citations coming from AAAI papers are pointing to share of each venue.

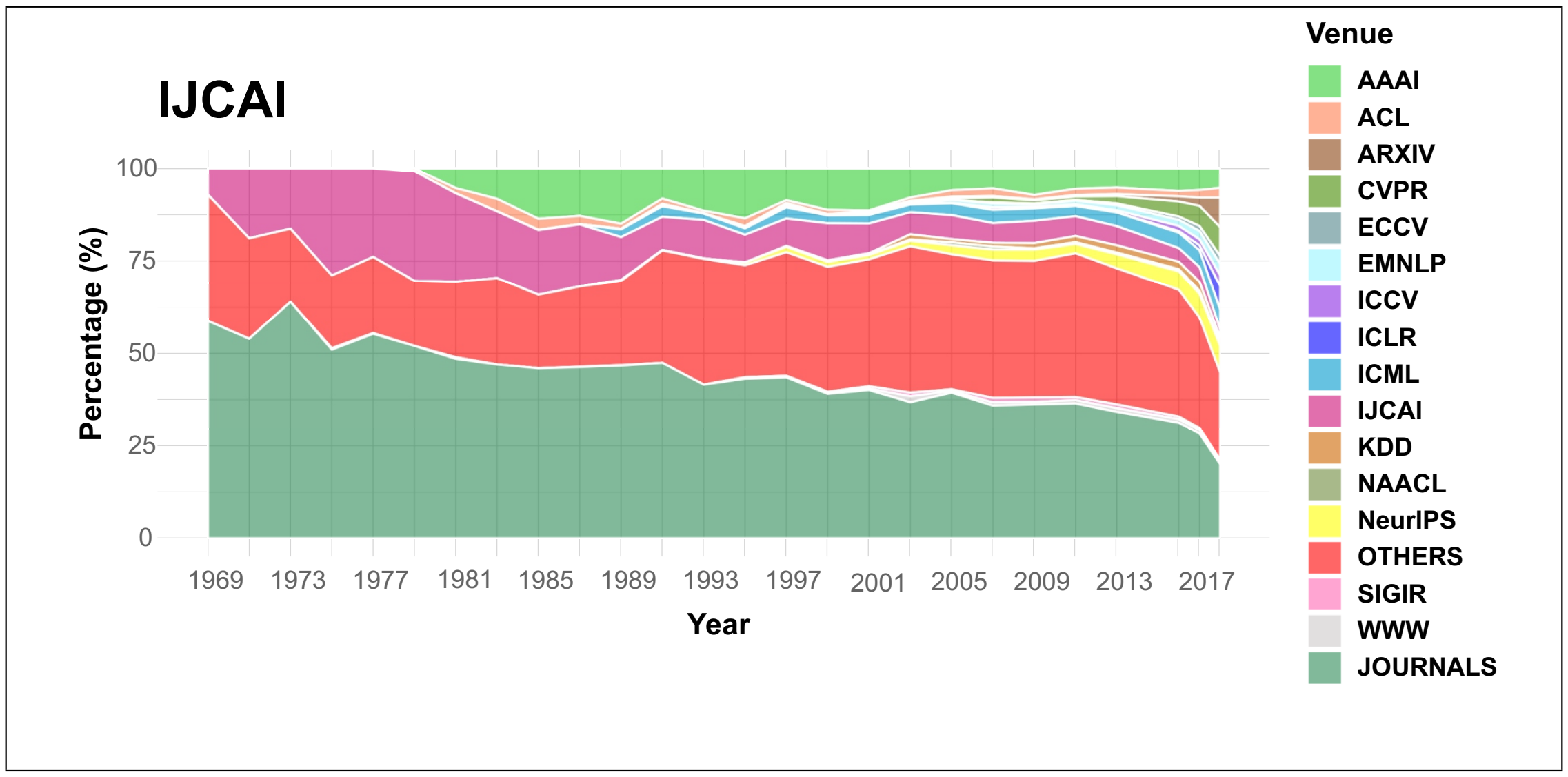

Figure 34: Where the citations coming from IJCAI papers are pointing to share of each venue.

### 4.5 Author-Paper Citation Graph

This graph was built to foster the research on recommender systems for papers – "given an author and his publication history, which are the most relevant papers he has not yet cited?"". In this sense, we have not developed any analysis on this graph and we expect to use it as a benchmark for future research on recommender systems: see Section 5.2.

### 4.6 Countries Citation Graph

This graph, computed for every country, is interesting because it has a different cardinality compared to the other graphs: while the others have tens of thousands of nodes, this graph only has 93 nodes with 4776703 edges.

Figure 35 shows the increase in the number of papers published at these conferences per year. The best-fitting line interpolates the data every 5 years. While the mid-1970s saw just a few countries participating in conferences (1974 and 1976 only had 2 countries: USA and United Kingdom, and USA and Canada, respectively) we have seen a large increase in countries participating in conferences in the last years, with 66 different countries having published in the conferences of interest in 2017. As mentioned above, in total authors from 93 different countries already published at these conferences.

Figure 36 shows a stacked percentage chart of the 15 countries with the most published papers. This data clearly shows the dominance of the United States in Artificial Intelligence research, with a slow increase in the number of papers published by authors in China.

Similarly, Figure 37 shows the same data but the pink bar at the bottom represents papers that we could not detect their country of affiliation.

Through this data representation, one can clearly see the years when IJCAI happened. Given the fact that IJCAI is commonly held outside the United States, and only in odd-numbered years, we can see a jagged-line pattern in the United States' share of papers, with a higher percentage in even-numbered years, and a lower percentage in odd-numbered years (when people from different countries have a higher chance of attending the conference, usually because of less strict visa requirements). For the same reason, after IJCAI started to be held annually (2013) the pattern disappeared. Figure 51 tries fixing this problem by creating a 2-year-wide sliding window and averaging the data before plotting it, creating a clearer view of the data.

An interesting outlier can be seen in 1979 when Japan had the highest share of published papers except for the USA. That happened exactly because IJCAI was held in Tokyo that year.

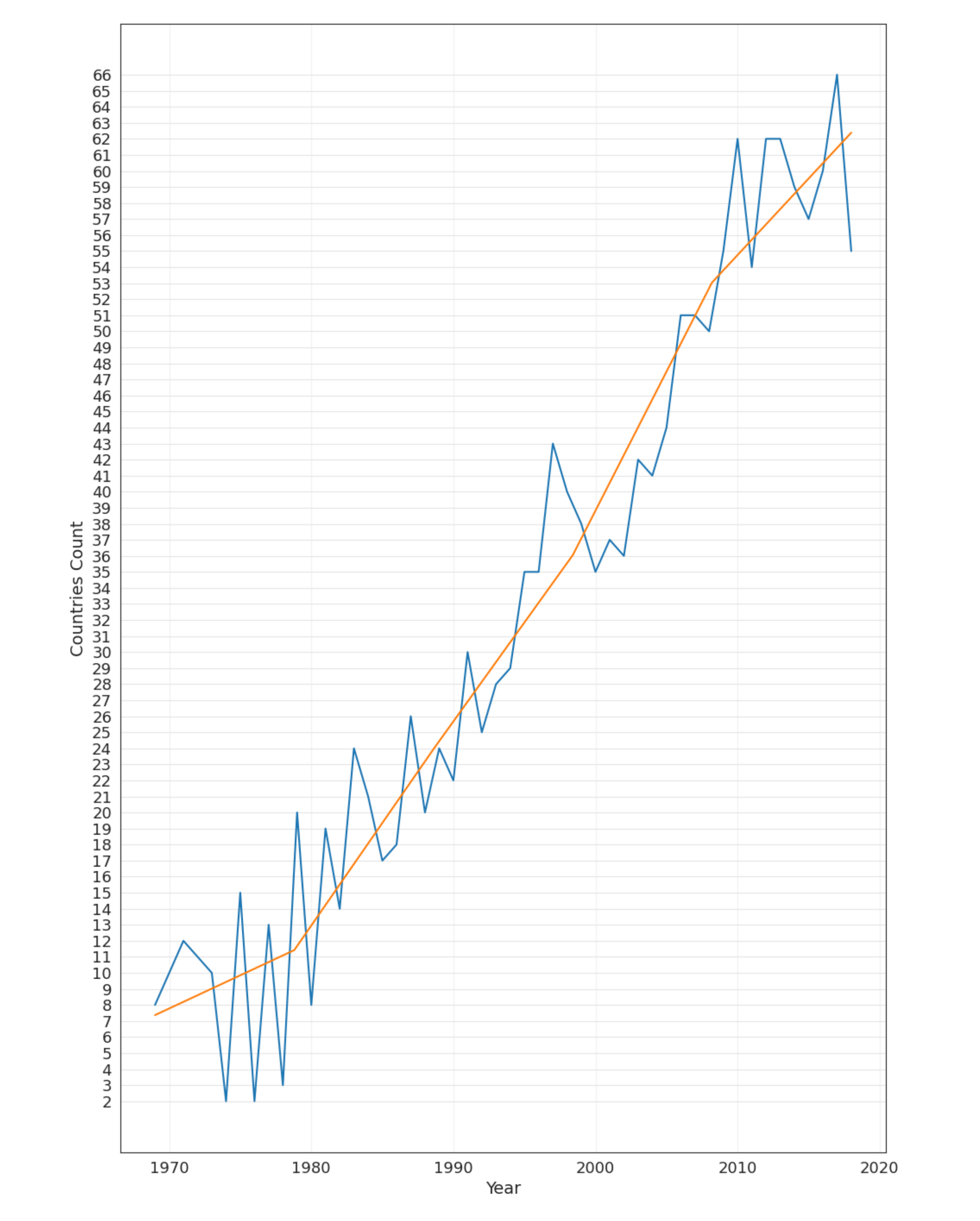

Figure 35: Number of Countries that published papers per year. The interpolating line is the best-fitting linear interpolation with 5 points

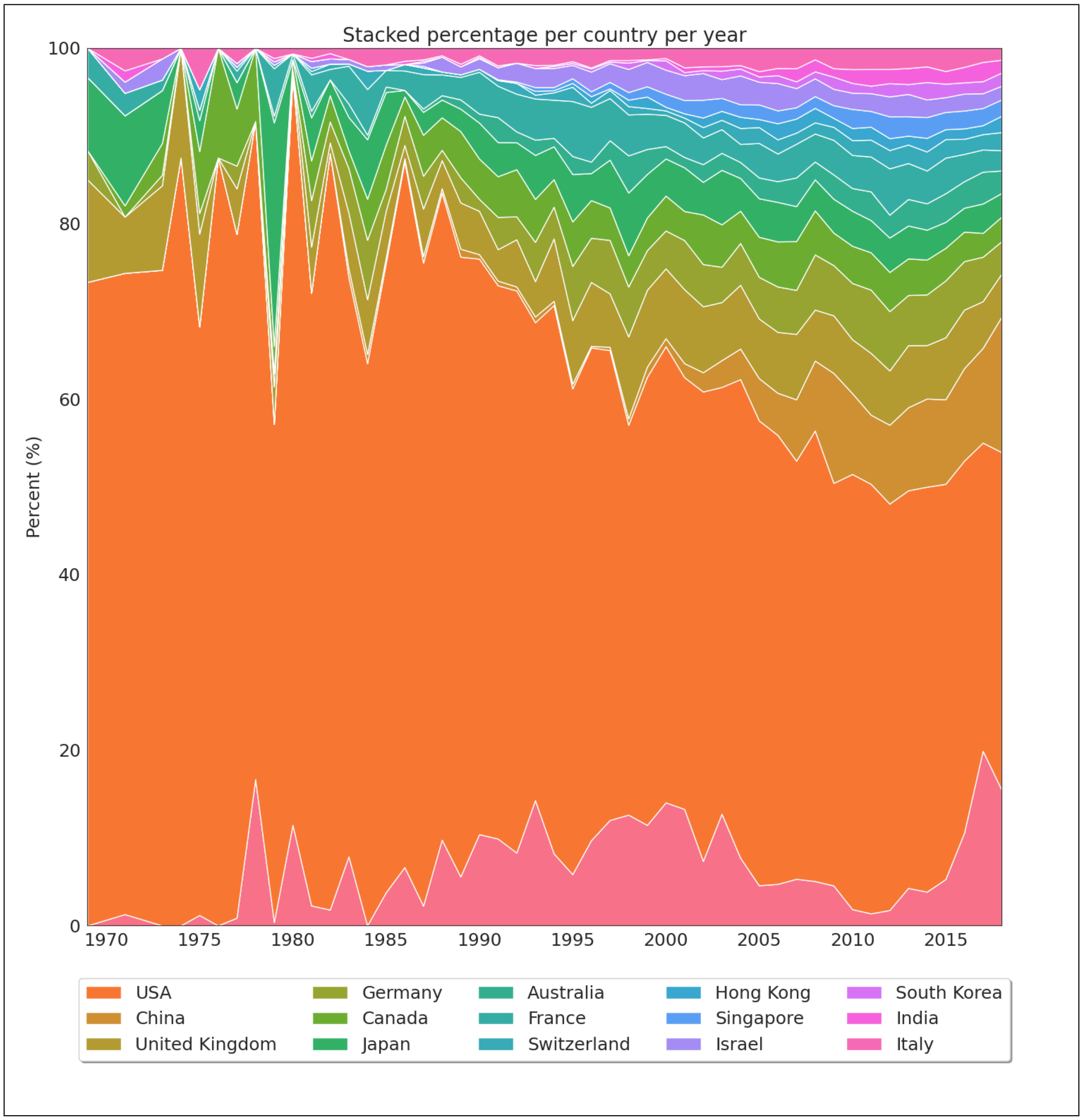

Figure 36: Stacked percentage of papers published per country per year including non-mapped ones. *The pink bar at the bottom indicates papers we could not identify which country they are from.*

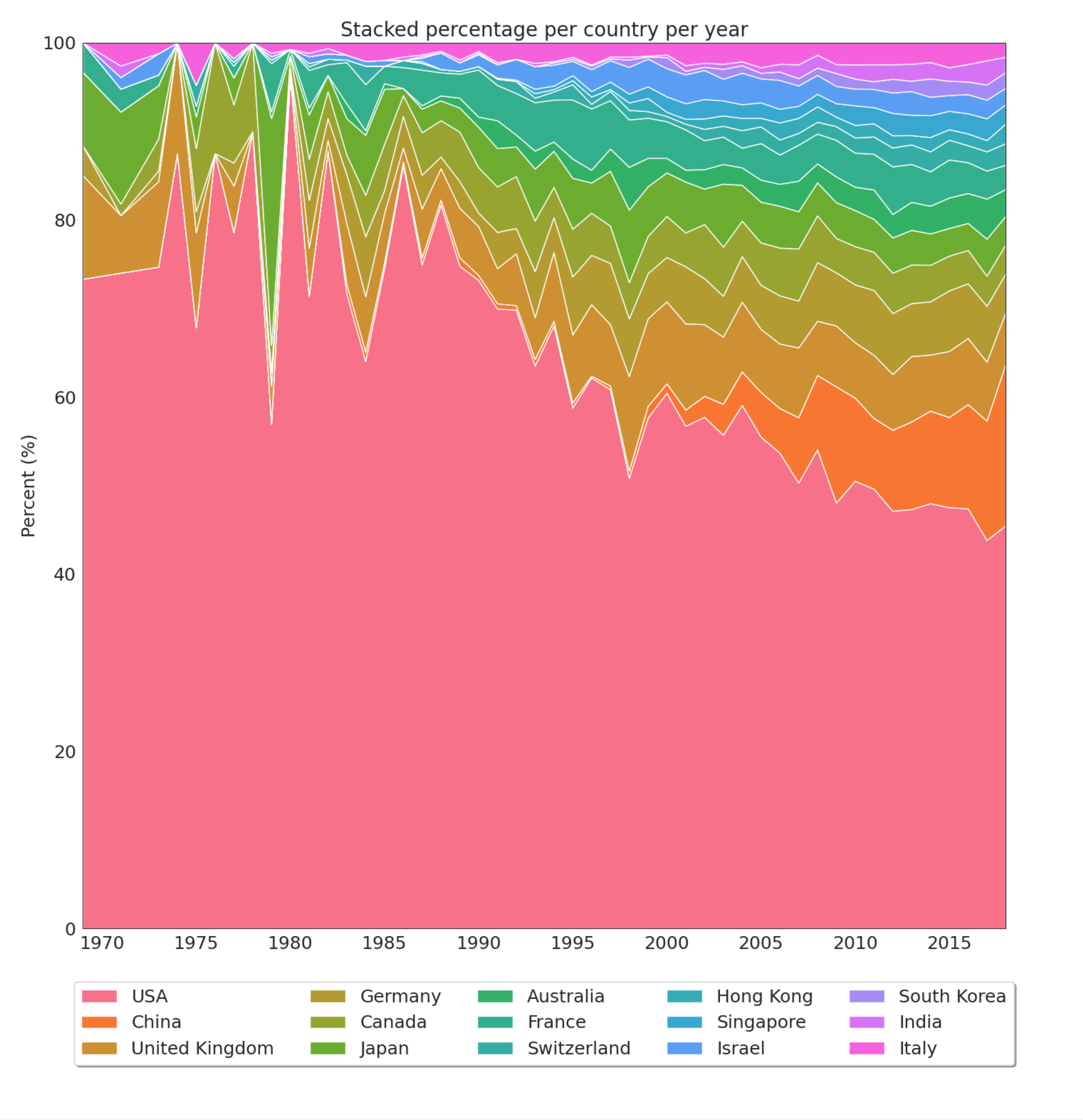

Figure 37: Stacked percentage of papers published per country per year, but not considering the ones we cannot identify.

Figure 38 shows the same data but with numbers in absolute terms instead of showing it with a stacked percentual. With it, we can see how the rate of acceptance for countries that are not the USA has grown faster than it has for the USA (the curve is steeper at the top). With it, we can also see the striking increase in papers accepted to these conferences in the last years, as already shown in numbers before.

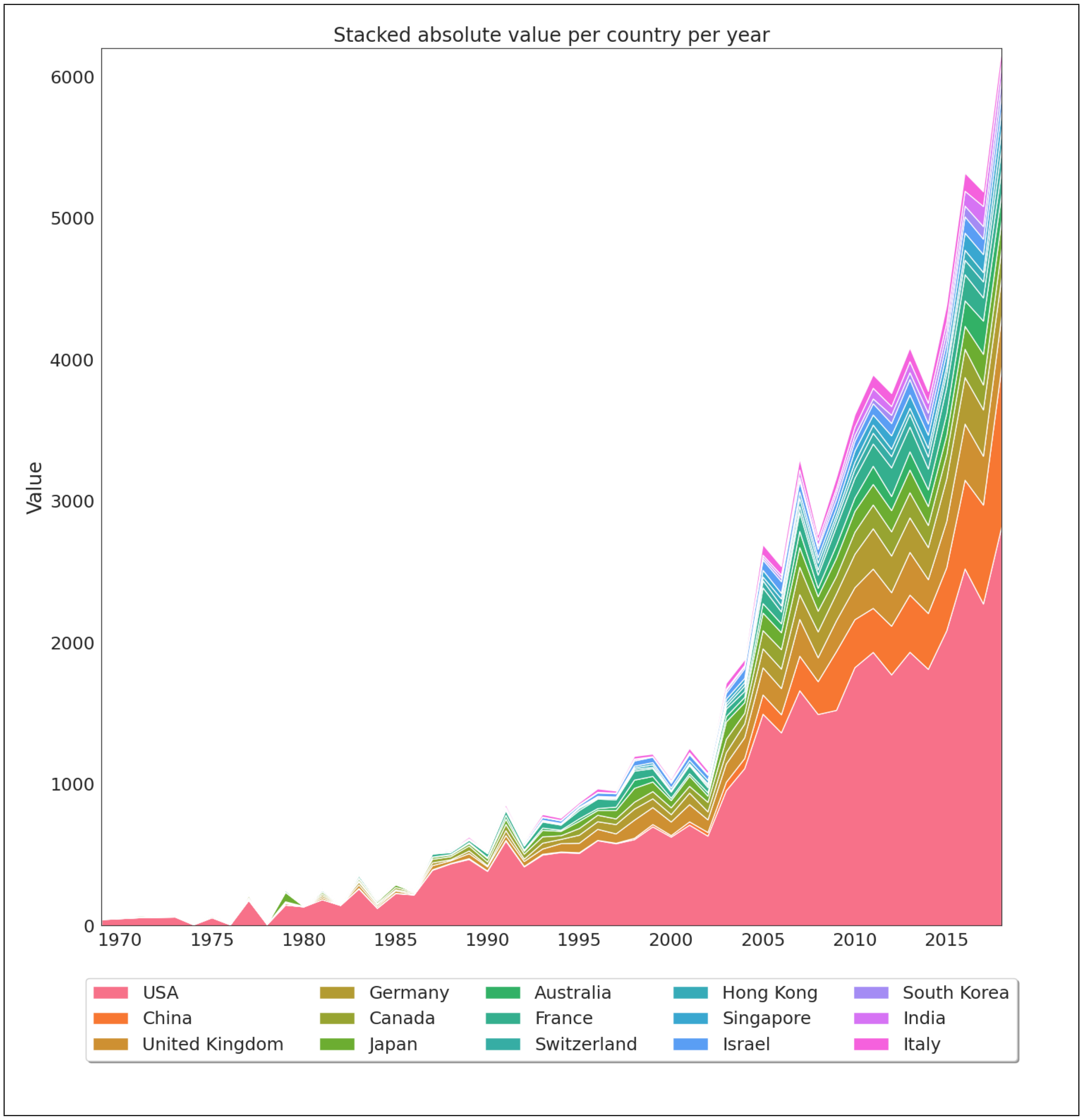

Figure 38: Quantity of papers per country per year

### 4.6.1 Analysis of More Recent Years

If we try using Arnet's v13 dataset to generate the above graphs, we can see in Figure 39 how much the data deteriorates. For 2019, we can only identify close to 5% of the paper country of affiliation, because for most of them, the "organization" field is empty. Note as well that even for the previous year the data is not as clear as it is in Figure 37, for instance.

We did not try applying any mapping similar to the one explained in Section 3.4 to this data because most of the non-identifiable organization fields are empty, as stated above.

## 4.7 Data Analysis of Turing Laureates

As previously stated, the Turing Award recognized seven researchers for their contributions to AI: Marvin Minsky (1969), John McCarthy (1971), Allen Newell and Herbert Simon (1975), Edward Feigenbaum and Raj Reddy (1994), Leslie Valiant (2010), Judea Pearl (2011) and Yoshua Bengio, Geoffrey Hinton and Yann LeCun (2018).

The Turing Award winners timeline (see Figure 40) depicts a change of focus of these highly prolific researchers over time: most recent awardees have their work divided into several venues (especially machine learning and computer vision-related ones, such as NIPS/NeurIPS and ICML and CVPR), while the older ones concentrated their efforts in AAAI or IJCAI. We also need to take into account that we are only considering conferences in this work, while most of the works published in the early days of Artificial Intelligence were published in other venues.

We also verified the Spearman correlation between the titles of papers published by Turing Award winners and the titles of papers published in the selected AI conferences (AAAI, IJCAI, and NIPS/NeurIPS) over time. To do so we compare the ranking of the TF-IDFs for the words in the Turing Award winner's paper titles in that year, related to the ranking of the TF-IDFs conference's (or group of conferences) papers titles in the same year. As the AI community, in general, has leaned towards the connectionist approach over the last years, we expected to see a decreasing trend regarding previous Turing Award winners who focused on symbolic AI and expert systems – or at least a very little correlation.

Nevertheless, the work of Marvin Minsky (1969 Turing Award laureate) is still quite in line with what is published in NIPS/NeurIPS, for instance, despite being poorly correlated with the three conferences when they are considered altogether (see Figure 41). These correlations may be however not very realistic since there are only two papers by Marvin Minsky in the entire dataset. The most positive

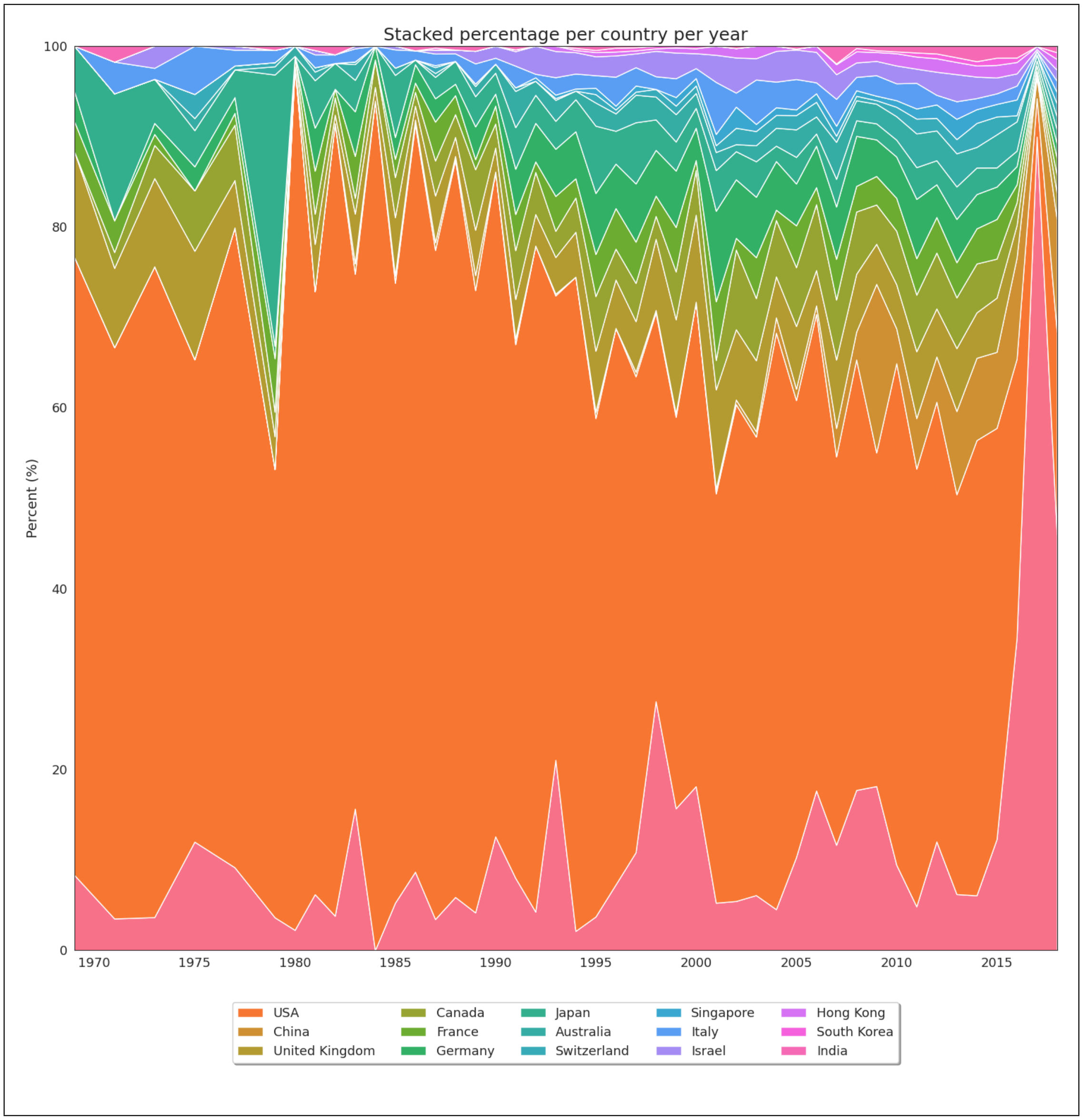

Figure 39: Deteriorated countries stacked chart with Arnet's V13

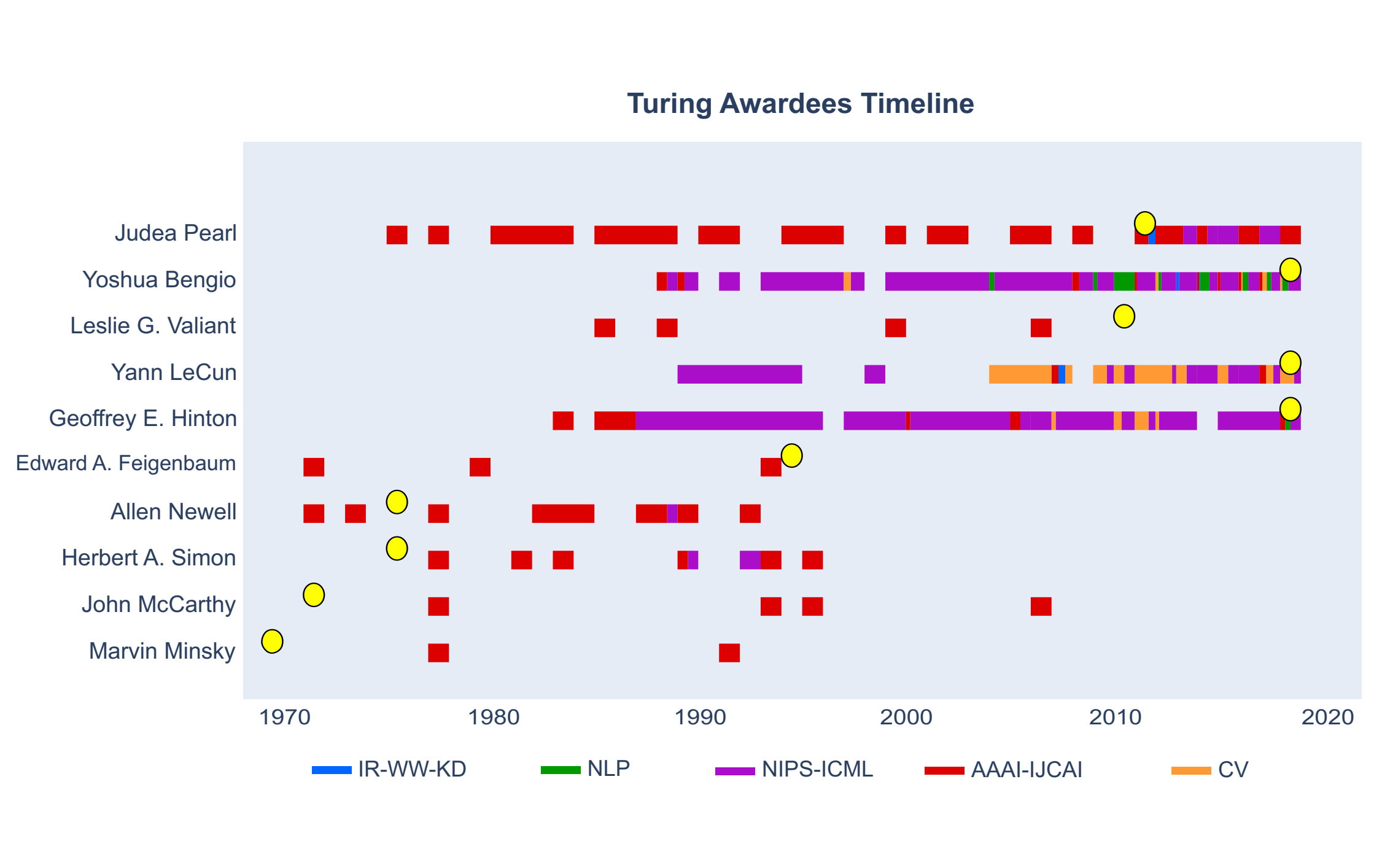

Figure 40: AI-related Turing Awardees timeline. *Each year is divided proportionally according to the number of papers published in each group of venues. Yellow ellipses are placed on the year each award was granted.*

and significant slope, however, comes from the work of the latest AI Turing Award laureates (Bengio, Hinton, and LeCun, 2018): their papers' titles have a positive and moderate correlation with all three conferences (and naturally with the average) and also show an increasing trend along the years, as depicted on Figure 42. The remaining plots can be found in Appendix G

Figure 42 clearly shows how the most recent AI Turing Awardees (Yoshua Bengio, Geoffrey Hinton, and Yann LeCun) influenced the area, with increasing rates of correlation over the years in all three main conferences from the Artificial Intelligence field. We predict that, in the next few years, if we were to plot the same data again, their correlation would likely have increased even further showing that they were able to influence AI research in general. We base our hypothesis on the fact that the 1969 Turing Award laureate Marvin Minsky, still has a positive correlation rate in some conferences such as NeurIPS, even though the same cannot be said for the AI field in general. Also, as noted in Section 2, Minsky possibly influenced the future

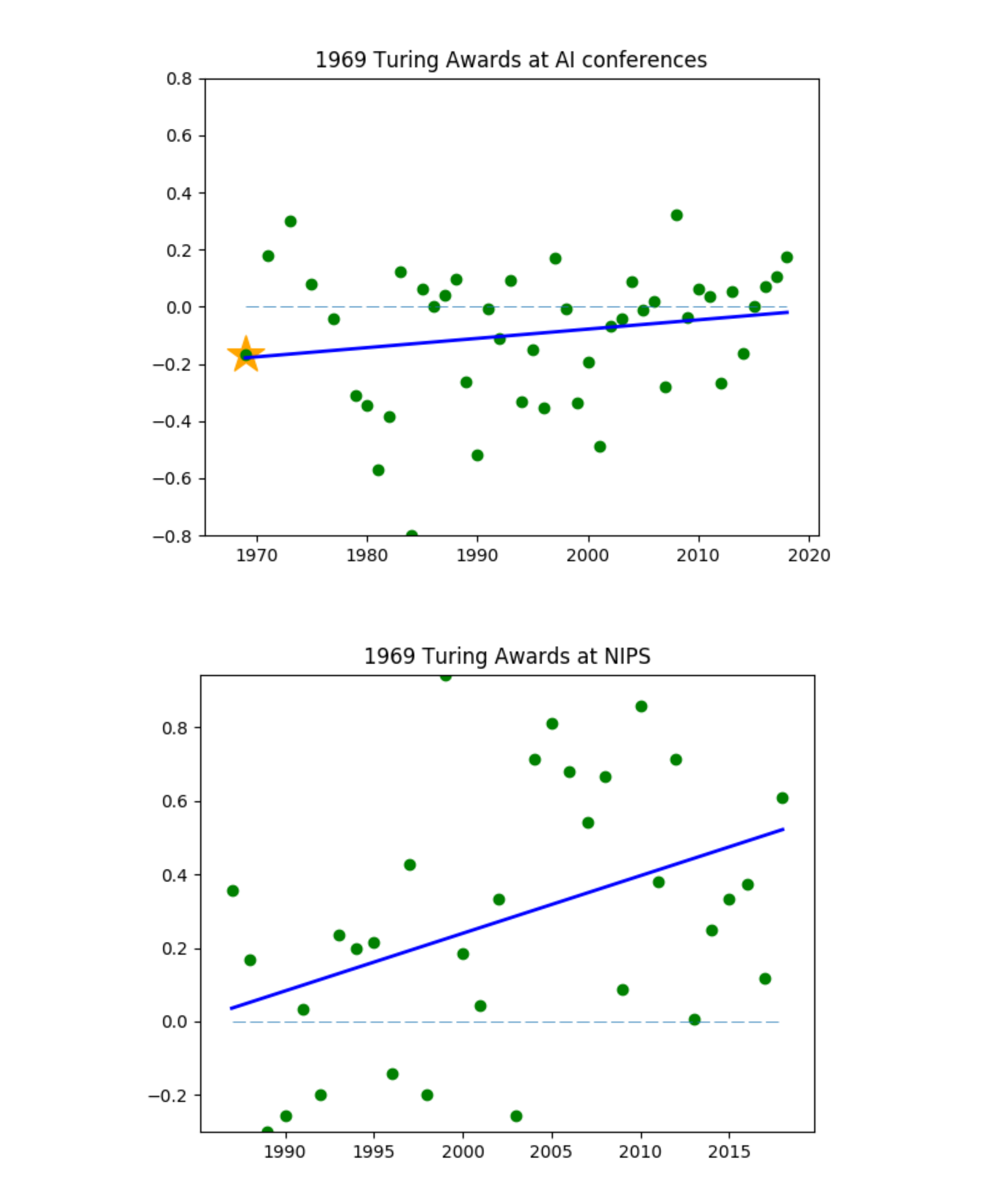

Figure 41: 1969 Turing Award Correlation with AI conferences and NIPS specifically

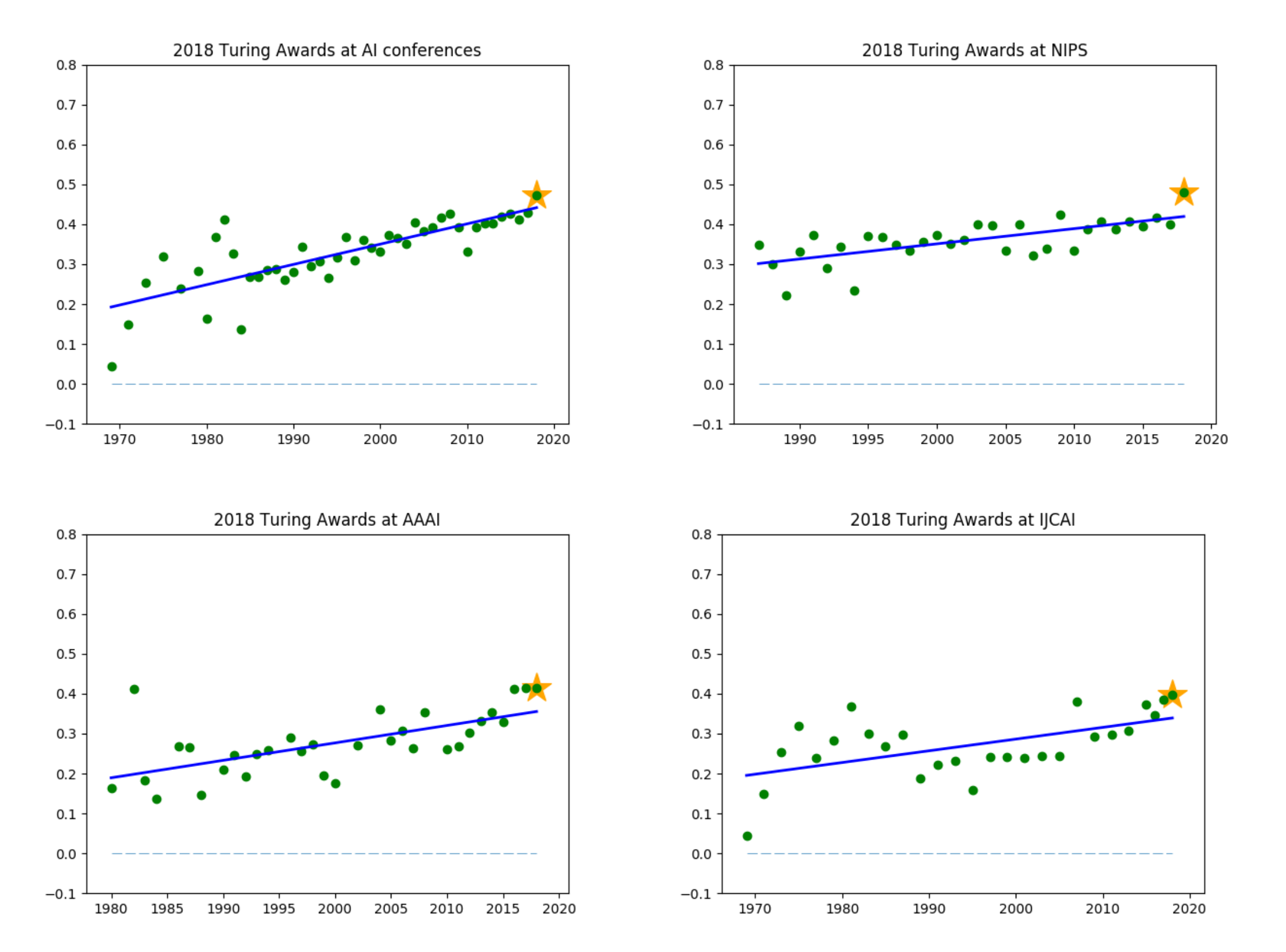

Figure 42: Correlation between titles of papers published by 2018 Turing Award winners and titles of papers published in the three AI flagship conferences.

of AI after publishing Minsky and Papert [1969].

Similarly, this data can also be seen from the opposite side, if we consider the 2018 laureates: they were closely following the trend of papers published in these conferences, therefore winning the Turing Award by researching the areas of interest. Even if possible, it is clear that their works are relevant and influenced the area in ways not influenced by others.

Also, it is worth noting that the Turing Award laureates do not appear in the ranking of authors according to the centrality measures in Section 3.3.1 and 4.3. This is probably because the Turing Awardees *have not* published a large number of papers in the venues analysed to be able to reach the top of the rankings, and also because their contributions are mostly based on some seminal, highly influential works.

# 5 Conclusions and Further Work

Artificial Intelligence research has accomplished much since Turing [1936], McCulloch and Pitts [1943], Turing [1950]. AI is now amply used in industry to power large high-tech corporations. Some of the processes behind this evolution are still not well understood this work aimed to make AI development, history, and evolution clear. We presented a short survey on AI history, describing the periods or seasons AI has already gone through. We have also included a quick survey on graph centrality measures, the Turing Award winners, and the flagship AI-related conferences, which is necessary to understand the overall picture of the area better.

By analyzing Arnet's v11 dataset, a dataset based on DBLP's corpus, and enhancing it to a graph-based format, we intended to ease paper/author citation/collaboration network research. This dataset generated insightful graphs and could generate even more in future works. Also, the code presented in this work makes it such that it is relatively easy to extend it to any other underlying dataset, making it possible to generate and compute the same statistics presented in this work for any other area than AI. These graphs show insights on self-citations, new authors, and author and paper importance throughout the years. We also proposed a new type of dataset intended to be used as a knowledge graph source for recommender systems, where authors, papers, citations, and collaborations are all defined in the same graph. With the Country Citation Graph, we also introduced an important dataset and pipeline capable of inferring the country of affiliation of an author based on its organization.

By investigating the Turing Award winners and comparing them against the published data, we find out that there is evidence that they actually "pull" their most published venues to their topic of research, at least for the most recent AI researchers winners. Finally, the study on countries' affiliation is, to the best of our knowledge, the first of its type, creating a new algorithm able to infer the country of affiliation of an author from their organization, as available at DBLP or Arnet.

## 5.1 Contributions

This work has the following specific contributions:

- Five new graph-based datasets, with fully computed centralities to ease paper/author, citation/collaboration network research.
- Analyses for these graphs, focusing both on their raw structure and the centrality rankings.

- Algorithms description, allowing anyone to replicate the graph building process in any programming language for any dataset.
- Spearman Correlation computation between Turing Award laureate papers and conference papers, showing they have a positive correlation.

Besides the theoretical contributions we also present a few important software contributions. They can be found in more detail in Appendix H, but we outline them as follows:

- Python library to convert an XML to JSON in a stream fashion, i.e. without loading the whole XML and JSON files in memory
- Parallel Python implementation for the Betweenness and Closeness Centralities
- Novel Python implementation for a Graph Parsing pipeline, avoiding duplicate work through data caching
- Python implementation of the proposed algorithm to infer a paper country of affiliation

## 5.2 Future Work

The dataset created in this work provides several possibilities for future work, especially when we think about the computed centralities. Our work presented several analyses of the dataset, and one might think of even more possible ways to visualize it. Additionally, it would be ideal if Coreness centrality (Section 2.4.6) was also computed for this dataset, as it displays the interesting feature of being a discrete value instead of a continuous one, thus allowing you to more easily identify the most important authors/papers according to it.

The Country Citation Graph has a lot of potential in understanding "brain-draining" by investigating the flow of authors from one country of affiliation to others – easily done with our dataset, without any extra work besides counting the number of "transitions" between countries. Similarly, we believe that comparing more advanced usages of this dataset with the Turing Awardees might bring even more interesting results. With a better dataset, one where there are abstract data available for every paper, one might be able to achieve better results when running a Spearman Correlation (Section 4.7) between the text in Turing Award winners' abstracts and the ones from the remaining venues papers.

**Acknowledgements**

This work is partly supported by CNPq and CAPES, Brazil, Finance Code 001.

# A Arnet Dataset

This section presents some extra data in the Arnet Dataset, which did not fit in the main work.

Tables 5 , 6, and 7 display the inner objects we referenced in Table 2.

Table 8 shows a manual count of papers published at AAAI, NeurIPS and IJCAI. This was built to support Figure 5 and bring the point that the v13 dataset did not have even half the data present at these conferences. In this table, a question mark ("?") indicates that we could not infer the number of papers for that conference in that year.

To aggregate the data present in this table, we used AAAI's website statistics, the NeurIPS API, and the IJCAI Proceedings page. The harder to get this data was IJCAI as they do not provide a paper account, only containing links to every paper published every year. Therefore we built a Javascript snippet that counted the number of links in those pages. Ultimately, we could not use this script in the years 1979 and 2001 because the available data format does not allow one to do such an analysis.

Figure 43 shows an example of data present in the Arnet dataset, and used throughout our work when we needed an example.

*"*" indicates the field was used in this work*
*"?" indicates the field is optional*

| Field Name | Type | Description |
|---|---|---|
| id* | *string* | Unique ID for the author - unique across papers |
| name* | *string* | Full author name |
| org?* | *string* | Organization this author was in when of this paper |

Table 5: Data structure for an *Author* entry in the Arnet JSON dataset

*"*" indicates the field was used in this work*
*"?" indicates the field is optional*

| Field Name | Type | Description |
|---|---|---|
| id* | *string* | Unique ID for the venue - unique across papers |
| raw* | *string* | The raw name of the venue |
| name? | *string* | Humand readable name of the venue |

Table 6: Data structure for a *Venue* entry in the Arnet JSON dataset

*"*" indicates the field was used in this work*

| Field Name | Type | Description |
|---|---|---|
| IndexLength | *integer* | How many words in the abstract |
| InvertedIndex* | *hash<string, integer[]>* | Inverted index with the position of every word in the paper abstract |

Table 7: Data structure for a *IndexedAbstract* entry in the Arnet JSON dataset

In Table 8, data for AAAI was extracted from their API; for NeurIPS it was extracted from their official statistics website; and for IJCAI it was manually (using JavaScript) counted on their website. Cells with a "?" text indicate the years we were not able to find an accurate count of papers for that conference, reinforcing the fact that this is a lower-bound estimate.

Table 8: Manual count of papers per main AI conference per year

| | AAAI | NeurIPS | IJCAI |
|---|---|---|---|
| 1969 | 0 | 0 | 63 |
| 1970 | 0 | 0 | 0 |
| 1971 | 0 | 0 | 58 |
| 1972 | 0 | 0 | 0 |
| 1973 | 0 | 0 | 77 |
| 1974 | 0 | 0 | 0 |
| 1975 | 0 | 0 | 141 |
| 1976 | 0 | 0 | 0 |
| 1977 | 0 | 0 | 200 |
| 1978 | 0 | 0 | 0 |
| 1979 | 0 | 0 | ? |
| 1980 | ? | 0 | 0 |
| 1981 | 0 | 0 | 106 |
| 1982 | ? | 0 | 0 |
| 1983 | ? | 0 | 233 |
| 1984 | ? | 0 | 0 |
| 1985 | 0 | 0 | 257 |
| 1986 | ? | 0 | 0 |
| 1987 | ? | 90 | 301 |
| 1988 | ? | 94 | 0 |

**Table 8 continued from previous page**

| | AAAI | NeurIPS | IJCAI |
|---|---|---|---|
| 1989 | 0 | 101 | 270 |
| 1990 | 0 | 143 | 0 |
| 1991 | *?* | 144 | 190 |
| 1992 | *?* | 127 | 0 |
| 1993 | *?* | 158 | 137 |
| 1994 | 341 | 140 | 0 |
| 1995 | 0 | 152 | 275 |
| 1996 | 336 | 152 | 0 |
| 1997 | 268 | 150 | 183 |
| 1998 | 269 | 151 | 0 |
| 1999 | 235 | 150 | 203 |
| 2000 | 265 | 152 | 0 |
| 2001 | 0 | 197 | *?* |
| 2002 | 256 | 207 | 0 |
| 2003 | 0 | 198 | 297 |
| 2004 | 250 | 207 | 0 |
| 2005 | 530 | 207 | 340 |
| 2006 | 718 | 204 | 0 |
| 2007 | 702 | 207 | 480 |
| 2008 | 648 | 250 | 0 |
| 2009 | 0 | 262 | 331 |
| 2010 | 780 | 292 | 0 |
| 2011 | 743 | 306 | 494 |
| 2012 | 707 | 370 | 0 |
| 2013 | 720 | 360 | 484 |
| 2014 | 912 | 411 | 0 |
| 2015 | 1101 | 403 | 656 |
| 2016 | 1163 | 569 | 658 |
| 2017 | 1049 | 679 | 782 |
| 2018 | 1201 | 1009 | 871 |
| 2019 | 1150 | 1428 | 965 |
| 2020 | 1591 | 1898 | 779 |
| 2021 | 1692 | 2334 | 722 |
| **Total** | 17627 | 13902 | 10553 |

```json
{
  "id": "1533861849",
  "title": "Understanding the difficulty of training deep
feedforward neural networks",
  "authors": [
    {
      "name": "Xavier Glorot",
      "id": "295353625"
    },
    {
      "name": "Yoshua Bengio",
      "id": "161269817"
    }
  ],
  "venue": {
    "raw": "international conference on artificial intelligence and
statistics",
    "id": "2622962978"
  },
  "year": 2010,
  "n_citation": 11,
  "page_start": "249",
  "page_end": "256",
  "doc_type": "Conference",
  "publisher": "",
  "volume": "",
  "issue": "",
  "references": [
    "1529808766",
    "1994197834",
    // [...]
    "2172174689"
  ],
  "indexed_abstract": {
    "IndexLength": 598,
    "InvertedIndex": {
      "Whereas": [0],
      "before": [1],
      "2006": [2],
      "it": [3, 452, 576, 593],
      // [...]
      "drastic": [596],
      "impact).": [597]
    }
  },
  "fos": [
    {"name": "Machine learning", "w": 0.45906812},
    {"name": "Gradient descent", "w": 0.454203367},
    // [...]
    {"name": "Artificial intelligence", "w": 0.0}
  ]
}
```

Figure 43: Example of a JSON entry for Glorot and Bengio [2010] in the Arnet dataset *[...] indicates some items in the array were abbreviated for sake of brevity.*

# B Codebase

In this section, we present Algorithm 4 used in Section 3.4 along Algorithm 3 to be able to infer a country of origin from an organization. This basically removes the clutter present in Arnet's data.

Table 9 shows every open-source library used to develop this work. We are very thankful for every library contributor's work to the open-source community.

| Library Name | Usage |
|---|---|
| click | Create CLI to run experiments with different parameters |
| fire | Create CLI to run experiments with different parameters |
| matplotlib | Plot the charts |
| networkx | Build the graph datasets |
| nltk | Tokenize words and detect stop words |
| numpy | Manipulate data arrays in a vector-fashion |
| scipy | Compute Spearman Correlations |
| seaborn | Improve matplotlib's plots look |
| sklearn | Generate linear models and compute TF-IDF |
| tqdm | Generate progress bars for long data processing pipelines |

Table 9: Python libraries used in this work

**Algorithm 4** Organization Name Cleaning Preprocessing

```
Require: org                                          ▷ Organization name
  org ← split(org, ",")          ▷ Split the text in every comma, turning it into a list
  org ← org[-1]                                       ▷ Last item in the array
  org ← replace(org, "#TAB#", "")                     ▷ Remove unknown tag
  org ← replace(org, "#tab#", "")                     ▷ Remove unknown tag
  org ← replace(org, /[\(\)\[\]\-_]/, "")             ▷ Regex-based replacement
      return org
```

# C Author Citation

The charts presented in this section are related to centralities from Section 3.3.1.

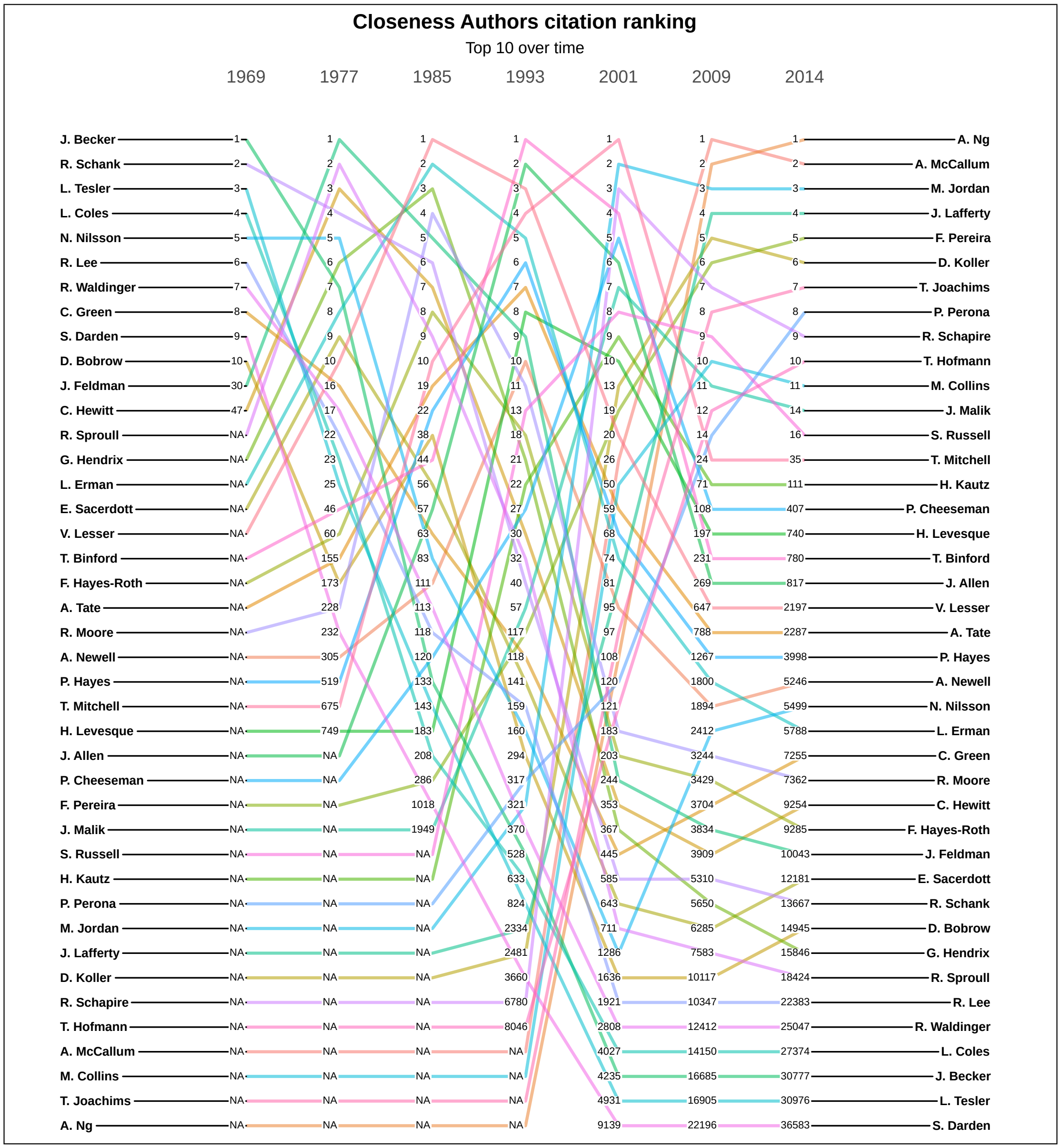

Figure 44: Authors citation ranking over time according to Closeness centrality. Figure refers to Appendix C

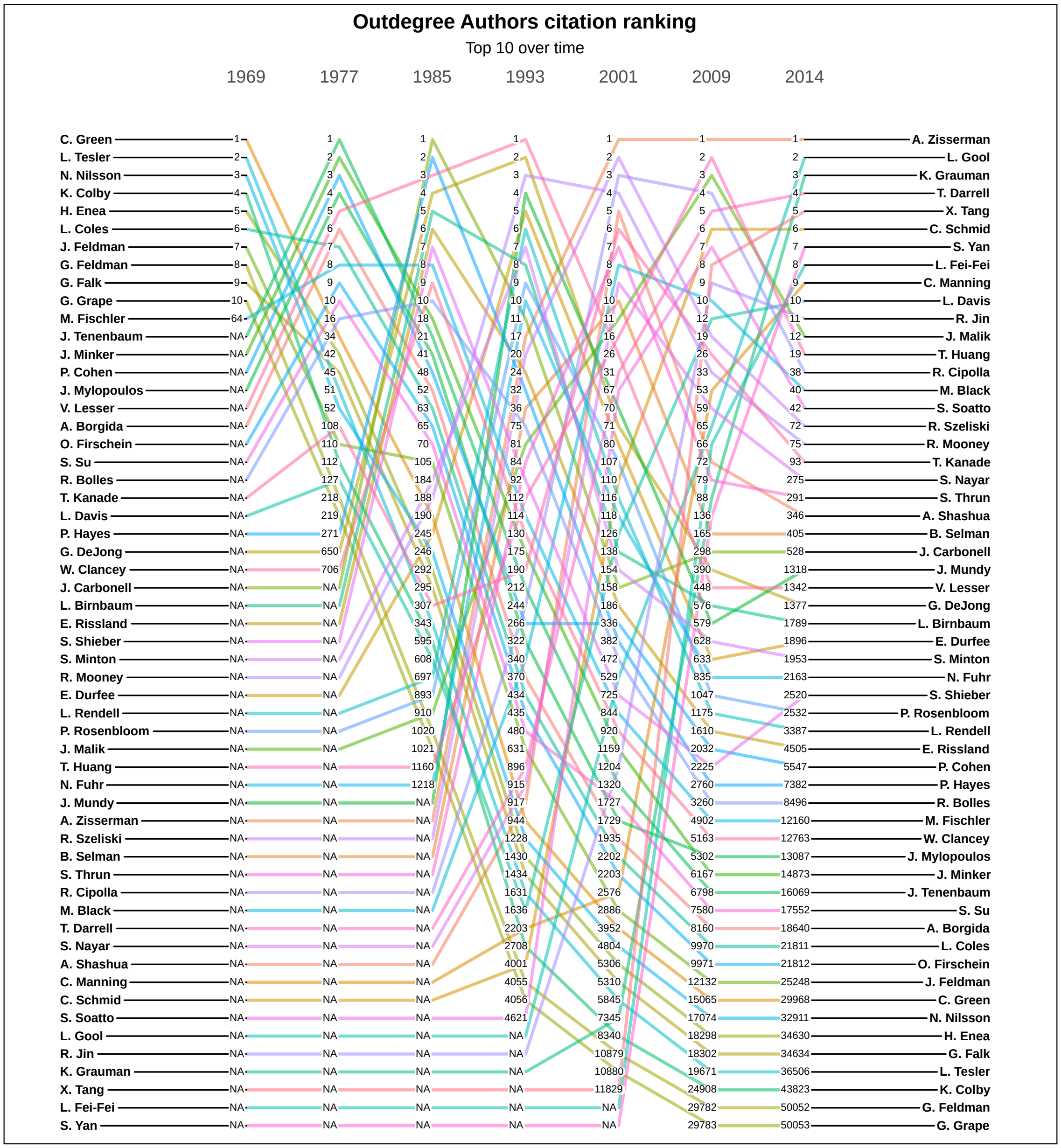

Figure 45: Authors citation ranking over time according to Out-degree centrality. Figure refers to Appendix C

# D Author Collaboration

The charts presented in this section are related to centralities from Section 4.3.

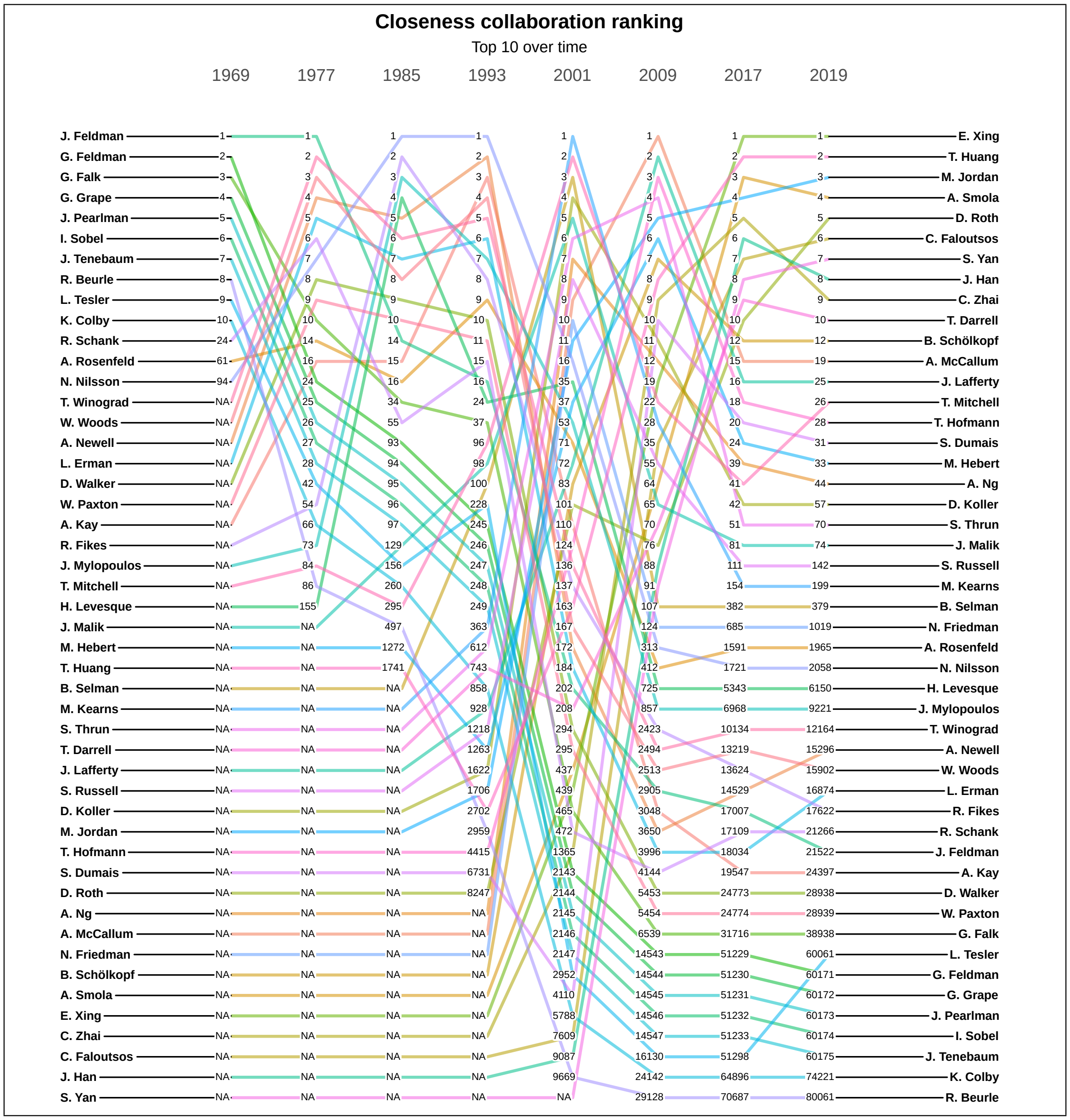

Figure 46: Authors collaboration ranking over time according to Closeness centrality. Figure refers to Appendix D

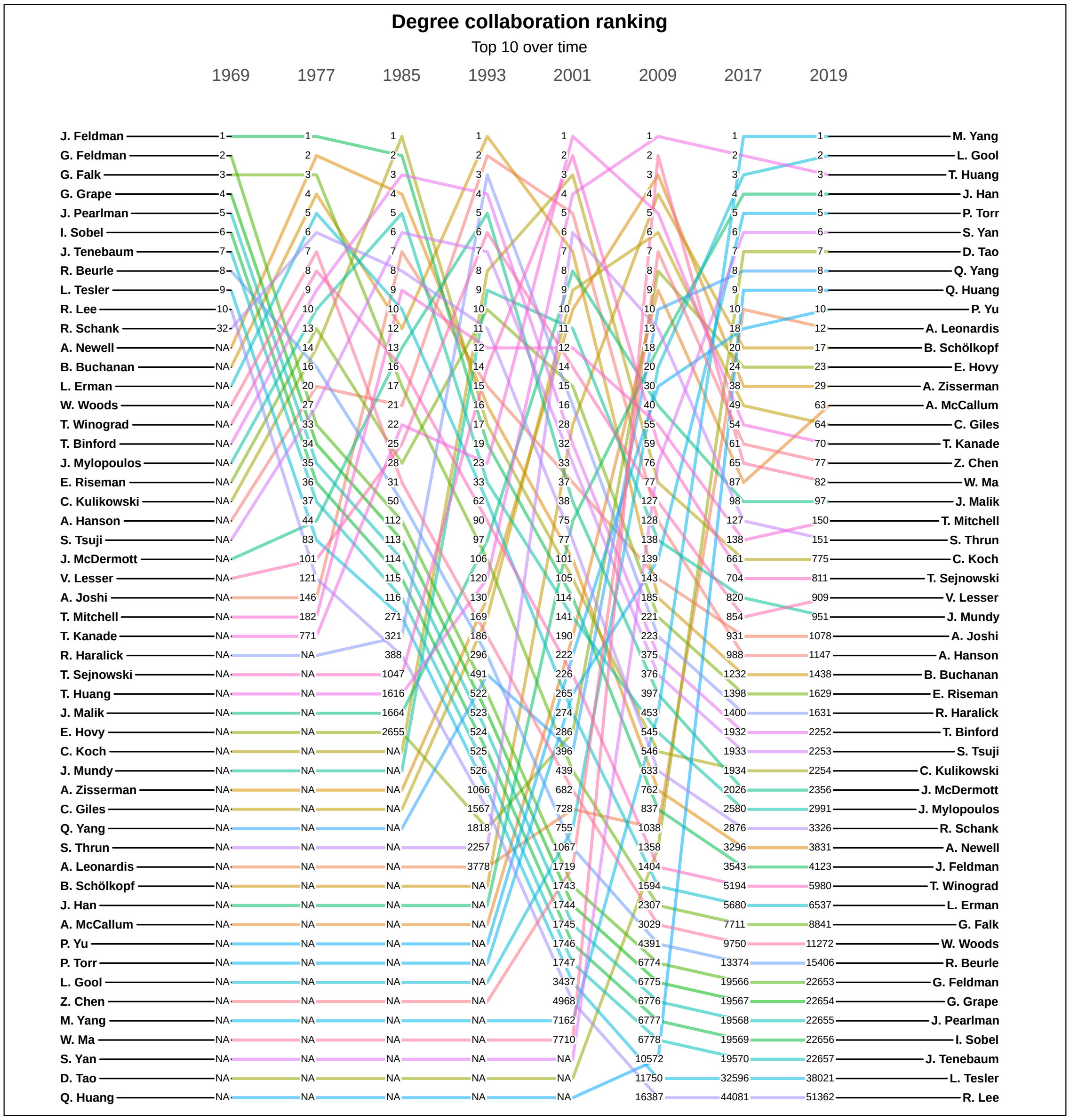

Figure 47: Authors collaboration ranking over time according to In-Degree centrality. Figure refers to Appendix D

# E Paper Citation

The charts presented in this section are related to centralities from Section 4.4. Similarly, Table 10 is used to map every paper entry in these charts to a paper title and year.

Table 10: Dictionary for the papers which appeared in the Top 5 rankings

| **Initials** | **Title** | **Year** | **Venue** |
|---|---|---|---|
| AAAOSL92 | An asymptotic analysis of speedup learning | 1992 | ICML |
| AALR85 | AI and legal reasoning | 1985 | IJCAI |
| AAOLTPAASP92 | An analysis of learning to plan as a search problem | 1992 | ICML |
| AASTNAP69 | An augmented state transition network analysis procedure | 1969 | IJCAI |
| ACDAAS90 | Accurate corner detection: an analytical study | 1990 | ICCV |
| ACPFNL69 | A conceptual parser for natural language | 1969 | IJCAI |
| ACRSFFL69 | A contextual recognition system for formal languages | 1969 | IJCAI |
| ACSOICI86 | A case study of incremental concept induction | 1986 | AAAI |
| AEARCFLG90 | AUTOMATICALLY EXTRACTING AND REPRESENTING COLLOCATIONS FOR LANGUAGE GENERATION | 1990 | ACL |
| AIIFIR83 | Artificial intelligence implications for information retrieval | 1983 | SIGIR |
| AIIRTWAATSV81 | An iterative image registration technique with an application to stereo vision | 1981 | IJCAI |
| ALOIAEB84 | A logic of implicit and explicit belief | 1984 | AAAI |
| AMAAAOAIT69 | A mobius automation: an application of artificial intelligence techniques | 1969 | IJCAI |
| AMEMFPT96 | A Maximum Entropy Model for Part-Of-Speech Tagging | 1996 | EMNLP |

*Continued on next page*

Table 10 – *Continued from previous page*

| Initials | Title | Year | Venue |
|---|---|---|---|
| AMOFPSUMDCSOK75 | A multi-level organization for problem solving using many, diverse, cooperating sources of knowledge | 1975 | IJCAI |
| ANSFSISDAR71 | A net structure for semantic information storage, deducation and retrieval | 1971 | IJCAI |
| AOMOAHC75 | Acquisition of moving objects and hand-eye coordination | 1975 | IJCAI |
| AOTPTPS69 | Application of theorem proving to problem solving | 1969 | IJCAI |
| APFFSC01 | A probabilistic framework for space carving | 2001 | ICCV |
| APFQCTSCOEG91 | A Procedure for Quantitatively Comparing the Syntactic Coverage of English Grammars | 1991 | NAACL |
| ARVOTHA77 | A retrospective view of the Hearsay-II architecture | 1977 | IJCAI |
| ASNAAMOHM73 | Active semantic networks as a model of human memory | 1973 | IJCAI |
| AUMAFFAI73 | A universal modular ACTOR formalism for artificial intelligence | 1973 | IJCAI |
| BAMFAEOMT02 | Bleu: a Method for Automatic Evaluation of Machine Translation | 2002 | ACL |
| BBOWEMDAFIIR08 | Beyond bags of words: effectively modeling dependence and features in information retrieval | 2008 | SIGIR |
| BTMANEFTEOVM97 | Boosting the margin: A new explanation for the effectiveness of voting methods | 1997 | ICML |

*Continued on next page*

Table 10 – *Continued from previous page*

| **Initials** | **Title** | **Year** | **Venue** |
|---|---|---|---|
| CCALMFIA11 | Combining concepts and language models for information access | 2011 | SIGIR |
| CKCWBK01 | Constrained K-means Clustering with Background Knowledge | 2001 | ICML |
| CLMFTC96 | Context-sensitive learning methods for text categorization | 1996 | SIGIR |
| CPS84 | Classification problem solving | 1984 | AAAI |
| CRFPMFSALSD01 | Conditional Random Fields: Probabilistic Models for Segmenting and Labeling Sequence Data | 2001 | ICML |
| CSS73 | Case structure systems | 1973 | IJCAI |
| CSTAE92 | Camera Self-Calibration: Theory and Experiments | 1992 | ECCV |
| DCOEW93 | DISTRIBUTIONAL CLUSTERING OF ENGLISH WORDS | 1993 | ACL |
| DRLFIR16 | Deep Residual Learning for Image Recognition | 2016 | CVPR |
| DROWAPATC13 | Distributed Representations of Words and Phrases and their Compositionality | 2013 | NIPS |
| EAAC89 | Execution architectures and compilation | 1989 | IJCAI |
| EAFMCVE94 | Efficient algorithms for minimizing cross validation error | 1994 | ICML |
| ETUOSNTP75 | Expanding the utility of semantic networks through partitioning | 1975 | IJCAI |
| EWANBA96 | Experiments with a new boosting algorithm | 1996 | ICML |

*Continued on next page*

Table 10 – *Continued from previous page*

| Initials | Title | Year | Venue |
|---|---|---|---|
| EWASAFTDBOAHBS69 | Experiments with a search algorithm for the data base of a human belief structure | 1969 | IJCAI |
| EWSRD13 | Explicit web search result diversification | 2013 | SIGIR |
| FAATIOAIOS73 | Forecasting and assessing the impact of artificial intelligence on society | 1973 | IJCAI |
| FCNFSS15 | Fully convolutional networks for semantic segmentation | 2015 | CVPR |
| FEFFUDT89 | Feature extraction from faces using deformable templates | 1989 | CVPR |
| FOAITHSUS77 | Focus of attention in the Hearsay-II speech understanding system | 1977 | IJCAI |
| FRUE91 | Face recognition using eigenfaces | 1991 | CVPR |
| GPN77 | Generating project networks | 1977 | IJCAI |
| HEFANLP77 | Human Engineering for Applied Natural Language Processing. | 1977 | IJCAI |
| HMME92 | Hierarchical Model-Based Motion Estimation | 1992 | ECCV |
| HOOGFHD05 | Histograms of oriented gradients for human detection | 2005 | CVPR |
| HTUWYK75 | How to use what you know | 1975 | IJCAI |
| IAA88 | Interpretation as Abduction | 1988 | ACL |
| IAFLPARBOADP90 | Integrated architecture for learning, planning, and reacting based on approximating dynamic programming | 1990 | ICML |
| ICWDCNN12 | ImageNet Classification with Deep Convolutional Neural Networks | 2012 | NIPS |

*Continued on next page*

Table 10 – *Continued from previous page*

| **Initials** | **Title** | **Year** | **Venue** |
|---|---|---|---|
| IEAUWA08 | Intelligent email: aiding users with AI | 2008 | AAAI |
| IFATSSP94 | Irrelevant features and the subset selection problem | 1994 | ICML |
| IMAMFEMFIS69 | Implicational molecules: a method for extracting meaning from input sentences | 1969 | IJCAI |
| INLG01 | Instance-based natural language generation | 2001 | NAACL |
| ISARASAFD77 | Information storage and retrieval: a survey and functional description | 1977 | SIGIR |
| LAAATFTOLA85 | Lexical ambiguity as a touchstone for theories of language analysis | 1985 | IJCAI |
| LELASTTITP89 | Lazy explanation-based learning: a solution to the intractable theory problem | 1989 | IJCAI |
| LPPKICE96 | Learning procedural planning knowledge in complex environments | 1996 | AAAI |
| LRCNFVRAD15 | Long-term recurrent convolutional networks for visual recognition and description | 2015 | CVPR |
| LTGCWCNN15 | Learning to generate chairs with convolutional neural networks | 2015 | CVPR |
| LTRFIR10 | Learning to rank for information retrieval | 2010 | SIGIR |
| LTRNLAAUA98 | Learning to resolve natural language ambiguities: a unified approach | 1998 | AAAI |
| LTRWPD08 | Learning to rank with partially-labeled data | 2008 | SIGIR |

*Continued on next page*

Table 10 – *Continued from previous page*

| Initials | Title | Year | Venue |
|---|---|---|---|
| MIFRVS91 | Multidimensional indexing for recognizing visual shapes | 1991 | CVPR |
| MRLUAV98 | Mobile Robot Localisation Using Active Vision | 1998 | ECCV |
| MSCBSSFVSTN3MSAR01 | Multi-view scene capture by surfel sampling: from video streams to non-rigid 3D motion, shape and reflectance | 2001 | ICCV |
| MSDIAAN71 | Managing semantic data in an associative net | 1971 | SIGIR |
| NATCA18 | Neural Approaches to Conversational AI | 2018 | SIGIR |
| NCFPS90 | NOUN CLASSIFICATION FROM PREDICATE-ARGUMENT STRUCTURES | 1990 | ACL |
| PASTAPW69 | PROW: a step toward automatic program writing | 1969 | IJCAI |
| PAUAODNPID83 | PROVIDING A UNIFIED ACCOUNT OF DEFINITE NOUN PHRASES IN DISCOURSE | 1983 | ACL |
| PEOKIP71 | Procedural embedding of knowledge in planner | 1971 | IJCAI |
| POACBC75 | Progress on a computer based consultant | 1975 | IJCAI |
| POOFT92 | Performance of optical flow techniques | 1992 | CVPR |
| PSGANL83 | Phrase structure grammars and natural languages | 1983 | IJCAI |
| RAKAA77 | Reasoning about knowledge and action | 1977 | IJCAI |
| RODUABCOSF01 | Rapid object detection using a boosted cascade of simple features | 2001 | CVPR |

*Continued on next page*

Table 10 – *Continued from previous page*

| **Initials** | **Title** | **Year** | **Venue** |
|---|---|---|---|
| SAATNICGWVA15 | Show, Attend and Tell: Neural Image Caption Generation with Visual Attention | 2015 | ICML |
| SCASFPM08 | Server characterisation and selection for personal metasearch | 2008 | SIGIR |
| SF83 | Scale-space filtering | 1983 | IJCAI |
| SMIS14 | Semantic Matching in Search | 2014 | SIGIR |
| STFSWA06 | Semi-Supervised Training for Statistical Word Alignment | 2006 | ACL |
| TAOAITACSOKE77 | The art of artificial intelligence: themes and case studies of knowledge engineering | 1977 | IJCAI |
| TAOALHWSE98 | The anatomy of a large-scale hypertextual Web search engine | 1998 | WWW |
| TAVOA77 | Towards automatic visual obstacle avoidance | 1977 | IJCAI |
| TCDAHOITP87 | The classification, detection and handling of imperfect theory problems | 1987 | IJCAI |
| TCOSP89 | Term clustering of syntactic phrases | 1989 | SIGIR |
| THSUSAEOTRP73 | The hearsay speech understanding system: an example of the recognition process | 1973 | IJCAI |
| TMOSAAIPIASMS69 | The modeling of simple analogic and inductive processes in a semantic memory system | 1969 | IJCAI |
| TSHP69 | The Stanford hand-eye project | 1969 | IJCAI |
| TUGIAIE83 | Tracking user goals in an information-seeking environment | 1983 | AAAI |
| TWARIE69 | Talking with a robot in English | 1969 | IJCAI |

*Continued on next page*

Table 10 – *Continued from previous page*

| **Initials** | **Title** | **Year** | **Venue** |
|---|---|---|---|
| UDTTICL93 | Using decision trees to improve case-based learning | 1993 | ICML |
| VMBARR79 | Visual mapping by a robot rover | 1979 | IJCAI |
| WCBSITDWAUSR92 | What can be seen in three dimensions with an uncalibrated stereo rig | 1992 | ECCV |
| WSNGVG88 | What Size Net Gives Valid Generalization | 1988 | NIPS |
| WYCRNTKYLIK92 | What your computer really needs to know, you learned in kindergarten | 1992 | AAAI |

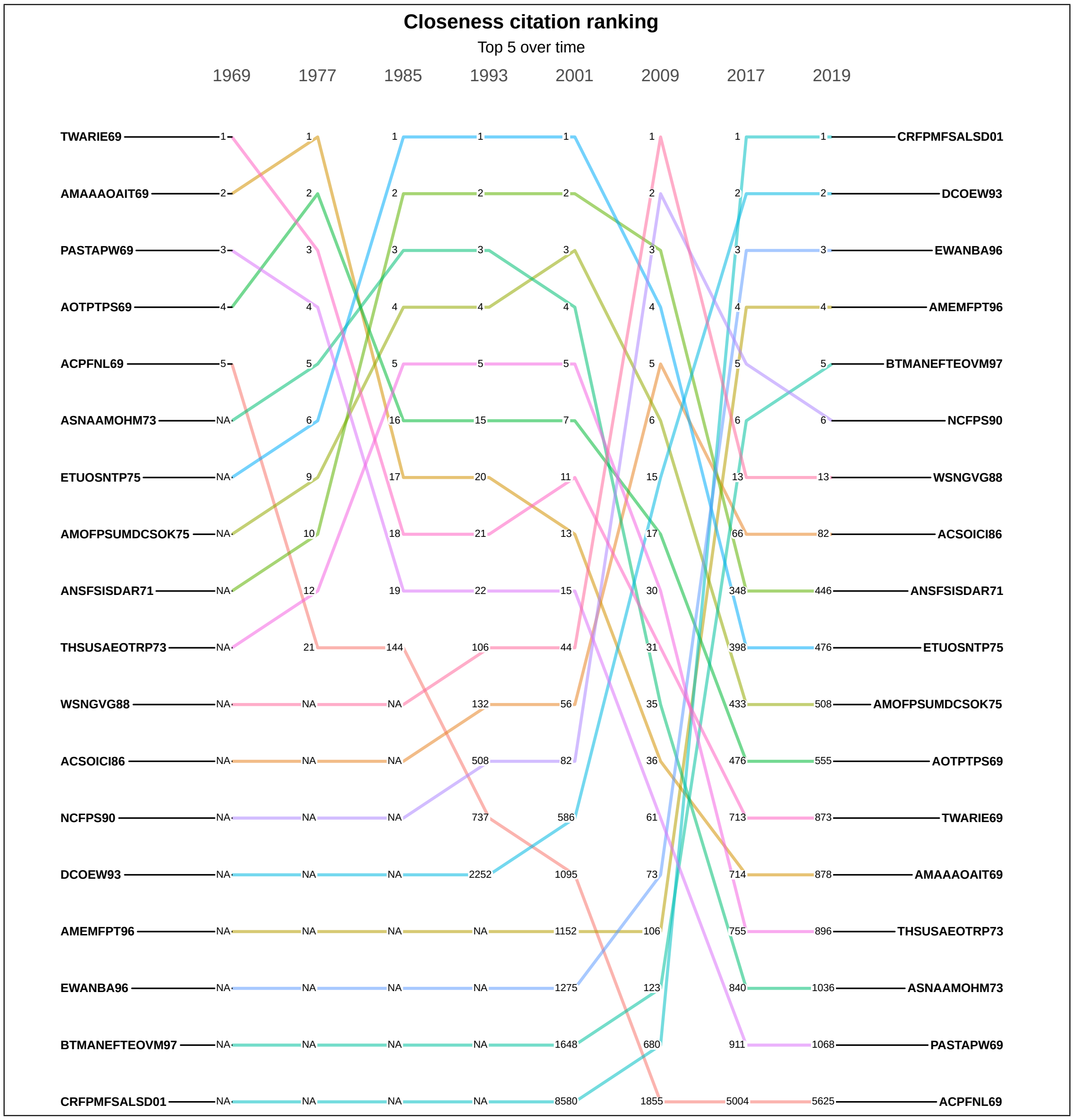

Figure 48: Papers citation ranking over time according to Closeness centrality. *N/A stands for papers that had not been published in the selected venues until that year. Please refer to Table 10 in Appendix E to see the details of each ranked paper.*

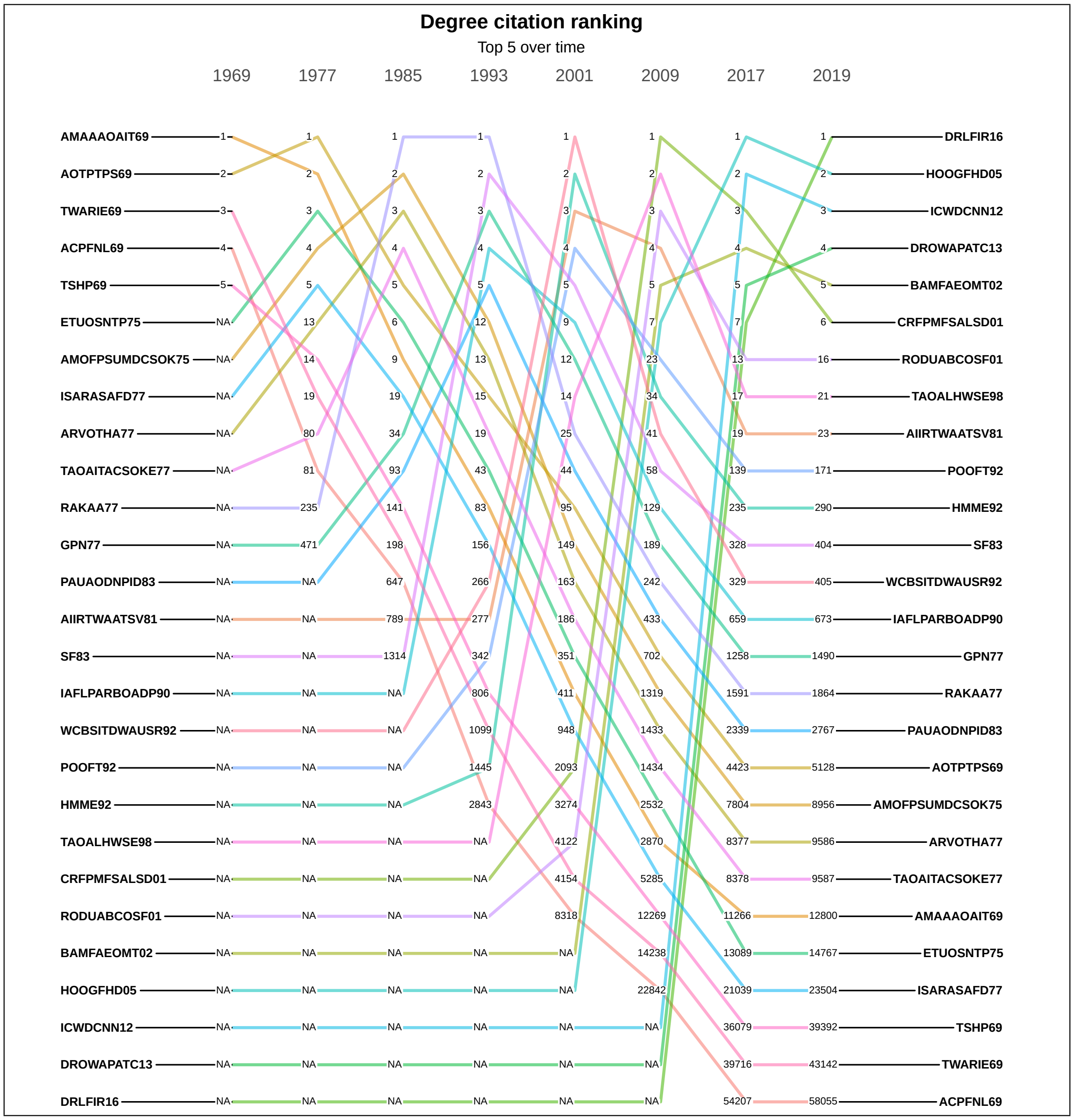

Figure 49: Papers citation ranking over time according to In-Degree centrality. *N/A stands for papers that had not been published in the selected venues until that year. Please refer to Table 10 in Appendix E to see the details of each ranked paper.*

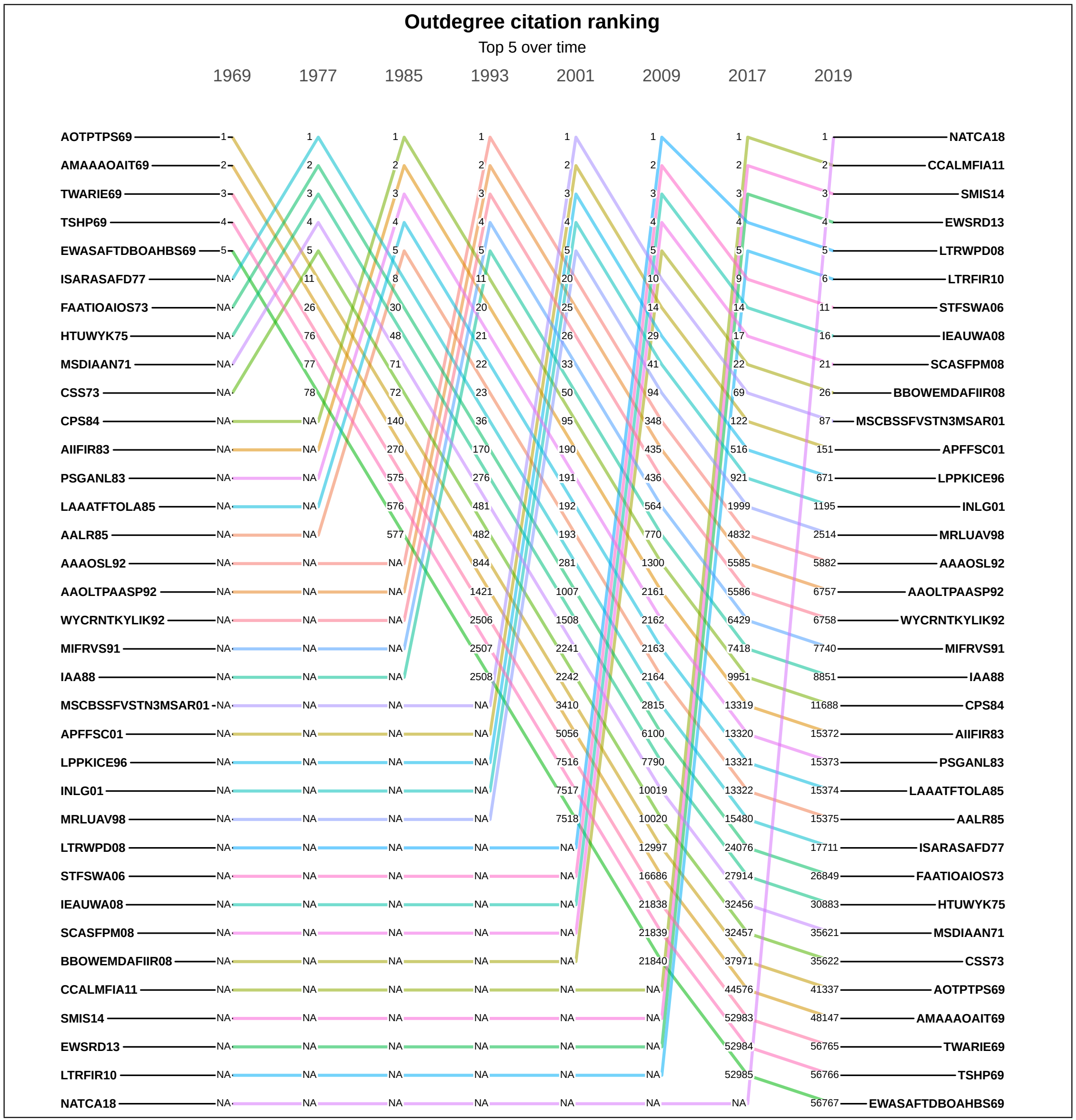

Figure 50: Papers citation ranking over time according to Out-degree centrality. *N/A stands for papers that had not been published in the selected venues until that year. Please refer to Table 10 in Appendix E to see the details of each ranked paper.*

# F Country Citation Graph

Figure 51 presents a different view than the one available at Section 4.6 by generating the stacked version of the countries but with a 2-year-wide sliding average window, i.e. every datapoint is actually the average between the year and its prior window, trying to avoid the variation seen in Figure 37 because of IJCAI being held only in odd-numbered years. The steady decline of the USA share in the graph is clearly seen.

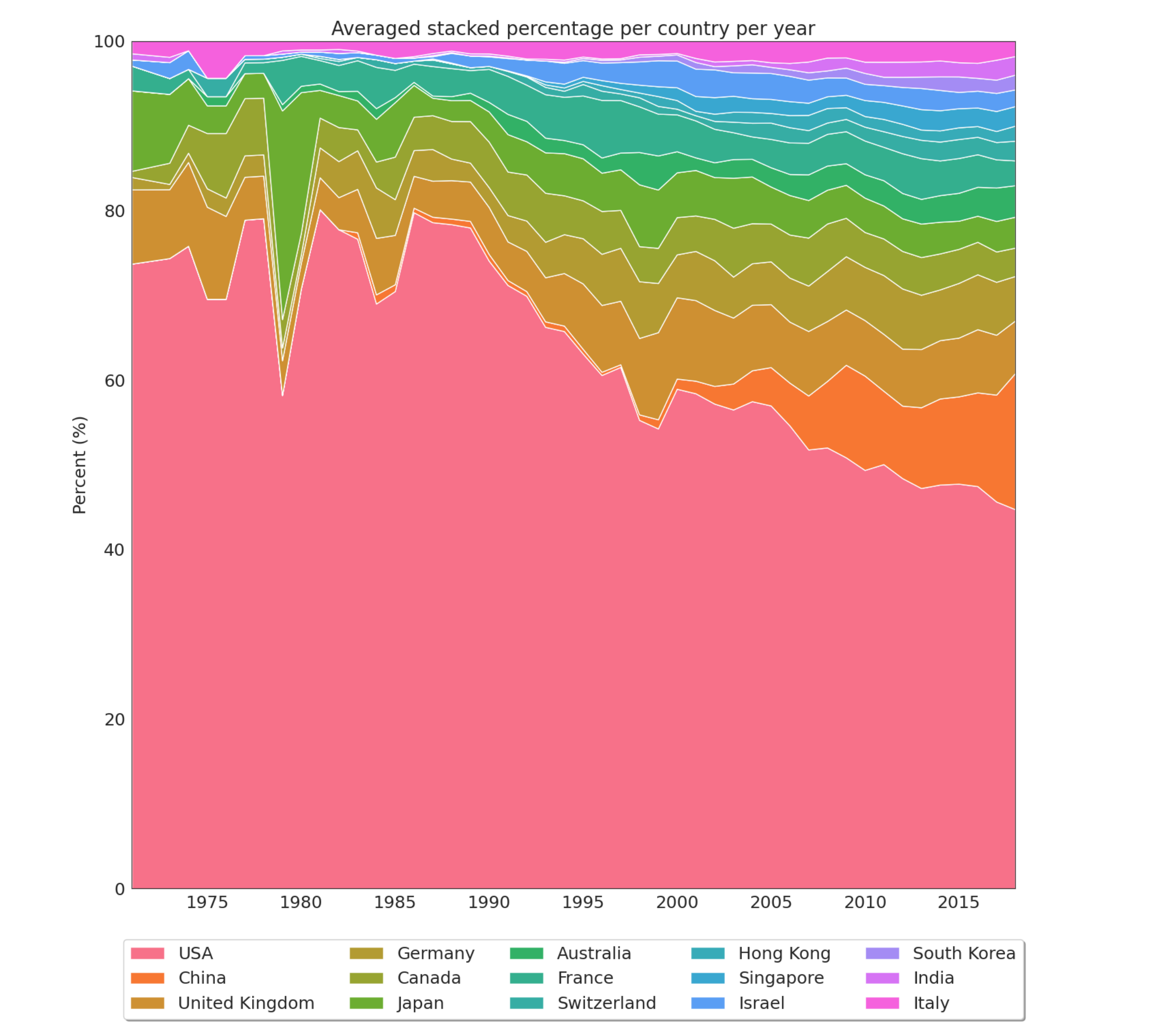

Figure 51: Stacked percentage of papers viewed with a 2-years-wide sliding average window

## G Turing Award Charts

This section presents some charts related to the correlation between Turing Awardees papers abstracts and the other authors' abstracts words. We used a Spearman Correlation to compute these.

Table 11 contains every single Turnig Award winner, with the year they won the prize and also their nationality.

Table 11: Turing Award Winners per year
** indicates the winner is deceased*

| Year | Winner | Nationality |
|---|---|---|
| 1966 | Perlis, Alan J. * | United States |
| 1967 | Wilkes, Maurice V. * | United Kingdom |
| 1968 | Hamming, Richard W. * | United States |
| 1969 | Minsky, Marvin * | United States |
| 1970 | Wilkinson, James Hardy ("Jim") * | United Kingdom |
| 1971 | McMarthy, John * | United States |
| 1972 | Dijkstra, Edsger Wybe * | Netherlands |
| 1973 | Bachman, Charles William * | United States |
| 1974 | Knuth, Donald ("Don") Ervin | United States |
| 1975 | Newel, Allen * | United States |
| | Simon, Herbert ("Herb") Alexander * | United States |
| 1976 | Rabin, Michael Oser | Israel |
| | Scott, Dana Stewart | United States |
| 1977 | Backus, John * | United States |
| 1978 | Floyd, Robert (Bob) W. * | United States |
| 1979 | Iverson, Kenneth E. ("Ken") * | Canada |
| 1980 | Hoare, C. Antony ("Tony") R. | United Kingdom |
| 1981 | Codd, Edgar F. ("Ted") * | United Kingdom |
| 1982 | Cook, Stephen Arthur | United States |

Table 11 – *Continued from previous page*

| Year | Winner | Nationality |
|---|---|---|
| 1983 | Ritchie, Dennis M. * | United States |
| | Thompson, Kenneth Lane | United States |
| 1984 | Wirth, Niklaus E. | Switzerland |
| 1985 | Karp, Richard ("Dick") Manning | United States |
| 1986 | Hopcroft, John E | United States |
| | Tarjan, Robert (Bob) Endre | United States |
| 1987 | Cocke, John * | United States |
| 1988 | Sutherland, Ivan | United States |
| 1989 | Kahan, William ("Velvel") Morton | Canada |
| 1990 | Corbato, Fernando J. ("Corby") * | United States |
| 1991 | Milner, Arthur John Robin Gorell ("Robin") * | United Kingdom |
| 1992 | Lampson, Butler W. | United States |
| 1993 | Hartmanis, Juris | United States |
| | Stearns, Richard ("Dick") Edwin | United States |
| 1994 | Feigenbaum, Edward A. ("Ed") | United States |
| | Reddy, Dabbala Rajagopal ("Raj") | India |
| 1995 | Blum, Manuel | United States |
| 1996 | Pnueli, Amir * | Israel |
| 1997 | Engelbart, Douglas * | United States |
| 1998 | Gray, James ("Jim") Nicholas * | United States |
| 1999 | Brooks, Frederick ("Fred") | United States |
| 2000 | Yao, Andrew Chi-Chih | China |
| 2001 | Dahl, Ole-Johan * | Norway |
| | Nygaard, Kristen | Norway |
| 2002 | Adleman, Leonard (Len) Max | United States |

Table 11 – *Continued from previous page*

| Year | Winner | Nationality |
|---|---|---|
| | Rivest, Ronald (Ron) Linn | United States |
| | Shamir, Adi | Israel |
| 2003 | Kay, Alan | United States |
| 2004 | Cerf, Vinton ("Vint") Gray | United States |
| | Kahn, Robert ("Bob") Elliot | United States |
| 2005 | Naur, Peter * | Denmark |
| 2006 | Allen, Frances ("Fran") Elizabeth * | United States |
| 2007 | Clarke, Edmund Melson * | United States |
| | Emerson, E. Allen | United States |
| | Sifakis, Joseph | France |
| 2008 | Liskov, Barbara | United States |
| 2009 | Thacker, Charles P. (Chuck) * | United States |
| 2010 | Valiant, Leslie Gabriel | United Kingdom |
| 2011 | Pearl, Judea | Israel |
| 2012 | Goldwasser, Shafi | United States |
| | Micali, Silvio | Italy |
| 2013 | Lamport, Leslie | United States |
| 2014 | Stonebraker, Michael | United States |
| 2015 | Diffie, Whitfield | United States |
| | Hellman, Martin | United States |
| 2016 | Bernes-Lee, Tim | United Kingdom |
| 2017 | Hennesy, John L. | United States |
| | Patterson, David | United States |
| 2018 | Bengio, Yoshua | Canada |
| | Hinton, Geoffrey E. | United Kingdom |

Table 11 – *Continued from previous page*

| Year | Winner | Nationality |
|---|---|---|
| | LeCun, Yann | France |
| 2019 | Catmull, Edwin E. | United States |
| | Hanrahan, Patrick M. | United States |
| 2020 | Aho, Alfred Vaino | Canada |
| | Ullman, Jeffrey David | United States |
| 2021 | Dongarra, Jack | United States |

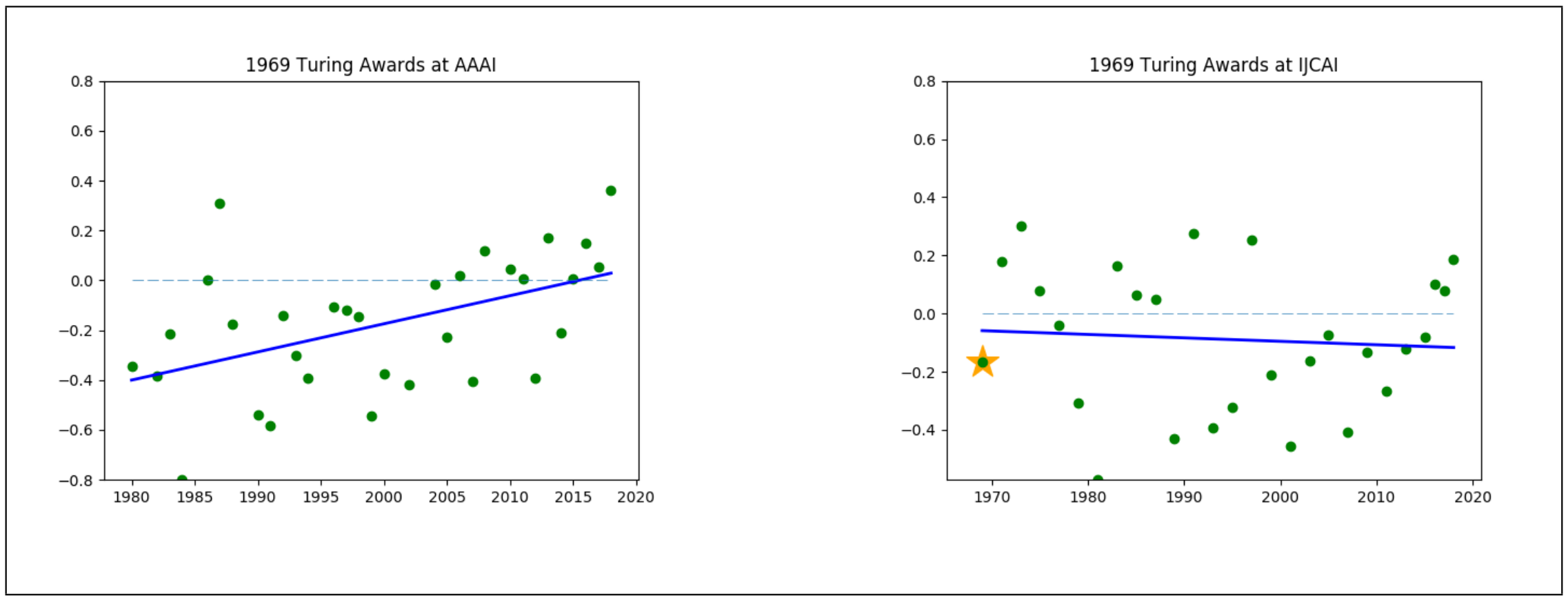

Figure 52: Correlation between 1969 Turing Award Winner papers and AAAI and IJCAI-published ones.

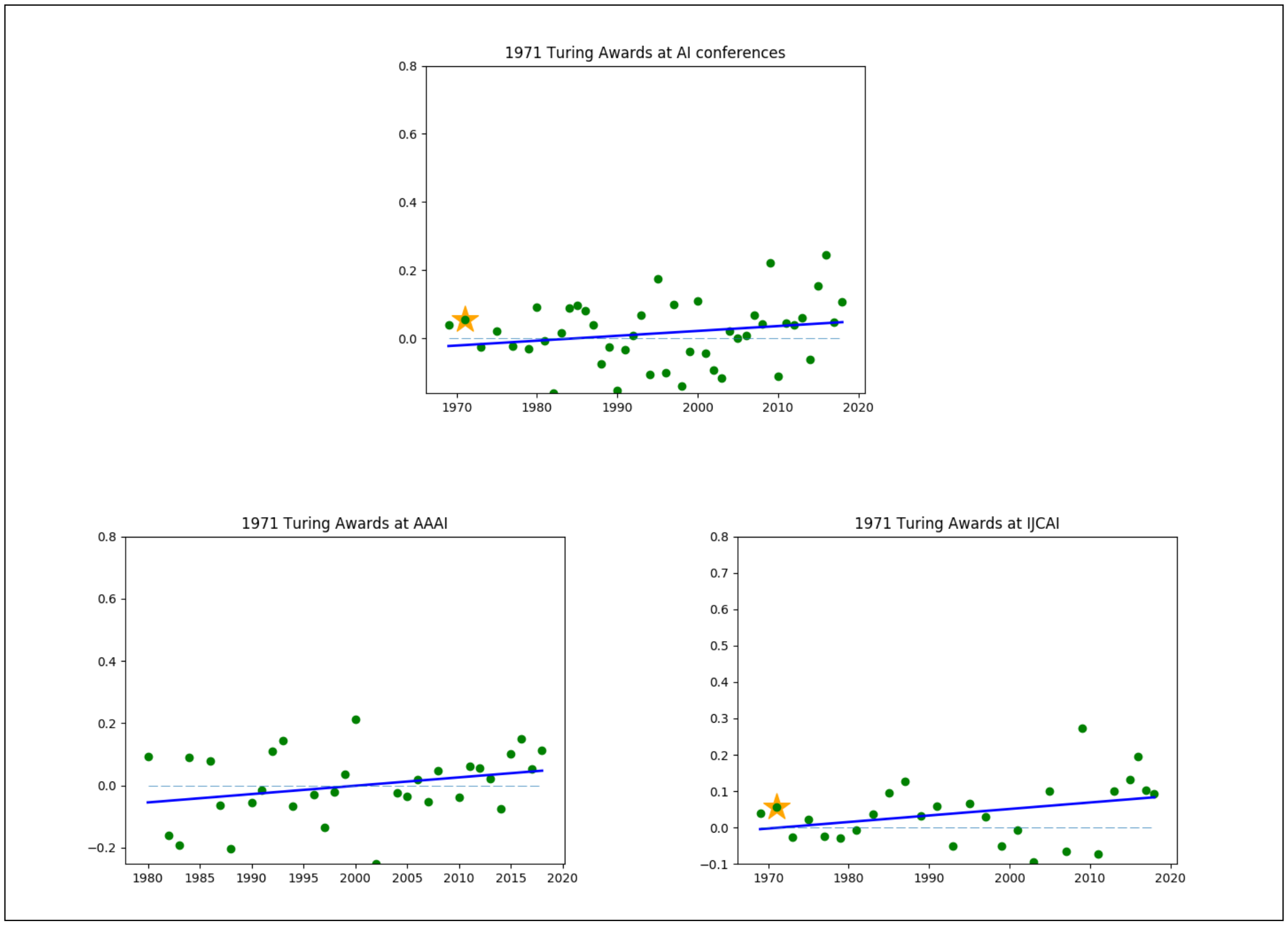

Figure 53: Correlation between titles of papers published by the 1971 Turing Award winner and titles of papers published in the three AI flagship conferences.

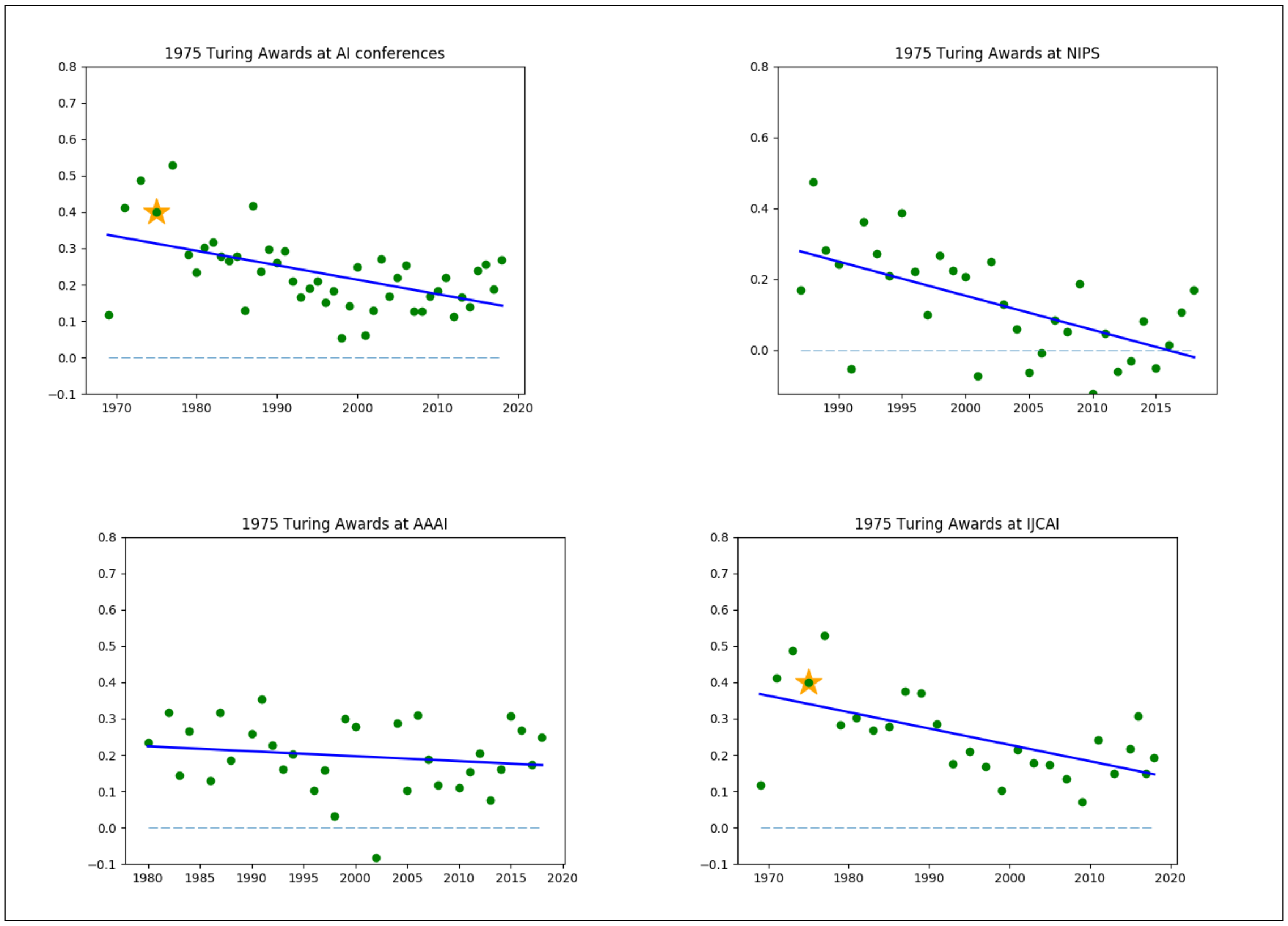

Figure 54: Correlation between titles of papers published by the 1975 Turing Award winners and titles of papers published in the three AI flagship conferences.

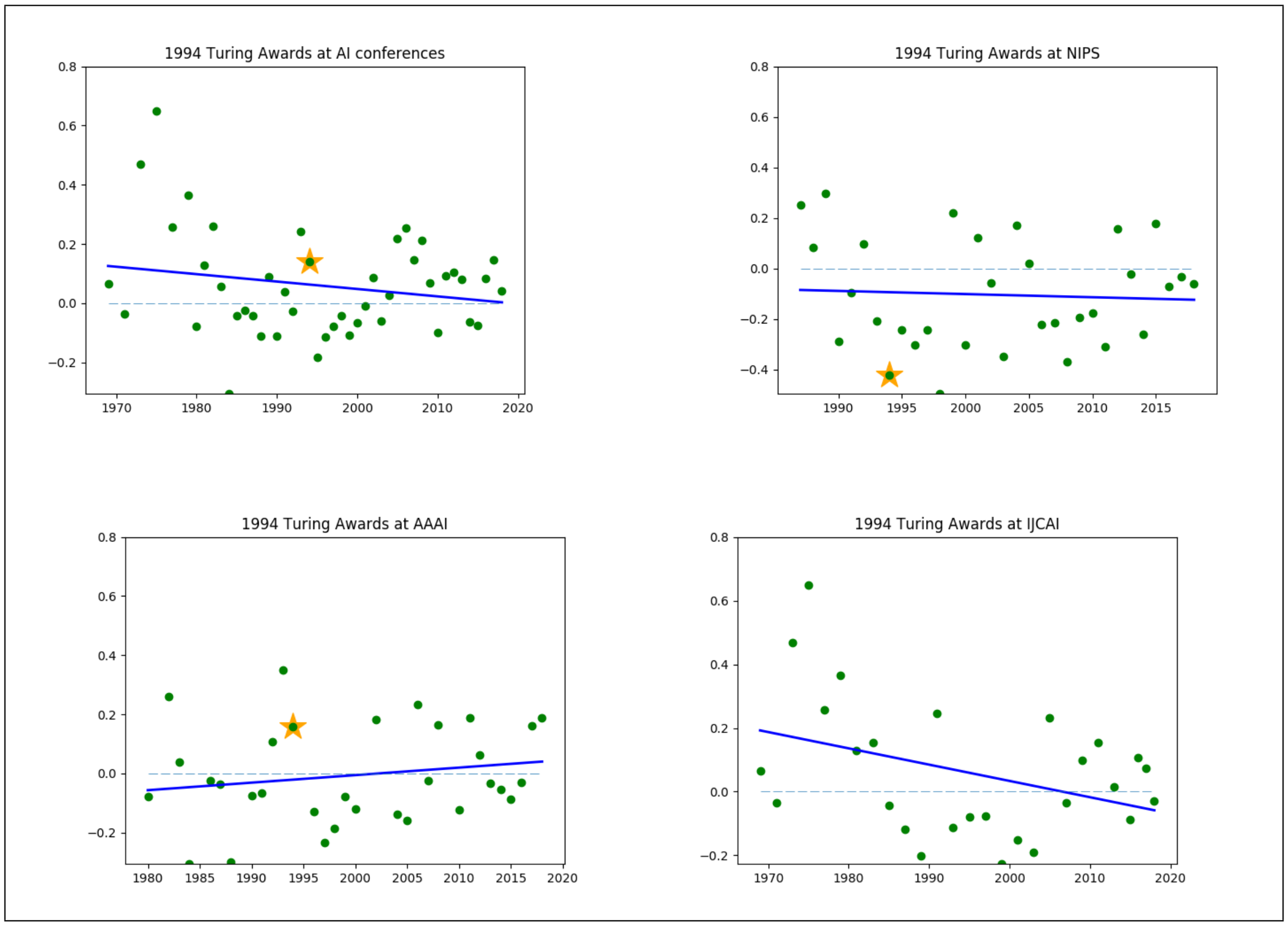

Figure 55: Correlation between titles of papers published by the 1994 Turing Award winners and titles of papers published in the three AI flagship conferences.

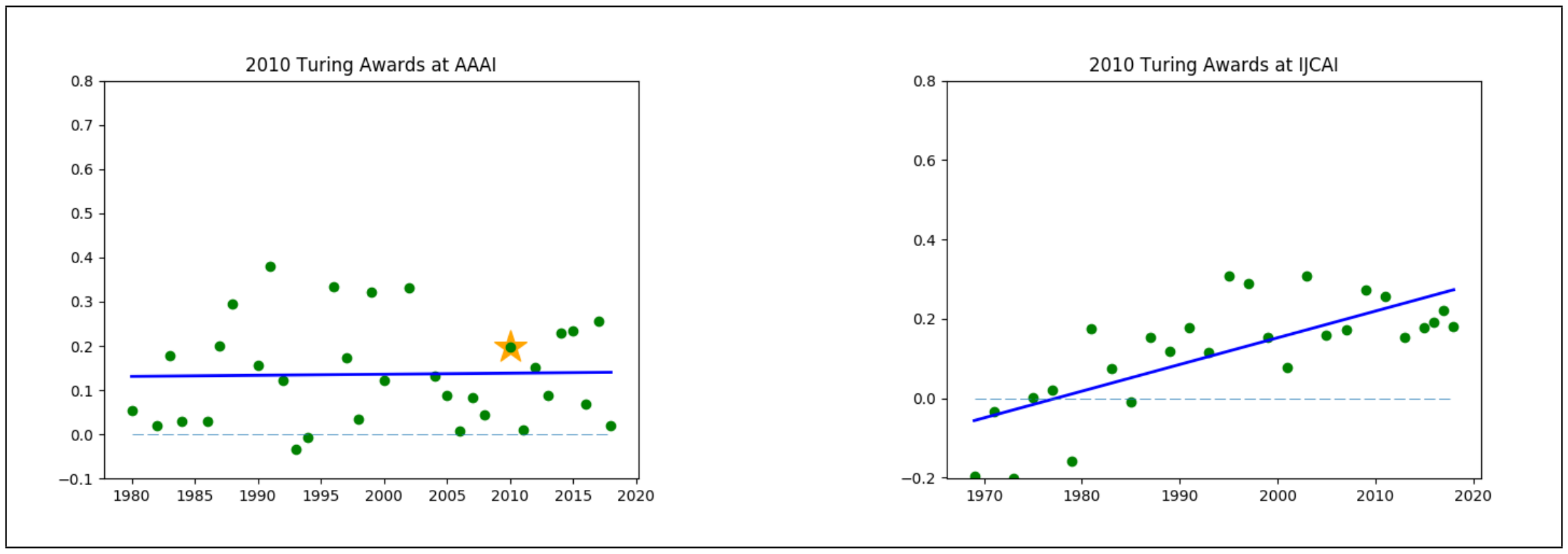

Figure 56: Correlation between titles of papers published by the 2010 Turing Award winners and titles of papers published in AAAI and IJCAI.

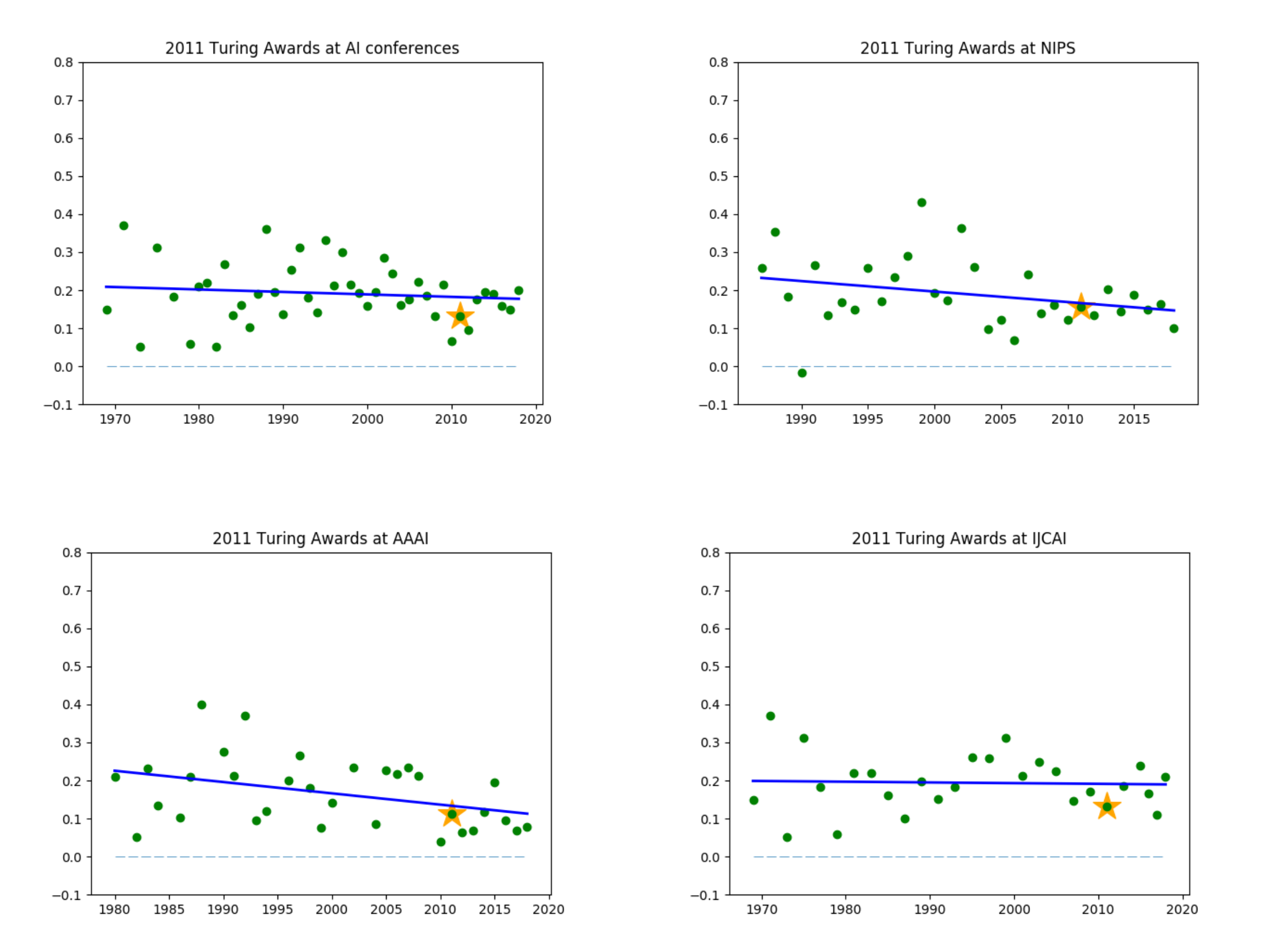

Figure 57: Correlation between titles of papers published by the 2011 Turing Award winner and titles of papers published in the three AI flagship conferences.

# H Software Contributions

To carry out this work, we have built some pieces of software that might be used by others in similarly sized tasks. They are briefly described and discussed below.

## H.1 *streamxml2json* Library

When we were trying to use the original DBLP dataset (See Section 3.1 for more context) we had some trouble when trying to convert the downloaded XML file to a JSON file we could more easily manipulate. The benefits of a JSON file over the XML file go from being more human-readable to the fact of it being a bit smaller (in our case, a 3.3GB XML yields a 3GB JSON file, a 10% size reduction) – therefore, easier to load in memory. It is a fact, however, that because we had the intention to parse this file in a CI environment, to be able to generate new charts (See Section 5.2) weekly we would need to be able to do this conversion from XML to JSON without loading the whole file into memory. After searching on Github and PyPi we realized that a tool to convert from XML to JSON without loading the whole file in memory did not exist.

That clarified, we decided we could build such a tool by using a few already existent libraries as building blocks: simplejson[48], jsonstreams[49] and xmltodict[50]. Streaming over any XML file and parsing only the necessary data, we can then output it to a JSON file, also through a file stream, without any substantial memory usage. Because our data was gzipped, the library supports reading directly from a *.xml.gz* file, not requiring the user to unzip it.

The library *streamxml2json* Audibert [2022a] Audibert [2022b] is available at Github in `https://github.com/rafaeelaudibert/streamxml2json` and publicly downloadable from PyPi on `https://pypi.org/project/streamxml2json/`. For the sake of completeness, the library can be downloaded if you have *pip* Developers [2008] installed in your machine by running *"pip install streamxml2json""*.

In the end, because we did not use this dataset, we do not use this library in our work, but the contribution was deemed important enough to the whole Python ecosystem in general so we are adding it to this section. We did keep in our main repository the file used to convert from XML to JSON, as a library usage example: `https://github.com/rafaeelaudibert/conferences_insights/blob/v11/scripts/xml2json.py`.

[48]`https://github.com/simplejson/simplejson`
[49]`https://github.com/dcbaker/jsonstreams`
[50]`https://github.com/martinblech/xmltodict`

## H.2 Python Parallel Centralities Implementation

Throughout our work, we used UFRGS HPC Group's (PCAD[51] supercomputers to be able to properly generate the graph we were building. We had to use their supercomputers because when we are computing graph centralities we need a lot of memory - for betweenness, we need to store the shortest path between every single node of our graph that contains more than 100,000 nodes. Computing these centralities, however, was still pretty slow because we have to do it for every single node in every single year. An easy way to increase speed in computation, especially when you are using supercomputers, is to parallelize your job across the available physical processors. In our case, we had access to a machine with 16 cores (32 threads) allowing us to compute our results a lot faster.

Therefore, using Networkx's implementations as a base, we developed a parallel Betweenness and a parallel Closeness algorithm capable of running close to 5x faster in a machine with 16 cores. The results are not 16x faster than expected because of Python's GIL which severely degrades Python's parallel performance.

The codes for these implementations can be found in Github. [52] [53].

## H.3 Graph Parsing pipeline

In our work, we had to generate several different types of graphs, with several different parameters in each of them. We also wanted to be able to easily cache data we had already computed, avoiding unnecessary computation.

To solve these problems, we devised a simple structure where we could extend a base *GenerateGraph* class (available in `https://github.com/rafaeelaudibert/conferences_insights/blob/v11/graph_generation/generate_graph.py`) that exposed several methods that made our job easier. Some of the exposed methods help us in the process of caching our data. Whenever we want to build a new graph, if we have no caching, we need to do these steps:

1. Filter papers from the required venues from DBLP's JSON file
2. Generate the full graph for every year
3. After the full graph is complete, compute the centralities

[51] `http://gppd-hpc.inf.ufrgs.br/`

[52] `https://github.com/rafaeelaudibert/conferences_insights/blob/v11/graph_generation/parallel_betweenness.py`

[53] `https://github.com/rafaeelaudibert/conferences_insights/blob/v11/graph_generation/parallel_closeness.py`

If we always followed these steps, whenever we made a code change to the centralities computation, we would need to run everything before. We can easily solve this by calling some of the base class helper methods that know how to save a pre-parsed list of papers from selected venues, or even an already partial graph if we had only built it until a given year (imagine you noticed something wrong or an exception was raised after you had parsed half the dataset).

Also, to be able to control which type of graph we wanted to run from the command line, we built a CLI on top of this class using Google's fire[54] library. It is used to automatically generate a CLI from the parameters of a function, effectively allowing us to simply add a new parameter to a function and then pass the parameter value from the command line to properly pass the parameters to our code.

When we want a new type of graph, therefore, we simply extend this *GenerateGraph* class and add a new parameter to the main function, allowing us to easily call this new type of graph generation.

It is worth noting, however, that, ideally, *fire* should be replaced by *click*. Click[55] is a more maintained library, with better features: automatically generated fully-customizable help command; subcommands to avoid the extra work of manually creating flags when creating new types of graphs; proper filename handling; and etc.

[54] `https://github.com/google/python-fire`

[55] `https://click.palletsprojects.com/en/8.1.x/`